\documentclass[10pt]{article} 
\usepackage[accepted]{tmlr}
\usepackage{float}
\usepackage{wrapfig}
\usepackage{enumitem}
\usepackage{graphicx}
\usepackage{soul,xcolor}
\usepackage{subfigure}
\usepackage{multirow}
\usepackage{multicol}
\usepackage{subcaption}

\usepackage{pifont}
\usepackage{amsmath}
\usepackage{xcolor}
\usepackage{subfigure}
\newcommand{\cmark}{\ding{51}}%
\newcommand{\xmark}{\ding{55}}%
\usepackage{url}

\usepackage{amsmath,amsfonts,bm}

\def\eqref#1{equation~\ref{#1}}

\def\1{\bm{1}}

\DeclareMathAlphabet{\mathsfit}{\encodingdefault}{\sfdefault}{m}{sl}
\SetMathAlphabet{\mathsfit}{bold}{\encodingdefault}{\sfdefault}{bx}{n}

\setstcolor{blue}
\usepackage{hyperref}
\usepackage{url}

\title{A Function-Centric Perspective on Flat and Sharp Minima}

\author{\name Israel Mason-Williams \email israel.mason-williams@kcl.ac.uk \\
      \addr Department of Informatics \\
      King's College London and Imperial College London (UKRI Safe and Trusted AI)\\
      London, United Kingdom
      \AND
      \name Gabryel Mason-Williams \email g.t.mason- williams@qmul.ac.uk \\
      \addr Queen Mary University of London\\
      London, United Kingdom
      \AND
      \name Helen Yannakoudakis \email helen.yannakoudakis@kcl.ac.uk\\
      \addr Department of Informatics\\
      King's College London \\
      London, United Kingdom}

\def\month{10}  
\def\year{2026} 
\def\openreview{\url{https://openreview.net/forum?id=LV1ffpXXeV}}

\begin{document}

\maketitle

\begin{abstract}
Flat minima are strongly associated with improved generalisation in deep neural networks. 
However, this connection has proven nuanced in recent studies, with both theoretical counterexamples and empirical exceptions emerging in the literature. In this paper, we revisit the role of sharpness in model performance and examine sharpness as a function-dependent property rather than as a direct indicator of poor generalisation. We conduct extensive empirical studies ranging from single-objective optimisation, synthetic non-linear binary classification tasks, to modern image classification tasks. In single-objective optimisation, we show that flatness and sharpness are relative to the function being learned: equally optimal solutions can exhibit markedly different local geometry. In synthetic non-linear binary classification tasks, we show that increasing decision-boundary tightness can increase sharpness even when models generalise perfectly, suggesting that sharpness is not reducible to memorisation alone. Finally, in large-scale experiments, we find that sharper minima often emerge when models are regularised (e.g., via weight decay, data augmentation, or SAM), and can coincide with better generalisation, calibration, robustness, and functional consistency. Our findings are consistent with a function-centric interpretation of minima geometry: flatness should not be treated as a universal proxy for generalisation, and sharper minima can reflect desirable inductive biases associated with improved predictive behaviour, calling for a function-centric reappraisal of minima geometry.
\end{abstract}

\section{Introduction}
\label{sec:introduction}

Neural network architectures with different implicit biases are known to exhibit distinct geometric properties, with flatness often associated with improved generalisation performance via reduced generalisation gaps \citep{li2018visualizing,han2025flatness,petzka2021relative,lee2025flat,cha2021swad,zhao2022penalizing}. This perspective is commonly associated with the idea that flat minima correspond to wide error margins and increased robustness, in line with Occam's Razor and Minimum Description Length (MDL) arguments~\citep{hochreiter1994simplifying}. 
Empirical and theoretical studies have reinforced this perspective~\citep{kaddour2022flat, foret2021sharpnessaware,petzka2021relative}, supporting the view that flatter solutions lead to better generalisation. Flatness has also been associated with benefits such as improved representation transfer~\citep{pmlr-v202-liu23ao} and the effects of architectural choices such as residual connections~\citep{li2018visualizing}. Notably, optimisation methods such as Sharpness Aware Minimization (SAM)~\citep{foret2021sharpnessaware}, which improve generalisation in the vision domain, were explicitly motivated by the idea of biasing training toward flatter minima. Yet generalisation is only one dimension of model quality.

At the same time, this connection has proven more nuanced than these accounts suggest. \cite{dinh2017sharp} showed that flat minima, under Hessian-based definitions, can be arbitrarily sharpened through reparameterisation, without changing the model's learned function or generalisation behaviour. This motivated the development of reparameterisation-invariant sharpness metrics, such as the Fisher-Rao norm~\citep{pmlr-v89-liang19a} and Relative-Flatness~\citep{petzka2021relative}, which have been used to recover the association between flatness and generalisation. However, these developments leave open a central question: does sharpness, under reparametrisation invariant metrics, indicate poor generalisation, or does its interpretation depend on the function learned by the model and the training conditions under which that function is obtained?

This question becomes especially important once model quality is assessed beyond test accuracy alone. In practical settings, reliability-related properties such as calibration \citep{guo2017calibration}, robustness to common corruptions \citep{hendrycks2019benchmarking}, and functional consistency across runs \citep{wang2024great} are also important for reliable deployment. However, their relationship to minima geometry remains underexplored. In particular, it is unclear whether flatter solutions consistently support better reliability, or whether higher-performing models on these dimensions may occupy sharper regions of the loss landscape.

In this paper, we revisit minima geometry through a function-centric lens. Specifically, we use the term \textit{function complexity} to refer to properties of the learned function such as representation constraints, decision-boundary structure, or degree of representational precision, that may be reflected in minima geometry. Our hypothesis is that the geometry of a solution reflects the complexity of the learned function, rather than directly determining performance. Our central claim is not that sharpness is universally beneficial, nor that flatness is irrelevant. Rather, we argue that sharpness should be interpreted relative to the learned function and the model's inductive biases. From this perspective, sharper minima need not indicate overfitting or memorisation, but instead may reflect more structured or better-regularised solutions, particularly in high-dimensional learning tasks.

Our function-centric framing is motivated by Occam's Razor and the MDL perspective commonly used in the flatness literature, where flatter minima are often interpreted as simpler, lower-complexity solutions that require less precision to represent \citep{hochreiter1994simplifying}. From this perspective, sharper minima can be viewed as corresponding to learned solutions that are more tightly constrained or that require more precise representation, and are therefore related to more complex decision-boundary structure 
in an Occam/MDL sense. We use function complexity as an interpretive lens for understanding how minima geometry relates to the learned function, inductive bias, and training regularisation.

We first study this in controlled synthetic settings. We begin with seven standard single-objective optimisation problems, where the global minima are known and their local geometry can be compared directly. These experiments illustrate that optimal solutions can be either sharp or flat, depending on the intrinsic complexity of the objective being fitted: some functions (e.g., Sphere) have flat global minima, while others (e.g., Rosenbrock) have inherently sharp global minima. This suggests that the geometry of the solution space is tied to function complexity, not optimality alone.
We then turn to non-linear binary classification tasks. 
Such settings have been used to show that sharp minima have an implicit relationship with memorised solutions, arguing that sharp minima are an indicator of poorly generalising models~\citep{pmlr-v137-huang20a}.
We revisit this relationship by explicitly manipulating decision-boundary tightness and showing that models can generalise perfectly while still exhibiting increased sharpness under tighter decision boundaries. This allows us to partially decouple sharpness from memorisation alone and to show that the decision boundaries of the data can shape the geometry of the solution. 
Additionally, we show that learning tight decision boundaries requires increased computational budgets (with all other training conditions fixed), indicating that such functions may be harder for neural networks to fit.

We then scale our analysis to high-dimensional problems, and use the CIFAR~\citep{krizhevsky2009learning} and TinyImageNet~\citep{le2015tiny} data sets to train the ResNet~\citep{he2016deep}, VGG~\citep{simonyan2014very} and ViT~\citep{dosovitskiy2020image} architectures under multiple matched-seed training controls. We compare baseline models to models trained with commonly used regularisation techniques (weight decay, data augmentation and SAM). Across these settings, we evaluate reparameterisation-invariant sharpness metrics, together with generalisation performance and reliability-related measures: average-case perturbation robustness, expected calibration error and functional/prediction agreement to provide a broad study of geometry, generalisation and reliability.

\begin{figure}[htb]
    \centering
    \subfigure[Fisher-Rao norm and Loss]{\includegraphics[width=0.32\linewidth] {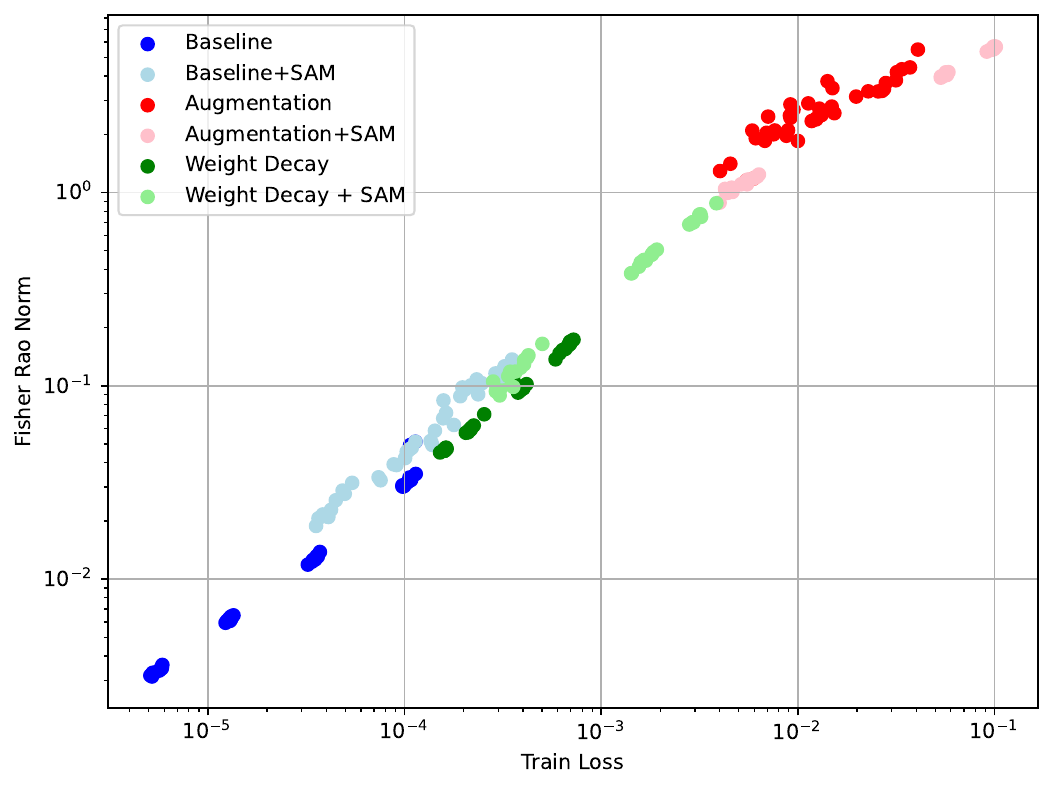}}
    \subfigure[Relative Flatness and Loss]{\includegraphics[width=0.315\linewidth] {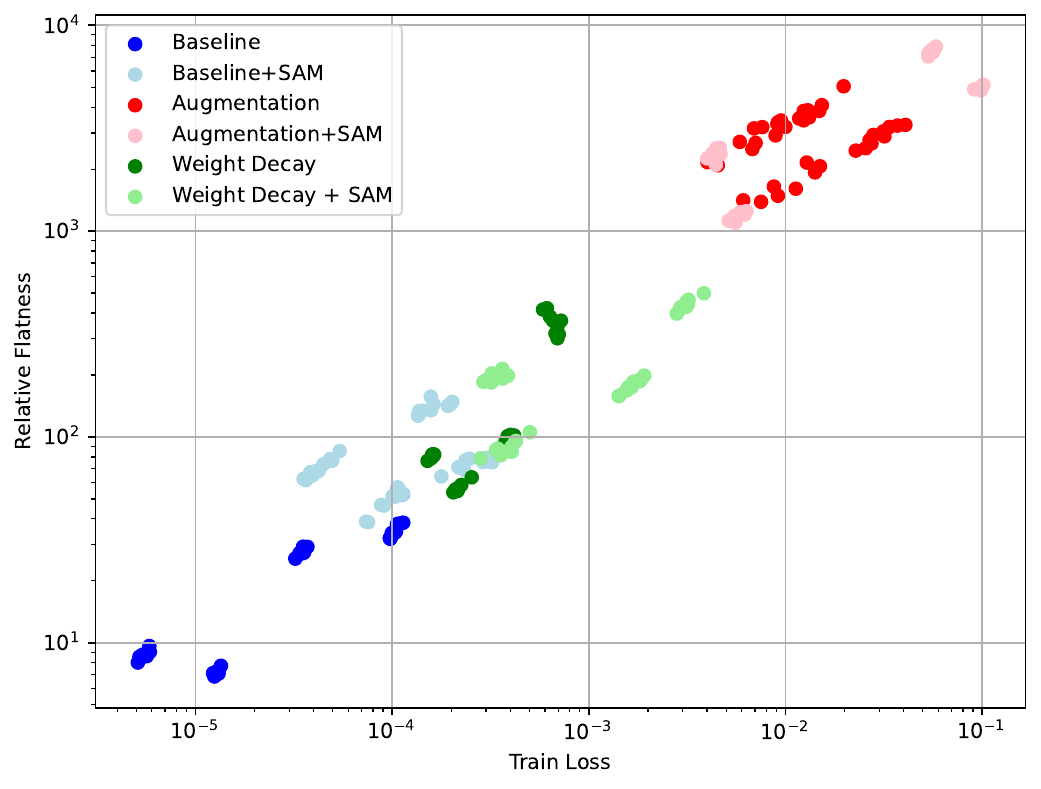}}
    \subfigure[Generalisation Gap and Loss]{\includegraphics[width=0.32\linewidth] {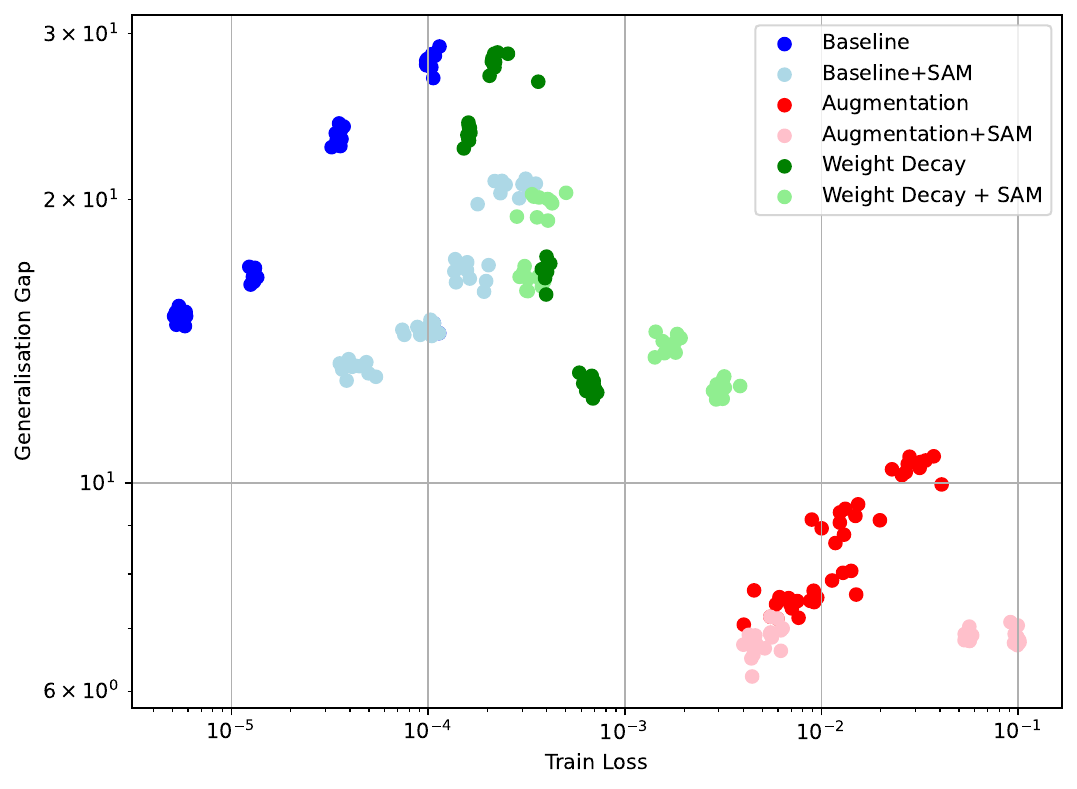}}
        \centering
    \subfigure[Fisher-Rao norm and Loss]{\includegraphics[width=0.32\linewidth] {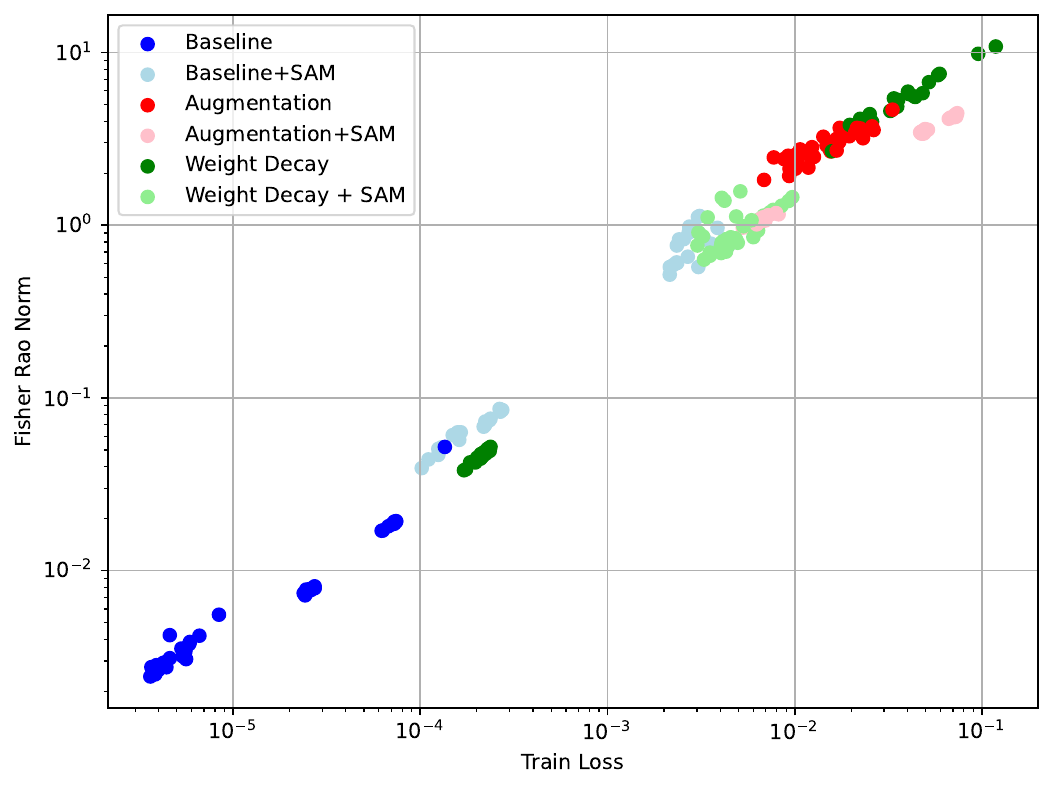}}
    \subfigure[Relative Flatness and Loss]{\includegraphics[width=0.315\linewidth] {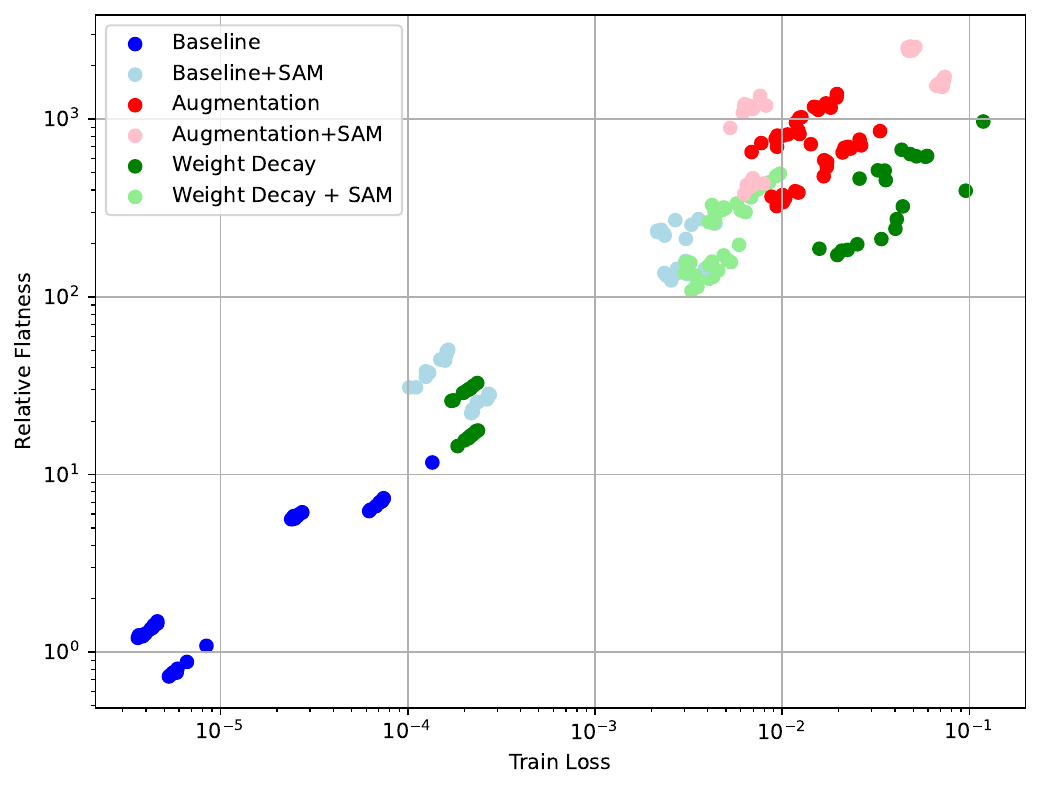}}
    \subfigure[Generalisation Gap and Loss]{\includegraphics[width=0.32\linewidth] {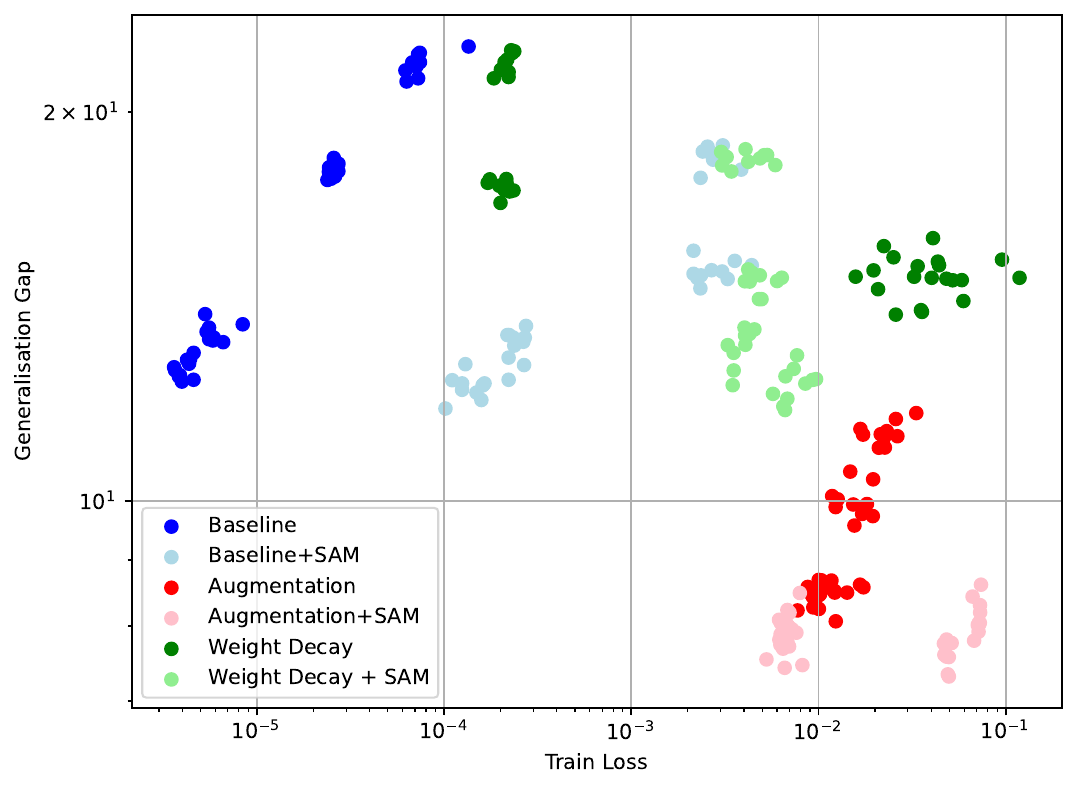}}
    \caption{Scatter plots of 240 minima for ResNet-18 (top) and VGG19 (bottom) on CIFAR-10 across batch size ${128, 256}$ and learning rate ${10^{-3},10^{-2}}$: (a) Fisher–Rao norm vs. train loss, (b) Relative Flatness vs. train loss, and (c) generalisation gap vs. train loss (log scale). Full results for ResNet and VGG in Appendix~\ref{sec:ResNet_param_sweep} and~\ref{sec:vgg_param_sweep}, respectively.}
   \label{fig:sharpness_gen_gap_plot}
\end{figure}

Our findings provide empirical support for a function-centric interpretation of sharpness: models trained with regularisation often reach sharper minima while also outperforming flatter, unregularised counterparts across reliability and generalisation metrics (Figure~\ref{fig:sharpness_gen_gap_plot}). This suggests that regularisation can give rise to learned solutions of greater functional complexity, 
yielding sharper yet more effective minima. At the same time, the relationship is not monotone or universal: distinct trends emerge across control conditions, with some favouring flatter solutions across seeds and others not, consistent with Simpson's paradox~\citep{simpson1951interpretation}. Minima geometry therefore requires a nuanced interpretation rather than a universal preference for flatness: while models without regularisation (\textcolor{blue}{\textbf{blue}} in Figure~\ref{fig:sharpness_gen_gap_plot}) have a strong correlation with small generalisation gaps and flatness, models using augmentation and SAM (\textcolor{pink}{\textbf{pink}} in Figure~\ref{fig:sharpness_gen_gap_plot}) show no such preference. While SAM and related methods were originally motivated by the goal of encouraging flatness, we show that their empirical benefits frequently arise alongside increased sharpness. Taken together, these results call into question the assumption that flatness is inherently beneficial and support a reappraisal of sharpness through the lens of function complexity. We further show that these findings are robust to training hyperparameters such as learning rate and batch size.

Concretely, we make the following contributions:
\begin{itemize}
    \item We present a function-centric interpretation of sharpness, in which the geometry of minima reflects the complexity of the learned function rather than serving as a universal proxy for (poor) generalisation.
    
    \item We provide a broad empirical reassessment of minima geometry across controlled optimisation problems, synthetic classification tasks, and high-dimensional deep learning tasks, and find that sharper minima can coincide with statistically significant improvements in generalisation, calibration, robustness and functional consistency.
    
    \item We find across architectures, data sets, hyperparameters and matched-seed training controls that widely used regularisation techniques (e.g., weight decay, augmentation and SAM) often induce statistically significant sharper minima than unregularised baselines, calling into question the assumption that regularisation generally promotes flatter solutions.
    
    \item We demonstrate that sharpness cannot be meaningfully interpreted across architectures or tasks without accounting for function complexity and inductive bias, cautioning against overgeneralised geometric claims.
\end{itemize}

Our findings can be summarised as follows: 
\begin{enumerate}
\itemsep0pt
    \item Sharpness varies across global minima in single-objective optimisation and with decision-boundary proximity in non-linear binary classification tasks, reflecting function complexity rather than solution quality alone. In high-dimensional problems, regularised models typically reach sharper minima, yet often achieve better generalisation, calibration, robustness and functional consistency than flatter unregularised baselines.

    \item We reconcile SAM's local robustness objective with increased global sharpness, showing that its empirical benefits need not arise from flatter minima and are consistent with a function-centric view of geometry.
    
    \item Our results support a function-centric view of sharpness: solution geometry is shaped by the complexity of the learned function and the model's inductive biases. Crucially, we find no clear Goldilocks zone for sharpness across architectures and data sets, as the interpretation of sharpness depends on the function being learned, an architectures implicit bias and the training conditions under which the model is optimised.
\end{enumerate}

\section{Related Work}
\cite{hochreiter1997flat} presented seminal empirical evidence that neural networks can adhere to Occam's Razor. They showed that a flat-minima search algorithm using a second-order Hessian approximation could yield the smallest generalisation gap on two-class classification problems. As a result of the observed empirical relationship between flatness and generalisation, the antipodal sharp minima were often viewed as undesirable. The importance of flatness in more complex learning tasks was later reaffirmed by~\cite{li2018visualizing} who introduced landscape visualisation to study the geometry of deep networks. They argued that skip connections help prevent explosions of non-convexity, reducing the occurrence of chaotic plateaus often associated with sharp minima. 
Building on this, Sharpness Aware Minimisation~\citep{foret2021sharpnessaware} was proposed as an optimisation method (motivated in part by~\cite{hochreiter1997flat}) that explicitly aims to reduce sharpness in the loss landscape. SAM has yielded strong empirical performance gains over traditional optimisation methods~\citep{foret2021sharpnessaware}. However, some recent work has challenged this interpretation, arguing that SAM does not necessarily find flatter minima~\citep{wen2023sharpness}. At the same time, sharp minima are still commonly associated in the literature with memorisation and poor generalisation~\cite{pmlr-v137-huang20a}. 

The necessity of flatness for generalisation has also been questioned more fundamentally. Notably,~\cite{dinh2017sharp} showed that sharpness can be arbitrarily increased through reparameterisation without affecting the learned function or its generalisation, casting doubt on the intrinsic value of flatness under naive Hessian-based definitions. In response, reparameterisation-invariant sharpness metrics were developed~\citep{petzka2021relative} and have since been used to reaffirm the association between flatness and generalisation. Together, these developments highlight a conceptual tension: although sharpness was shown to be manipulable through reparameterisation, reparametrisation invariant metrics have reaffirmed a preference for flatness and, as such, flatness is still widely used as a desirable indicator of generalisation. 

More recent work has continued to recover a relationship between flat minima and strong generalisation in neural network optimisation.~\cite{stutz2021relating} conducted a study which related average-case flatness to robustness under adversarial perturbation.~\cite{cha2021swad} introduced Stochastic Weight Averaging Densely (SWAD), which they argue improves domain generalisation by seeking flatter minima, supported by loss-landscape visualisations.~\cite{zhao2022penalizing} built on sharpness-aware minimisation through gradient-norm constraints designed to find flatter minima more efficiently and improve generalisation. In reinforcement learning,~\cite{lee2025flat} related the robustness of trained agents to flatter minima at the end of optimisation. Finally, flatness has also been explored as a necessary component for generalisation in grokking, where a phase of extreme memorisation is followed by generalisation~\citep{han2025flatness}. In that setting, the authors showed that preventing Relative Flatness minimisation during training inhibited the escape from the memorisation phase of grokking.
Taken together, there is a strong empirical and theoretical basis for favouring flat minima in deep learning.    

In contrast to the prevailing view that sharpness necessarily signals poor generalisation, we find that sharper solutions can emerge precisely when models generalise better and exhibit improved reliability-related properties in high-dimensional setting. Additionally, we find in synthetic tasks that models can be sharp or flat and still perfectly generalise which can be characterised through decision boundary geometry. We propose that sharpness should be interpreted through the complexity of the learned function and the model's inductive bias, rather than through a universal preference for flatness.

\section{Sharpness, Generalisation and Reliability-Related Evaluations}
In this section we define the sharpness metrics used in our empirical experiments (Section~\ref{subsec:sharpness_metrics}), and the reliability-related metrics used to assess model behaviour beyond test accuracy (Section~\ref{subsec:saftey_metrics}). Our aim is to examine whether minima geometry aligns not only with generalisation, but also with broader properties relevant to reliable deployment, we leverage a range of flatness measures to ensure (where possible) that our insights are not over index on one particular geometric formalisation.

\subsection{Sharpness Metrics}
\label{subsec:sharpness_metrics}
We employ three established measures of sharpness from the literature, namely Fisher-Rao norm \citep{pmlr-v89-liang19a}, Relative Flatness \citep{petzka2021relative}, and average-case SAM-Sharpness \citep{foret2021sharpnessaware, mason-williams2024explicit}. Hessian-based metrics, such as the top eigenvalue or the trace of the Hessian, were shown not to be reparameterisation invariant as they can be manipulated through linear reparameterisations. These reparameterisations do not change the function of the model but can make a minimum appear arbitrarily sharper~\cite{dinh2017sharp}, thereby undermining any direct relationship between such measures and generalisation.
For this reason, we focus in particular on Fisher-Rao norm \citep{pmlr-v89-liang19a} and Relative Flatness \citep{petzka2021relative}, both of which are reparametrisation invariant, meaning that they quantify sharpness without being impacted by the types of layer wise reparametrisations considered by~\cite{dinh2017sharp}. This ensures that our study, and the conclusions drawn from it, are robust to such layer wise reparametrisations which traditionally undermine flatness/sharpness metrics. 

Below we provide the formal definitions of the three sharpness metrics used throughout this work:
Fisher-Rao norm, SAM-Sharpness and Relative Flatness. Information Geometric Sharpness (IGS)~\citep{informationGeometricSharpness} is also a relevant candidate metric; however, we omitted it from this study because its computation is not feasible for the network sizes and data sets considered here. 
For Fisher-Rao and Relative Flatness we use the implementation provided by~\cite{petzka2021relative}\footnote{Code base for Fisher-Rao norm and Relative Flatness from~\cite{petzka2021relative}:~\url{https://github.com/kampmichael/RelativeFlatnessAndGeneralization/blob/main/CorrelationFlatnessGeneralization/measure_comparison.py}}.

\textit{Fisher-Rao:} Fisher-Rao norm~\citep{pmlr-v89-liang19a} uses information geometry to provide a norm-based measure of model complexity. In architectures satisfying the assumptions under which it is derived, Fisher-Rao norm also provides a reparametrisation-invariant measure of minima sharpness, as verified by~\cite{petzka2021relative}. In this paper, we use Fisher-Rao norm as an empirical geometry measure following~\cite{petzka2021relative}. In line with~\cite{petzka2021relative}, we use the analytical expression for cross-entropy loss from the appendix of \cite{pmlr-v89-liang19a}, reproduced in~\eqref{FRnorm}, as the empirical estimator throughout this work. To be calculated Fisher-Rao norm needs a differentiable parametric statistical model where a probability or a conditional probability distribution can be defined. This is the case for all the models examined with this metric in the paper. Though different activation functions may impact calculations, this metric still provides precursory insights into geometry for architectures such as the ViT and, as such, is appropriate for the architectures we use.

In our notation, $L$ denotes the number of hidden layers in the sense of~\cite{pmlr-v89-liang19a}, so that $L+1$ is the number of weight matrices in the predictor. For practical implementation, we compute this quantity by counting the number of Linear, Conv1d, Conv2d, Conv3d, and Embedding layers in each network, and then matching this count to the $L+1$ factor in~\eqref{FRnorm}. Under this convention, the corresponding counts are 21 for ResNet18, 18 for VGG19, and 26 for ViT. Additionally, $N$ denotes the number of data points and $L_{i}$ the loss at a specific batch of data in $N$. 

\begin{equation}
\label{FRnorm}
\mathrm{FR}_{\mathrm{norm}} = \sqrt{ (L+1)^2 \cdot \frac{1}{N} \sum_{i=1}^{N} \left( \frac{\partial \mathcal{L}_i}{\partial \theta} \cdot \theta \right) }
\end{equation}

\textit{SAM-Sharpness:} When using the SAM optimiser~\citep{foret2021sharpnessaware} the $\rho$ parameter determines the radius at which an adversarial weight-space perturbation is taken. A small $\rho$ value indicates a reduced radius from the existing weight space. To study the loss regions in the weight spaces around the minima at the end of training we define SAM-sharpness as the average loss difference across 100 perturbed parameter locations at distance 0.005$\rho$ from the original model,  
following~\cite{foret2021sharpnessaware,mason-williams2024explicit}. This provides a local measure of how rapidly the loss changes under small parameter perturbations. A sharp minima would therefore be expected to exhibit a larger increase in loss from the baseline minima under a new adversarial weight-space sampling, whereas for a flat minima the loss would be expected to remain approximately constant, broadly relating to the robustness of representation under perturbation. 
\\
\\
\begin{equation}
S(\theta) = \frac{1}{N} \sum_{n=1}^N 
\left| \frac{\mathcal{L}(\theta + \Delta \theta_n) - \mathcal{L}(\theta)}{\rho} \right|.
\end{equation}

\textit{Relative Flatness:}~\cite{petzka2021relative} define Relative Flatness as a reparameterisation-invariant sharpness measure, and show that it has one of the strongest correlations with small generalisation gaps among the measures they consider.
In our setting, Relative Flatness sharpness is computed between the feature extraction layer and the classifier, and is computationally expensive because it requires the trace of Hessian blocks over the output matrices. We use the formulation shown in~\eqref{eq:relative_flatness}. 
\begin{equation}
\kappa^{\phi}_{\mathrm{Trace}}(\mathbf{w})
\;:=\;
\sum_{s,s'=1}^{d}
\langle \mathbf{w}_{s}, \mathbf{w}_{s'} \rangle
\,\cdot\,
\mathrm{Trace}\!\left( H_{s,s'}(\mathbf{w}, \phi(S)) \right)
\label{eq:relative_flatness}
\end{equation}

where $\mathbf{w}_{s}$ denotes the $s$-th row of $\mathbf{w}$, $\langle \cdot, \cdot \rangle$ is the scalar product, and $ H_{s,s'}(\mathbf{w}, \phi(S))$ is the Hessian of
the empirical loss with respect to $\mathbf{w}_{s}$ and $\mathbf{w}_{s'}$, evaluated at $\phi(S)$ , where where $\phi$ represents the fixed feature map, i.e. the penultimate layer in the model, and $S$ is a sample set ~\citep{han2025flatness}. We interpret Relative Flatness as an empirical geometry measure that is informative about the learned solution.

In line with the function-centric perspective of this paper, we interpret these sharpness measures, particularly Fisher-Rao norm, as capturing aspects of the geometry of the learned function, rather than treating them as universal scalar proxies for generalisation.

\subsection{Reliability-Related Metrics}
\label{subsec:saftey_metrics}
Beyond the generalisation properties of a neural network, there are several additional quantities of interest that are relevant to reliable and safe deployment of neural networks in.
In this section, we provide the reliability-related metrics used in this study to examine how minima geometry relates to broader model behaviour. 

\textit{Calibration:} Calibration measures how well a model's predicted confidence aligns with its empirical likelihood of correctness. Typically, deep neural networks, including ResNets, have been shown to be systematically overconfident~\citep{guo2017calibration}. The practical impact of this is that outputs of deep neural networks that are uncalibrated can provide misleading confidence in their predictions. In this paper we measure calibration using Expected Calibration Error (ECE)~\citep{guo2017calibration}, where lower values indicate better calibration and higher trustworthiness. To compute Expected Calibration Error (ECE), we use the Lighting AI Pytorch Metrics implementation of Multiclass Calibration Error following~\cite{kumar2019verified}\footnote{Calibration error documentation from Lighting AI: \url{https://lightning.ai/docs/torchmetrics/stable/classification/calibration_error.html}\label{footnote2}}.

\begin{equation}
    \mathrm{ECE} = \sum_{i=1}^{B} b_i \, |p_i - c_i| 
\end{equation}

\noindent where $p_i$ denotes the accuracy in bin $i$, and $c_i$ the average confidence in that bin under uniform binning~\citep{kumar2019verified}\textsuperscript{\ref{footnote2}}.

\textit{Functional Consistency:} Functional diversity reflects how similar neural networks are in function or representation space~\citep{wang2024great, mason-williams2024knowledge, mason-williams2024neural}. Prior work has linked diversity in function space to improved ensemble performance~\citep{fort2019deep, lu2024sharpness}, while other
work argues that representation convergence can also benefit ensembles~\citep{wang2024great}. In this paper, we focus on the consistency of the learned functions across runs by measuring prediction disagreement on the test set, which captures how often two models disagree on their top-1 outputs.
Lower disagreement implies greater agreement on individual predictions given the same training data, and we interpret this as a form of functional consistency under training stochasticity. 

Concretely, we use \textbf{Prediction Disagreement}, defined in~\eqref{pred_dis} following \citet{fort2019deep}, where $f(x;\theta)$ denotes the top-1 predicted class for sample $x$ under parameters $\theta$. Lower Prediction Disagreement indicates that two models agree more frequently on their top-1 predictions which indicates increased functional consistency. 
\begin{equation}
\label{pred_dis}
\frac{1}{N} \sum_{n=1}^{N} \bigl[\, f(x_n;\theta_1) \neq f(x_n;\theta_2) \,\bigr].
\end{equation}

\textit{Corruption Robustness:} Robustness assesses how well a model performs under distribution shift or input perturbations, which is important for reliable deployment. We evaluate robustness on CIFAR10-C , CIFAR100-C and TinyImageNet-C~\citep{hendrycks2019benchmarking}, which include common corruptions such as impulse noise, JPEG compression, and contrast distortions. Performance is quantified via mean corruption accuracy, where higher values indicate greater robustness. These perturbations represent average-case corruption robustness, rather than worst-case adversarial robustness \citep{hendrycks2019benchmarking}.
We employ these data sets to study how geometric properties relate to corruption robustness.  
Each corruption is provided at five levels of severity. 
The metric we use is \textbf{Corruption Accuracy}, defined as the average accuracy of a classifier $f$ on the corrupted test set $\mathcal{D}_{\mathrm{corruption}}$ across all corruption types and severity levels~\citep{hendrycks2019benchmarking}.

\begin{equation}
    \text{Corruption Accuracy} = \frac{1}{C} \sum_{c=1}^C \frac{1}{S} \sum_{s=1}^S \frac{1}{N_{c,s}}\sum_{n=1}^{N_{c,s}} \mathbf{1}(f(x_n^{(c,s)};\theta) = y_n).
\end{equation}
\noindent where $C$ is number of corruption types, $S$ is the number of severity levels, $N_{c,s}$ the number of samples under corruption $c$ and severity $s$, $f(x_n;\theta)$ the top-1 prediction for input $x_n$ under parameters $\theta$, and $y_n$ the corresponding label.

\section{Single-Objective Optimisation}
\begin{wrapfigure}{R}{0.35\textwidth}
\vspace{-0.2cm}
{\centering
\includegraphics[width=\linewidth]{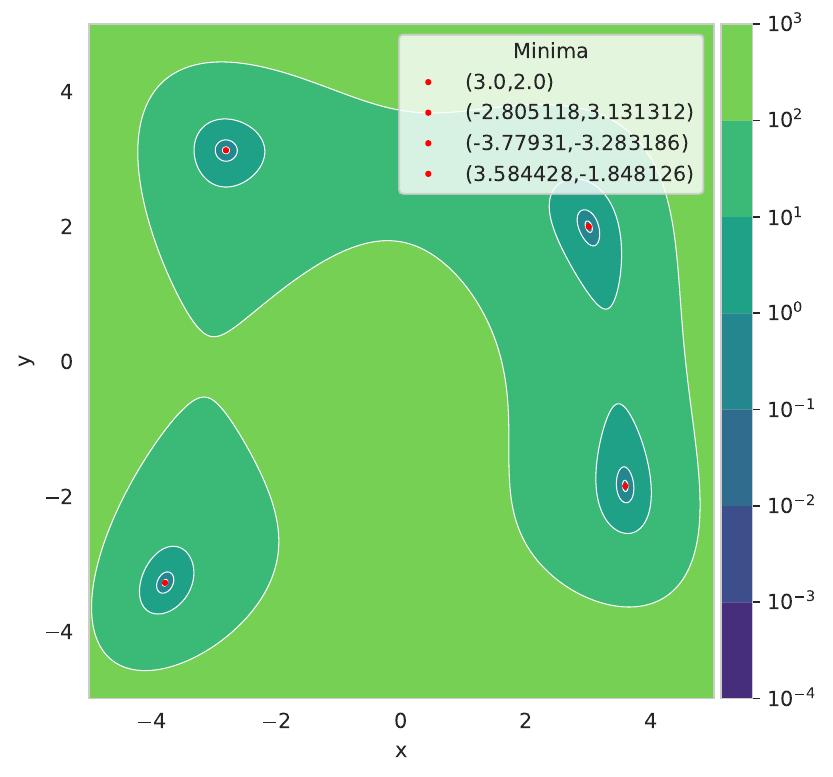}
    \caption{Himmelblau's function landscape.}
    \label{fig:himmelblaus}}
\end{wrapfigure}
\label{sec:single_objective_optimisation}
We posit that the sharpness reached by a model depends on the geometric properties of the function it is trained to approximate, given that neural networks are universal function approximators. To illustrate that loss-landscape geometry can be tied to solution complexity / the function being fitted, we begin with a toy setting: single-objective optimisation. 
Toy settings have previously been used to study geometric properties of neural networks; for example~\cite{pmlr-v137-huang20a} used the Swiss Roll data set to explore generalisation and flat minima. 

Consider Himmelblau's function in~\eqref{eq:himmelblauspaper} (visualised in Figure~\ref{fig:himmelblaus}). It has four global minima whose local geometry differs markedly (Table~\ref{tab:global-minima}), yet each achieves zero loss. Thus, no minimum is intrinsically preferable from an optimisation-objective standpoint. Under flatness-centric views, flatter minima would be deemed preferable; however, any network that represents the target function can plausibly reach any of these minima. Flatness is therefore not a necessary criterion for optimality in this setting.
\begin{equation}
    {\displaystyle f(\boldsymbol {x,y} )= (x^2 + y - 11)^2 + (x + y^2 - 7)^2}
    \label{eq:himmelblauspaper}
\end{equation}
Moving beyond this example, we examine a set of single-objective problems with a single global minimum. Figure~\ref{fig:single-objective-optimization} visualises six such functions (definitions are provided in Appendix~\ref{app:Single-Objective-Optimisation-Functions}). Each exhibits a distinct landscape, implying different local curvature at its global minimum. Table~\ref{tab:global_minima_sharpness} reports sharpness statistics at the global minimum. For instance, the Sphere function is the flattest across metrics, whereas functions with more intricate landscapes (e.g., Rosenbrock and Beale) have sharper optima. Accurately representing these objectives entails reaching minima with geometry commensurate to the target function.
\begin{table}[H]
    \centering
    \resizebox{\textwidth}{!}{
    \begin{tabular}{|l|c|c|c|c|}
    \hline
     \textbf{Global Minimum  }       &   \textbf{Condition Number} &   \textbf{Hessian Trace} &   \textbf{Hessian determinant} &   \textbf{Max Eigenvalue} \\
    \hline
     (3.0, 2.0)            &            3.200 &          108.000    &               2116.000    &          82.284 \\ \hline
     (-2.805118, 3.131312) &            1.242 &          145.39 &               5222.890 &          80.550   \\ \hline
     (-3.77931, -3.283186) &            1.892 &          204.500  &               9460.560 &         133.786  \\ \hline
     (3.584428, -1.848126) &            3.674 &          134.110 &               3024.540 &         105.419  \\
    \hline
    \end{tabular}
    }
    \caption{Local geometric properties at the four global minima of Himmelblau's function.}
    \label{tab:global-minima}
\end{table}

\begin{figure}[H]
    \centering
    \includegraphics[width=0.85\linewidth]{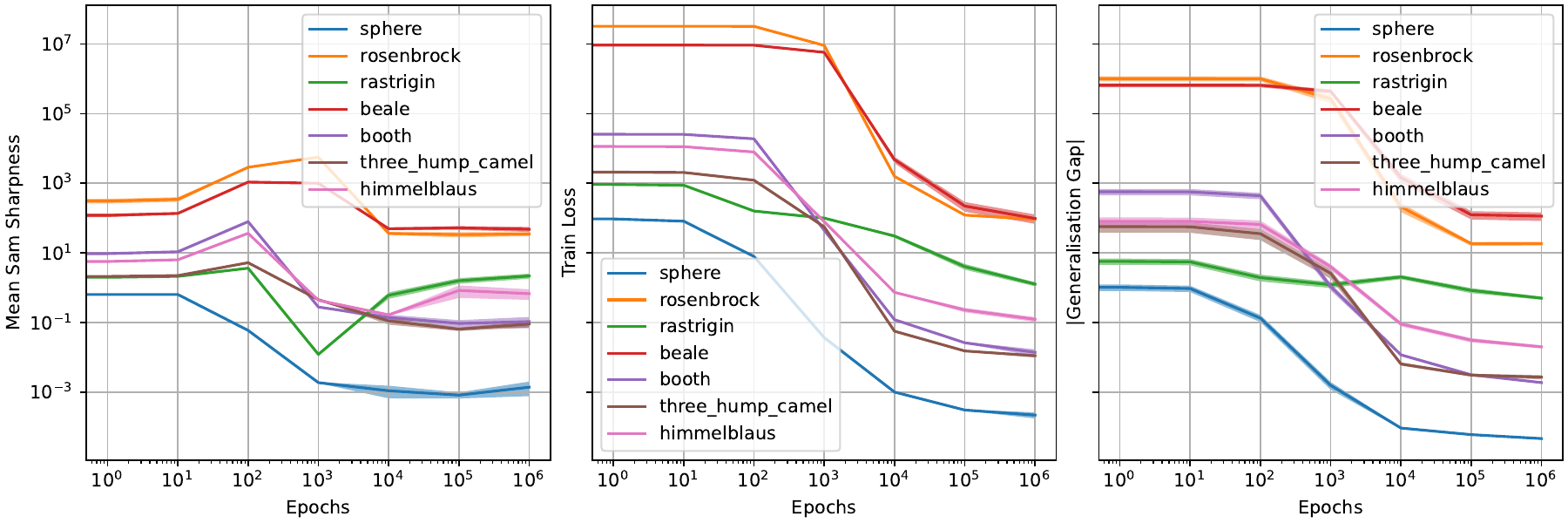}
     \caption{Training an MLP on single-objective problems over epochs: mean sharpness, training loss, and absolute generalisation gap (averaged over 10 runs).}
    \label{fig:sharpness_adam}
\end{figure}

We next fit an MLP to each objective using the same initialisation and average over ten models. As shown in Figure~\ref{fig:sharpness_adam}, the sharpness of the local minima encountered during training reflects the sharpness of the global optimum: under a fixed training budget, model sharpness, training loss, and generalisation gap vary systematically with the complexity/structure of the target function (see Figure~\ref{fig:single-objective-optimization}). Although absolute generalisation gaps differ across objectives, they exhibit similar relative reductions over training. Appendix~\ref{sec:equlivaent_loss} further shows that matching the final loss across functions still yields different sharpness levels, as expected from their intrinsic geometry and in Appendix~\ref{app:sam-sharpness-rho} we show that the SAM sharpness relationships seen for the single objective optimisation functions are robust to a sweep of the $\rho$ hyperparameter.

From this perspective, flatness is desirable only when demanded by the target function (e.g., Sphere, Booth and Three hump camel). Seeking flat solutions for intrinsically sharper objectives (e.g., Rosenbrock) may therefore be suboptimal: their structure is consistent with the need for tighter solution geometry and thus sharper minima. It is important to note that this section is intended only as an illustration of how neural network minima geometry can relate to function complexity and the function being fitted. In the regression setting considered here, this analysis does not extend directly to measures such as Relative Flatness, which require locally constant labels~\cite{petzka2021relative}. 
In the following section, we show how the same broader perspective extends to reparameterisation-invariant measures, including Fisher-Rao norm and Relative Flatness, in classification settings where the locally constant label condition holds. 
\begin{figure}[H]
    \centering
    \includegraphics[width=\linewidth]{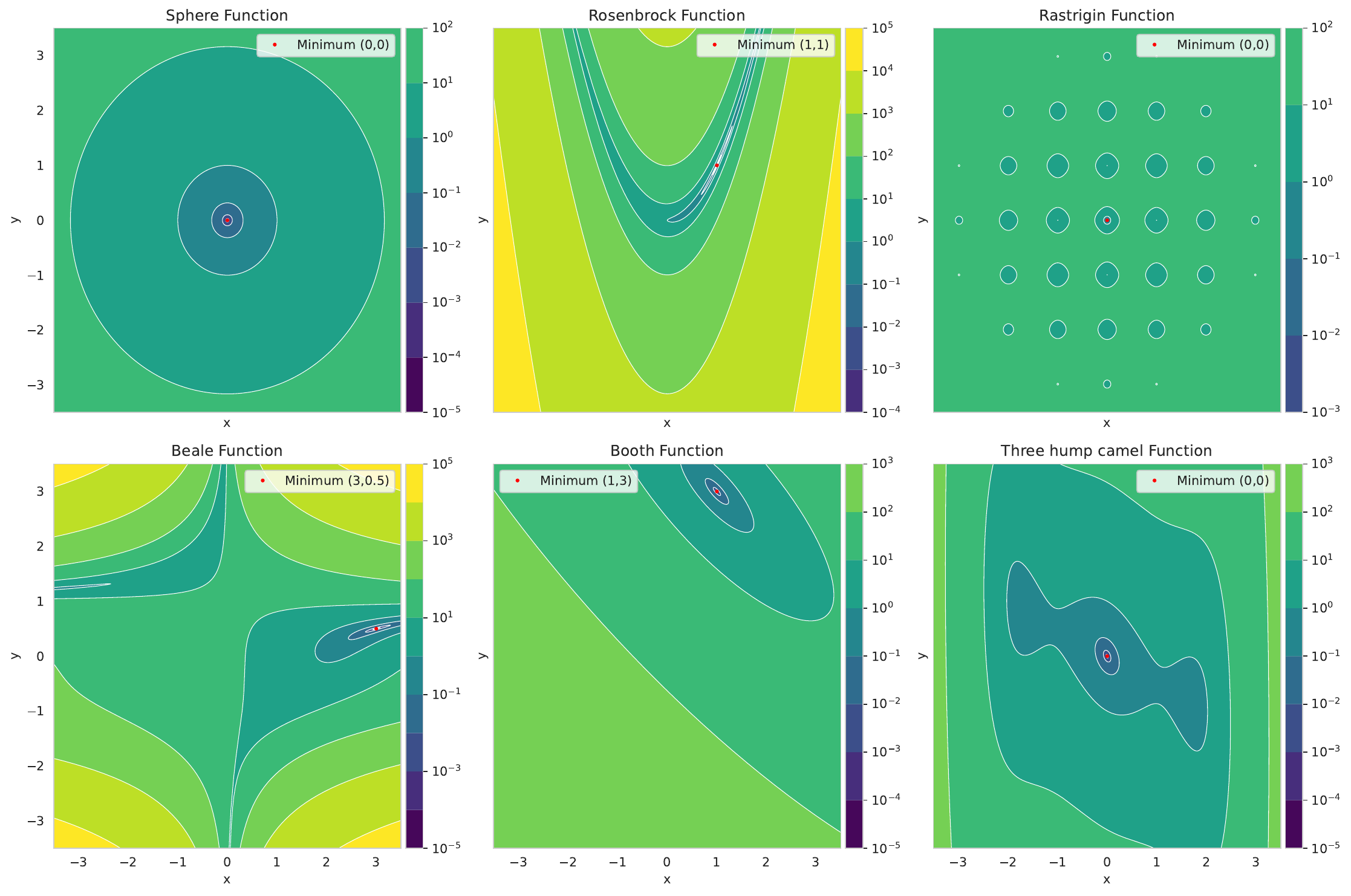}
     \caption{Landscapes for six single-objective functions. 
     }
    \label{fig:single-objective-optimization}
\end{figure}

\begin{table}[H]
    \centering
    \resizebox{\textwidth}{!}{
    \begin{tabular}{|l|c|c|c|c|}
    \hline
     \textbf{Function}         &   \textbf{Condition Number} &   \textbf{Hessian Trace} &   \textbf{Hessian determinant} &   \textbf{Max Eigenvalue} \\
    \hline
     Sphere           &            1.000       &          4.000      &                4.000      &          2.000       \\ \hline
     Rosenbrock       &         2508.010    &       1002.000      &              400.000      &       1001.600     \\ \hline
     Rastrigin        &            1.000       &        793.568  &           157438.000      &        396.784   \\ \hline
     Beale            &          162.473   &         49.281 &               14.766 &         48.980  \\ \hline
     Booth            &            9.000       &         20.000      &               36.000      &         18.000       \\ \hline
     Three hump camel &            2.784 &          6.000      &                7.000      &          4.414 \\ 
    \hline
    \end{tabular}
    }
    \caption{Sharpness at the global minimum for six single-objective optimisation functions.}
    \label{tab:global_minima_sharpness}
\end{table}

Furthermore, in Section~\ref{Sec:Increasing_func_complexity} we show that arbitrarily increasing the complexity of the training data in a classification setting results in models reaching sharper minima under Fisher-Rao norm and Relative Flatness. The findings from that experiment further support the relationship between function complexity and minima geometry illustrated in this toy setting.

\section{Decision Boundaries, Memorisation and Sharpness}
\label{sec:decision_boundaries}
In this section, we explore and extend the analysis conducted by~\cite{pmlr-v137-huang20a}, who studied sharp and flat minima through the lens of memorisation and generalisation.~\cite{pmlr-v137-huang20a} showed that neural networks that reach sharp minima, either through ``bad" minima or through training on memorised data, exhibit sharper landscape visualisations than models trained on standard data. As such, they introduced the perspective that the geometric properties of a model relate to the decision boundaries of solution, with tighter decision boundaries corresponding to models that have memorised training points. 

\begin{figure}[H]
    \centering    \includegraphics[width=0.3\linewidth]{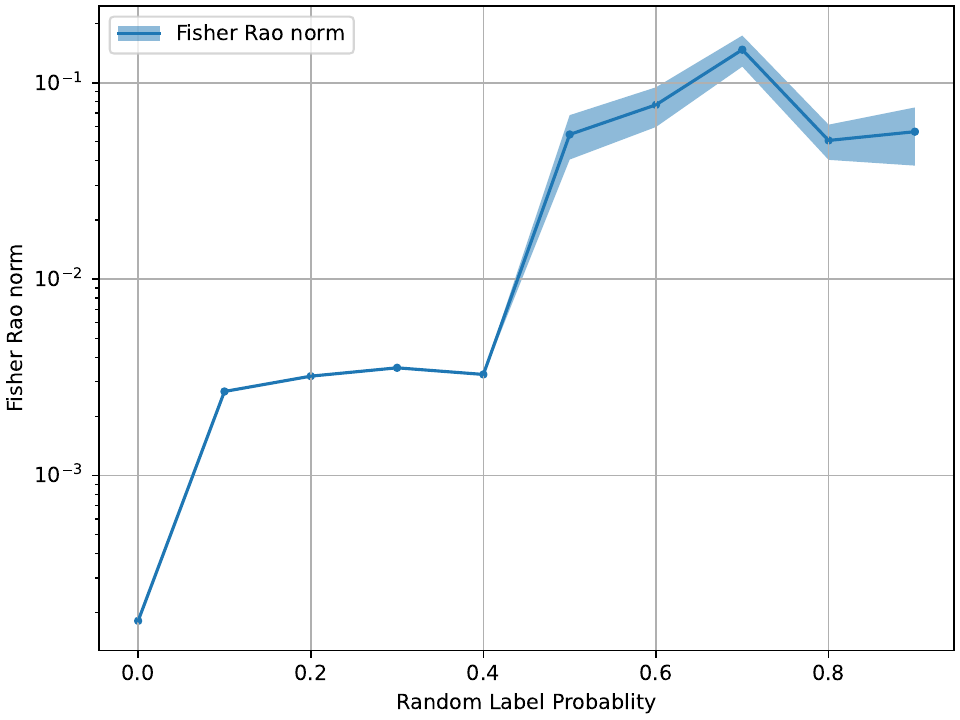}
    \includegraphics[width=0.3\linewidth]{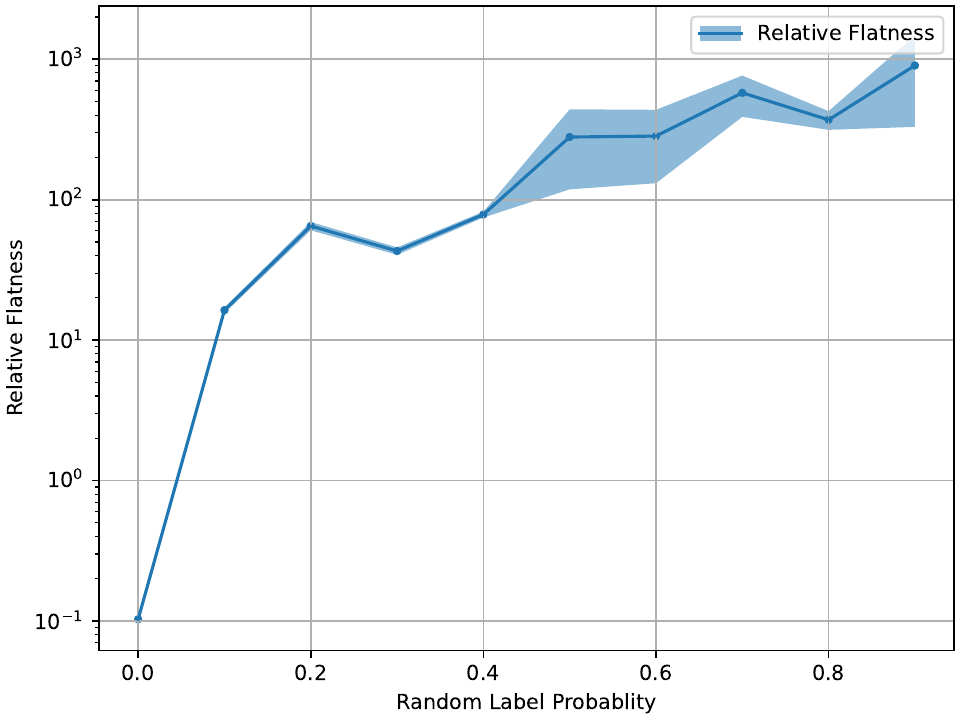}

     \caption{Fisher-Rao norm (Left) and Relative Flatness (Right) of the resulting minima after increasing the proportion of randomised data.}
    \label{fig:random_data_sharpness}
\end{figure}

We first reproduce the results of~\citet{pmlr-v137-huang20a} on the make\_circles data set, which is a non-linear binary classification task with 100 data points split evenly between 2 classes~\citep{scikit-learn}\footnote{Sklearn make\_circles data set: \url{https://scikit-learn.org/stable/modules/generated/sklearn.data sets.make_circles.html}}. We train a 3-layer MLP to classify the two classes. In the standard setting with no randomisation (random label probability 0.0), the model is trained with SGD using a learning rate of 0.01, momentum 0.9, batch size 5, for 300 epochs using cross-entropy loss. To show that increasing the randomisation of data points results in sharper minima, we scale the random label probability from 0.0 to 0.9. In the randomised setting, the MLP is trained for 20,000 epochs, with all other hyperparameters unchanged to allow convergence to 100\% accuracy on the training data set. For both the standard 0.0 and randomised conditions (0.1--0.9), we initialise 100 models across random seeds 0--99 and use the same training-data order for each model. The increased training budget / epochs required for the randomised data already suggests that such training data are harder for the MLP to fit.
\begin{figure}[H]
    \centering
    \includegraphics[width=\linewidth]{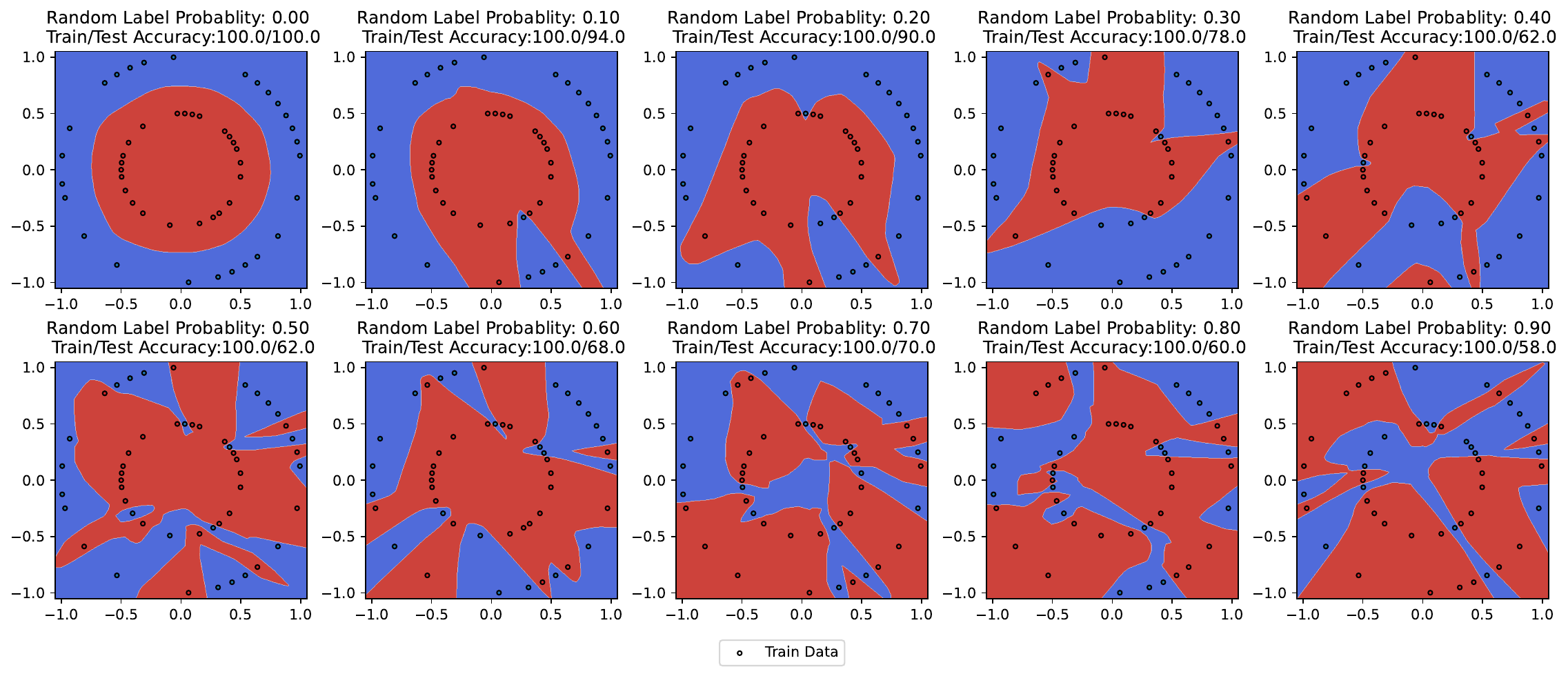}
    
     \caption{Visualising the decision boundaries of an MLP trained on the make\_circles data set. The proportion of randomised data is increased in 0.1 intervals between 0.0 and 0.9. \textbf{\textcolor{blue}{Blue}} indicates the decision boundary for class 1 and \textbf{\textcolor{red}{red}} indicates the decision boundary for class 2. As the data is increasingly randomised, the learned decision boundaries become more complex (irregular and tightly constrained).}
    \label{fig:random_data_boundaries}
\end{figure} 
Figure~\ref{fig:random_data_boundaries} shows the corresponding decision boundaries for seed 0, with the two class regions shown in \textcolor{red}{\textbf{red}} and \textcolor{blue}{\textbf{blue}}. As the decision boundaries become tighter and more irregular/complex, we also observe a general reduction in test accuracy, although the decrease is not strictly monotonic, likely due to stochasticity in the data-randomisation process. It can be observed that as the proportion of randomised data increases (0.1-0.9), the decision boundaries in the visualisation become tighter and more irregular. 
In Figure~\ref{fig:random_data_sharpness}, we show the mean sharpness of the minima found at the end of training (measured by Fisher-Rao norm and Relative Flatness)  together with $\pm{1}$ Standard Error of the Mean (SEM)~\citep{belia2005researchers}. 
Under both reparametrisation-invariant sharpness metrics, the models trained on randomised data exhibiting poor generalisation are substantially sharper than the baseline model trained without randomisation. Figure~\ref{fig:random_data_boundaries} also shows that, without randomisation, the decision boundaries are wide and the model generalises perfectly. This reproduces the central observation of ~\cite{pmlr-v137-huang20a} that memorisation coincides with sharp minima and tight decision boundaries, and therefore that increased minima sharpness indicates memorisation.

Building on our results in Single-Objective Optimisation discussed in Section~\ref{sec:single_objective_optimisation}, we now adopt a broader function-centric view. Rather than associating minima geometry only with memorisation or generalisation, we consider it to be tied more fundamentally to the function being learned and to the way in which the training data are fitted. To investigate this hypothesis under experimental conditions similar to those studied by~\cite{pmlr-v137-huang20a}, we modify the training data not by randomising labels, but by reducing the distance between the two classes while preserving perfect generalisation. 
Specifically, we vary the scale factor between the inner and outer circles, while controlling the experiment using the same 100 initialisations and data-order seeds.
A higher scale factor produces tighter class boundaries, while a lower scale factor leads to wider separation between classes. Thus, the scale factor directly controls the margin of the decision boundary.

\begin{figure}[H]
    \centering
    \includegraphics[width=\linewidth]{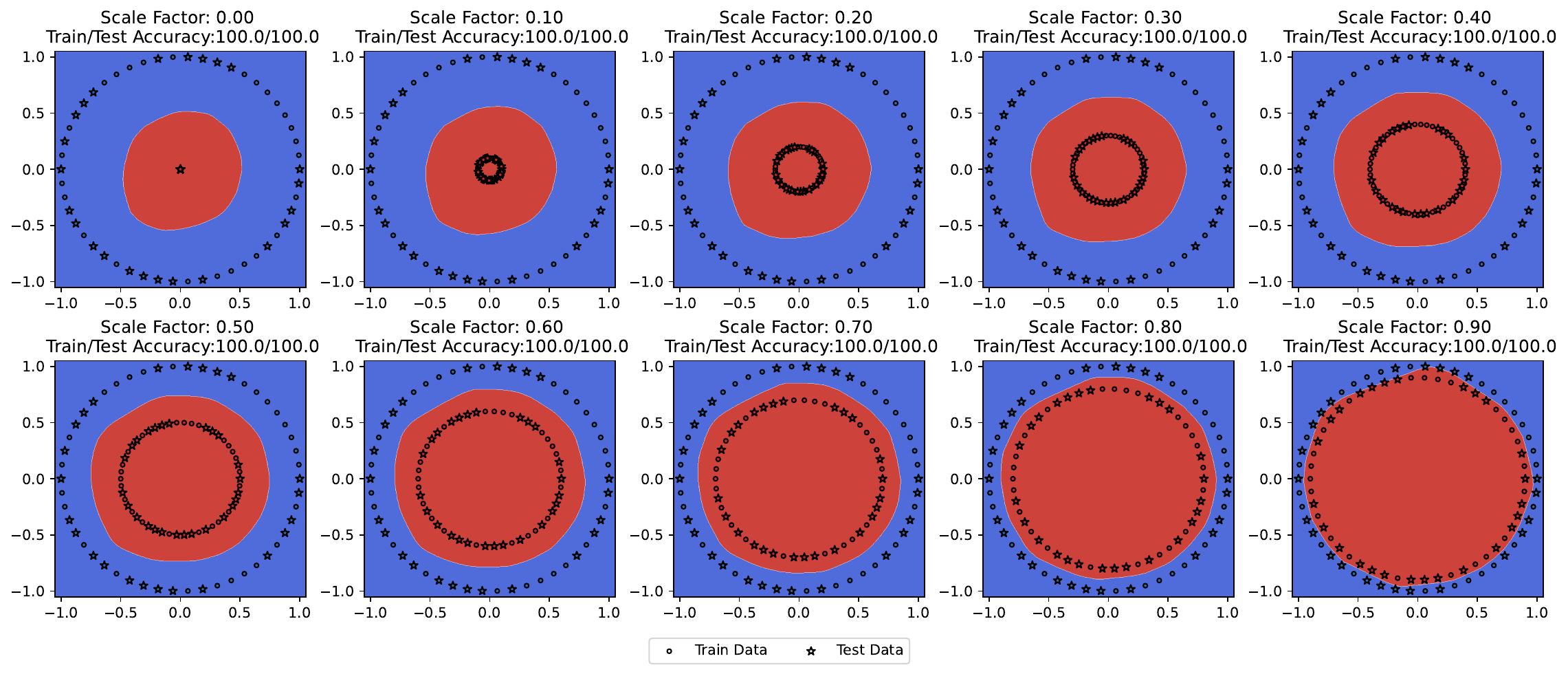}
     \caption{Visualising the decision boundaries of an MLP trained on the make\_circles data set. The scale factor is increased in 0.1 intervals between 0.0 and 0.9. \textbf{\textcolor{blue}{Blue}} indicates the decision boundary for class 1 and \textbf{\textcolor{red}{red}} indicates the decision boundary for class 2. As the Scale factor increases, the decision boundaries between class 1 and 2 become progressively tighter.}
    \label{fig:scale_factor_boundaries}
\end{figure}

\begin{figure}[H]
    \centering    \includegraphics[width=0.3\linewidth]{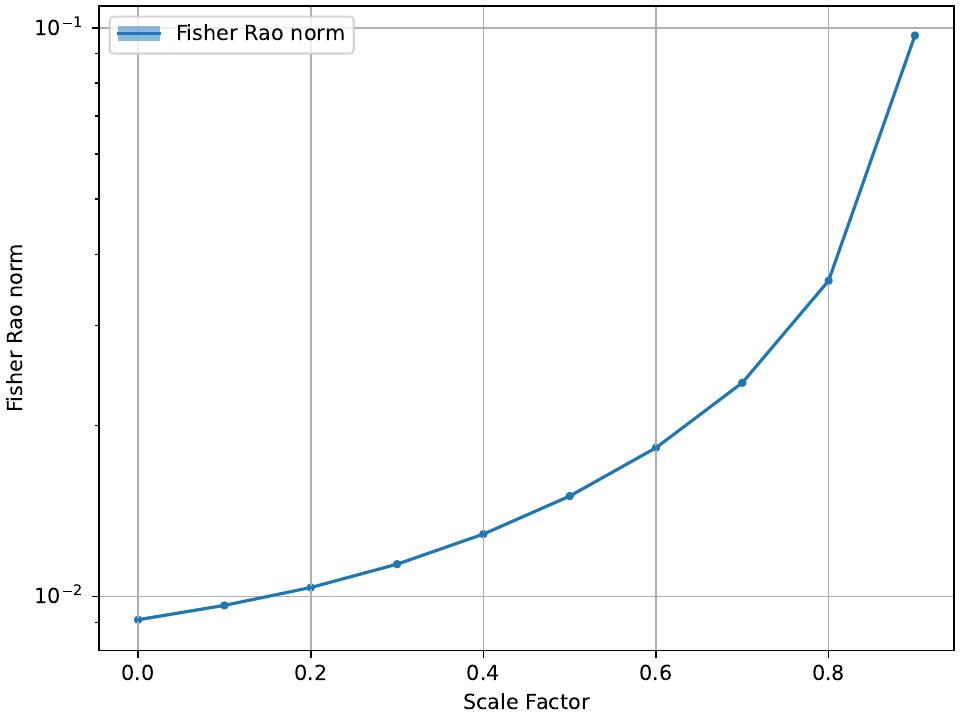}
    \includegraphics[width=0.3\linewidth]{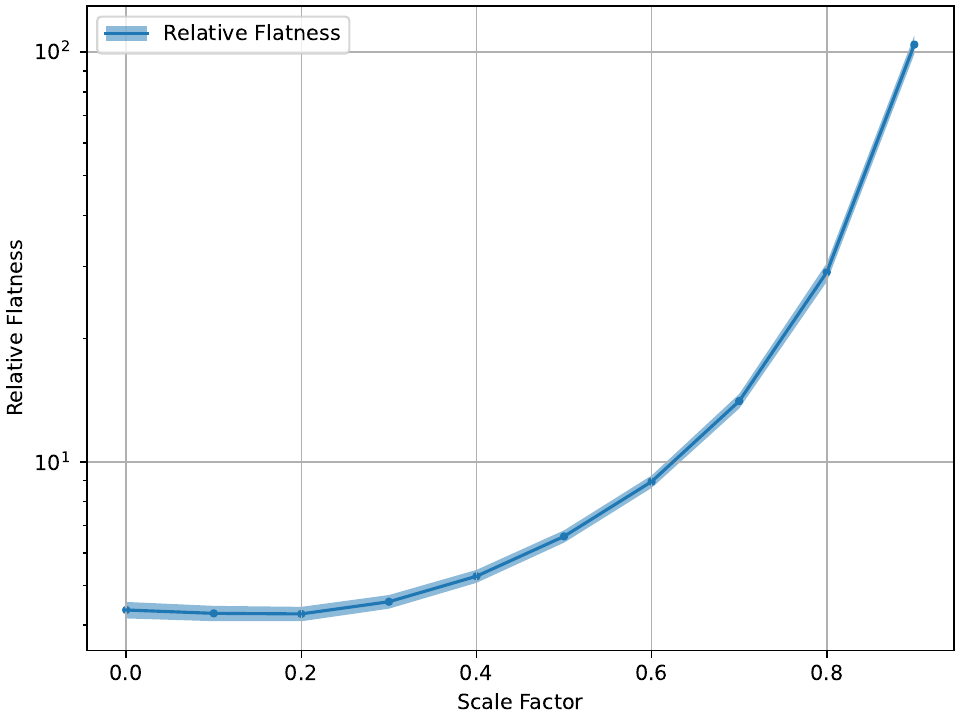}

     \caption{Fisher-Rao norm (Left) and Relative Flatness (Right) of the resulting minima after increasing the Scale factor of the data.}
    \label{fig:scale_factor_sharpness}
\end{figure}

To probe this relationship, we train the same MLP architecture across scale factors from 0.0 to 0.9 using the training hyperparameters as described in the standard experiment above. Our objective is to determine how increasing the tightness of the decision boundary affects the recorded sharpness of the minima at the end of training, outside the memorisation regime. At every scale factor, all models perfectly fit the training data and generalise perfectly to the test data. In Figure~\ref{fig:scale_factor_boundaries}, we observe that as the scale factor increases, the decision boundaries of the two classes move closer together. This is expected, since the scale factor explicitly reduces the separation between the two classes shown in \textcolor{red}{\textbf{red}} and \textcolor{blue}{\textbf{blue}}. Under existing interpretations of flat and sharp minima, one might expect all of these models to be found at relatively flat minima, given that they all achieve 100\% test accuracy and therefore generalise perfectly regardless of the scale factor. However, Figure~\ref{fig:scale_factor_sharpness} shows that increasing the scale factor is accompanied by increased sharpness under both Fisher-Rao norm and Relative Flatness. As the scale factor explicitly modifies the decision boundaries while preserving perfect generalisation, this suggests that sharpness can provide an indication of tighter decision boundaries rather than memorisation alone. In this way, we can partially decouple the relationship between sharp minima and memorisation, and instead interpret increased sharpness, more appropriately, as reflecting tighter and more complex decision-boundary structure in the learned function.

It is also interesting to compare the sharpness values in the randomised-data and the scale-factor settings (Figures~\ref{fig:random_data_sharpness} and~\ref{fig:scale_factor_sharpness}). 
The randomised models require a substantially larger training budget to fit the training data and reach sharpness values, particularly for Relative Flatness, that are orders of magnitude larger than those in the scale-factor setting. These results suggest that functions with extremely tight or irregular decision boundaries are harder for neural nets to fit than functions with wider or more structured decision boundaries, even when both are associated with increased sharpness.  

Our findings in this section therefore support a more nuanced relationship between sharp minima, memorisation, and decision-boundary structure:

\noindent\textit{While models that memorise training data may occupy sharp minima with tight decision boundaries, not all sharp minima with tight decision boundaries imply memorisation. } 

More broadly, these results strengthen the function-centric perspective developed throughout this paper: the manner in which data are learned can shape the geometry of the solution and, in doing so, reflect the complexity of the learned function rather than poor generalisation or memorisation alone.

\section{High-Dimensional Optimisation Problems} 
Building on the view developed in the previous sections, namely that minima geometry reflects properties of the function being fit and the way in which that function is learned, we now extend our analysis to high-dimensional settings in the vision domain. 
In practice, deep vision neural networks are routinely trained with regularisation~\citep{Goodfellow-et-al-2016, kukacka2017regularization}, yet why specific regularisers improve generalisation remains only partially understood despite extensive prior work~\citep{regularisationSurvey1,regularisationSurvey2,regularisationSurvey3}. This makes vision an ideal test-bed to study how geometry relates not only to generalisation, but also to reliability properties (calibration, robustness, and prediction agreement) at scale. Our contribution in this section is to examine these phenomena through the lens of learned solution (function-space) complexity, explicitly associating minima geometry with both generalisation and broader model behaviour. This function-centric perspective offers a complementary reading of flat and sharp minima.

\textit{Function Complexity:} 
Occam's Razor, or the principle of parsimony, states that when two competing hypotheses, $\mathcal{H}$ and $\mathcal{H'}$  both adequately explain an event, preference should be given to the hypothesis that relies on fewer assumptions. Formally, if $\mathcal{H}$ is characterised by an assumption set $\{\mathcal{A}_1,\mathcal{A}_2,\dots,\mathcal{A}_{\mathcal{K}}\}$ and $\mathcal{H}'$ by an assumption set $\{\mathcal{A}_1,\mathcal{A}_2,\dots,\mathcal{A}_{\mathcal{J}}\}$, with $\mathcal{K}<\mathcal{J}$, then Occam's Razor favours $\mathcal{H}$ over $\mathcal{H}'$~\citep{good1977explicativity}. 
This principle is commonly invoked in the flatness literature to motivate a preference for simpler learned solutions. In neural networks, this idea has often been linked to Minimum Description Length (MDL), where flatter minima are interpreted as requiring less precision to describe because the loss remains approximately constant under perturbation~\cite{hochreiter1994simplifying}. Here we posit that the minima geometry reached at the end of training can offer insights into how a model's effective capacity is leveraged to fit specific properties of the data and training dynamics. An extension of this being that model capacity measures alone may not be sufficient to make assumptions of a geometric bias of a particular model towards a flat or sharp minima. As a result, mathematically formalising such interactions may become important in the future for understanding geometric priors and robustness of neural networks.

From this perspective, sharper minima may be viewed as corresponding to learned solutions that are more tightly constrained, require more precise representation, or induce more complex decision-boundary structure. 
Through this lens, we study how sharpness (function complexity) is governed by regularisation to better understand generalisation and training dynamics. We extend this by connecting this study to reliability-related evaluations (calibration, robustness and prediction consistency) as well as optimisation dynamics. Despite the arguments of~\cite{dinh2017sharp}, flat minima continue to be widely regarded as important for strong generalisation~\citep{petzka2021relative,cha2021swad,zhao2022penalizing,lee2025flat,han2025flatness}. However, the relationship between minima geometry, reliability-related metrics, and function complexity remains underexplored. Existing perspectives in the flatness literature suggest that neural networks with small generalisation gaps and, by extension, strong calibration or robustness, should be found at flatter minima. Our single-objective and non-linear binary classification analyses suggest a more nuanced picture: regularisation can yield sharper minima when the learned function is more structured, more tightly constrained, or requires tighter decision-boundary geometry.  
We therefore examine, in a controlled manner, how commonly used regularisers affect sharpness and the corresponding reliability evaluations across matched seeds.

\subsection{Training Controls}

We train neural networks (ResNets, VGGs and ViTs) across a range of data sets (CIFAR10, CIFAR100 and TinyImageNet) using $n=10$ matched seeds. For each seed, all controls share the same initial weights and data order. This ensures that models trained under different controls begin from the same point in the loss landscape and, in principle, could traverse toward (and even reach) similar minima, enabling controlled geometric comparisons. Each control is applied independently, while all other training details (optimiser, learning-rate schedule, epochs, etc.) are held fixed within each architecture--data set setting, isolating the effect of each control on the resulting solution geometry.

Our objective is not to optimise for state-of-the-art performance, but to characterise, under controlled conditions, how different training controls affect minima geometry and reliability-related behaviour. We define the controls as follows: 
\\
\\
\textit{Baseline}: A model trained in a vanilla setting that has no extra regularisation terms applied -- for each architecture and data set, we define how the baseline model is created in Appendix Section~\ref{app:settings}. For each seed, the baseline provides a reference point for the geometric, generalisation, and reliability-related evaluations that should be expected for each architecture and dataset.
\\
\\
\textit{Weight Decay, Augmentation and Sharpness Aware Minimisation (SAM):} We consider weight decay ($5\times 10^{-4}$), data augmentation (random rotation and crop), and Sharpness Aware Minimisation (SAM), applied individually or in combination, as defined in Subsection~\ref{sec:experimental_setup}, as the regularisation controls used in this study. We record their effect on the sharpness metrics and the reliability evaluations under the matched-seed setup, allowing a controlled and precise causal insight into the exact impact of each regularisation control on the resulting geometric landscapes for each architecture--dataset pair.  

\subsection{Experimental Setup}
\label{sec:experimental_setup}
More formally, given a training control (regulariser) $\mathcal{C}$, we examine how it affects sharpness, and the corresponding generalisation and reliability evaluations. 
Let the set of controls (training conditions) be
$\mathcal{C}=\{\text{Baseline},\ \text{Baseline+SAM},\ \text{Augmentation},\ \text{Augmentation+SAM},\ \text{Weight Decay},\ \text{Weight Decay+SAM}\}$.
Let $\mathcal{M}=\{\text{SS},\text{FR},\text{RF}\}$ denote sharpness metrics
(SAM sharpness, Fisher-Rao, Relative Flatness), $\mathcal{G}$ denote generalisation related metrics (generalisation gap, and test accuracy), $\mathcal{G}=\{\text{Gen}_{\text{gap}},\ \text{Test}_{\text{acc}}\}$
and let $\mathcal{R}$ denote reliability-related metrics (calibration, corruption-robust accuracy and prediction disagreement) where $\mathcal{R}=\{\text{ECE}, \text{Acc}_{\text{corr}}, \text{Disagree}\}$. We run seeds $i\in\{0,\dots,9\}$ with identical initialisation and data order across controls. For each control $c\in\mathcal{C}$ and seed $i$, we record
$
S^{(c)}_{i,m}\, (m\in\mathcal{M}),\,\,
G^{(c)}_{i,g}\, (g\in\mathcal{G}),\,\,
R^{(c)}_{i,r}\, (r\in\mathcal{R}).
$
We report per-control, per-metric summaries as means across seeds:
\[
\bar S^{(c)}_{m}=\frac{1}{n}\sum_{i=0}^{n-1} S^{(c)}_{i,m},\qquad
\bar G^{(c)}_{g}=\frac{1}{n}\sum_{i=0}^{n-1} G^{(c)}_{i,g},\qquad
\bar R^{(c)}_{r}=\frac{1}{n}\sum_{i=0}^{n-1} R^{(c)}_{i,r},
\]
and present mean $\pm$ SEM~\cite{belia2005researchers} across ten seeds for each control. 
\\
\\
Our hypothesis for the high-dimensional experiments is as follows:\begin{itemize}
\setlength{\parskip}{0pt}
\setlength{\itemsep}{0pt}
\item[$H_0$:] Regularisation controls ($\mathcal{C}$) reduce sharpness ($\mathcal{M}$) relative to the baseline, i.e., $\bar S^{(c)}_{m}$ $\ll$ $\bar S^{(baseline)}_{m}$, while improving evaluations in $\mathcal{G}$ and $\mathcal{R}$, i.e., reduced generalisation gap, higher accuracy-based metrics, and lower ECE and disagreement. 

\item[$H_a$:] Regularisation controls ($\mathcal{C}$) increase sharpness ($\mathcal{M}$) relative to the baseline, i.e., $\bar S^{(c)}_{m}$ $\gg$ $\bar S^{(baseline)}_{m}$, while still improving evaluations in $\mathcal{G}$ and $\mathcal{R}$, i.e., reduced generalisation gap, higher accuracy metrics, and lower ECE and disagreement. 
\end{itemize}

This hypothesis states the experiment-level expectation in terms of per-control average behaviour across seeds. The actual statistical tests are conducted on seed-matched paired differences between each control and the corresponding baseline run. Specifically, for each architecture--data set pair and metric, we use one-sided Wilcoxon signed-rank tests~\citep{woolson2007wilcoxon} on the paired seed-level differences to test whether a control improves performance in the preferred direction relative to the baseline. 

For each architecture--data set pair and metric, we compare each control to the \textit{Baseline} model across the 10 matched runs ($\mathcal{C}=\{\textit{Baseline+SAM},\ \textit{Augmentation},\ \textit{Augmentation+SAM},\ \textit{Weight Decay},\ $ $\textit{Weight Decay+SAM}\}$) using one-sided Wilcoxon signed-rank tests~\citep{woolson2007wilcoxon}. Our scientific question is directional: for sharpness metrics in $\mathcal{M}$, the alternative tests whether a regularisation control yields sharper minima than the baseline; for $\mathcal{G}$, this means lower values for $\text{Gen}_{\text{gap}}$ and higher $\text{Test}_{\text{acc}}$; for $\mathcal{R}$, this corresponds to higher values for $\text{Acc}_{\text{corr}}$ and lower values for ECE and $\text{Disagree}$. 

To control the expected false discovery rate across the reliability and sharpness metrics for each control, we apply Benjamini--Hochberg False Discovery Rate (FDR) correction ~\citep{yekutieli1999resampling} and report significance after FDR for $\mathcal{M}$ and $\mathcal{R}$. This means that FDR is conducted independently for \{ECE, $\text{Acc}_{\text{corr}}$, Disagree\} for $\mathcal{R}$ and for $\mathcal{M}$ \{\textit{SS},\textit{FR},\textit{RF}\}. For $\mathcal{G}$ $\{\text{Gen}_{\text{gap}},\ \text{Test}_{\text{acc}}\}$, we retain the significance threshold at $0.05$, since $\text{Gen}_{\text{gap}}$ is defined as $100\times(\text{Train}_{\text{acc}} - \text{Test}_{\text{acc}})$ and is therefore dependent on $\text{Test}_{\text{acc}}$.
We consider a regularisation control to be \textit{statistically supported} 
if the corrected significance threshold is satisfied for the majority of reliability measures against the baseline. Similarly, a regularisation control leads to \textit{statistically supported} increased sharpness if all sharpness values under correction reach the significance threshold against the baseline. In significance tables when there is a statistically significance result we report the absolute matched pairs rank-biserial correlation (RBC)~\citep{kerby2014simple} effect size. The absolute RBC values range from 0 to 1 where 0 indicates no meaningful effect and 1 indicates the largest possible effect. We follow this experimental setup to promote reproducibility of our findings and conclusions~\citep{mason-williams2025reproducibility}. 

\section{High Dimensional Results and Discussion}

Our central aim in this section is to test whether the common flatness narrative survives under controlled modern training settings. More specifically, we examine whether regularised models are consistently flatter than unregularised baselines, whether flatter models consistently achieve stronger downstream behaviour, and whether these relationships depend on the architecture and training control. It is important to note that despite the utility of our synthetic results for understanding minima properties and decoupling memorisation and generalisation, there is no guarantee of transfer across scales in highly non-convex optimisation settings, which warrants further exploration at higher dimensional settings.  The first two questions are addressed primarily in Section~\ref{sec:sharpness_regularisation}. The role of training control is examined further in Section~\ref{sec:rho_SAM_seep}, while Section~\ref{Sec:Increasing_func_complexity} provides additional support for our function-centric interpretation by showing that artificially increasing task complexity can also lead to sharper minima. Evidence across data set complexities (CIFAR10, CIFAR100 and TinyImageNet) is provided in the main body for ResNet18, and in Appendix Sections~\ref{app:VGG} and~\ref{app:ViT} we present results for VGG and ViT, respectively.

We present results for ResNet18 trained on CIFAR10, CIFAR100, and TinyImageNet. For each control in $\mathcal{C}$, we report generalisation metrics ($\mathcal{G}$) together with reliability-related evaluations ($\mathcal{R}$) and sharpness metrics ($\mathcal{M}$) across 10 matched seeds. Appendix~\ref{app:settings} provides the full training and sharpness-evaluation details. 
Results for VGG and ViT are given in Appendix~\ref{app:VGG} and~\ref{app:ViT}, where they support the same overall picture. 
Tables~\ref{tab:c10_resnet18},~\ref{tab:c100_resnet18},~\ref{tab:TiN_resnet18} below summarise how each training control affects sharpness and reliability evaluations. Means $\pm$ SEM are reported per metric. TinyImageNet results exclude Relative Flatness due to the Relative Flatness calculation exceeding the memory of an 80GB A100 GPU for any sample size of the training set. Additional batch-size (256 and 128) and learning-rate ($1e^{-3}$ and $1e^{-2}$) sweeps for ResNet and VGG are reported in Appendix~\ref{sec:ResNet_param_sweep} and~\ref{sec:vgg_param_sweep}, and support the same trends. 
In Section~\ref{sec:rho_SAM_seep} we further explore how increasing the SAM radius $\rho$ can increase a model's sharpness relative to the baseline while also improving performance. Additionally, in Section~\ref{Sec:Increasing_func_complexity}, we artificially increase the complexity of the training data and observe that minima become sharper when classes have increasingly disjoint examples. Finally, in appendix Section~\ref{Language_model_results} we show that our findings extend to Language data domains, that regularisation can result in the best performing models across generalisation and reliability measures while increasing sharpness.
Taken together, these results suggest that function complexity and minima geometry are closely related in high-dimensional optimisation.

\subsection{Regularisation and Sharpness}
\label{sec:sharpness_regularisation}
\textit{Regularisers Improve Generalisation and Often Increase Sharpness.} Across all CIFAR data sets and architectures, we observe a recurrent trend: the baseline condition yields the flattest minima (lowest values across FR, RF, SAM), yet performs worst on test accuracy and reliability-related metrics: calibration (ECE), robustness (Corruption Accuracy), and functional consistency (Prediction Disagreement) (Tables~\ref{tab:c10_resnet18},~\ref{tab:c100_resnet18}). Conversely, controls with stronger regularisation tend to yield sharper solutions while also achieving stronger evaluations. For example, in Table~\ref{tab:c10_resnet18} for the conditions of augmentation and augmentation + SAM, the sharpness values are two orders of magnitude larger for Fisher-Rao norm and Relative Flatness and several orders of magnitude larger for SAM-Sharpness against the baseline, while test accuracy increases by more than 15\%. This calls into question the conventional view that flatter minima are inherently preferable, and instead supports the function-centric perspective that sharper minima can reflect more complex, well-generalising solutions. Crucially, we also find that sharper minima can coincide empirically with better reliability-related performance. 
Across the majority of regularisers, the increase in reliability performance against the baseline condition is statistically supported. We observe in Tables~\ref{tab:res-sign-c10},~\ref{tab:res-sign-c100} and ~\ref{tab:res-sign-TiN}, that regularisers often coincide with a statistically significant increase in sharpness against the baselines. However, there are some instances in the TinyImageNet significance Table~\ref{tab:res-sign-TiN} where the SAM and weight decay controls do not increase sharpness when applied independently, but when applied together, they result in a statistically significant increase in sharpness. As a result, we argue that, for the training conditions explored in this paper, regularisation of neural networks under equal training budgets often leads to models navigating to sharper minima, which may indicate tighter or more constrained learned solutions rather than poorer generalisation. 

\begin{table}[H]
\centering
\resizebox{\textwidth}{!}{
\begin{tabular}{|c|c|c|c|c|c|c|c|c|}
\hline
\textbf{\begin{tabular}[c]{@{}c@{}}Control\\ Condition\end{tabular}} & \textbf{\begin{tabular}[c]{@{}c@{}}Generalisation \\ Gap\end{tabular}} & \textbf{\begin{tabular}[c]{@{}c@{}}Test \\ Accuracy\end{tabular}} & \textbf{\begin{tabular}[c]{@{}c@{}}Test \\ ECE\end{tabular}} & \textbf{\begin{tabular}[c]{@{}c@{}}Corruption \\ Accuracy\end{tabular}} & \textbf{\begin{tabular}[c]{@{}c@{}}Prediction \\ Disagreement\end{tabular}} & \textbf{\begin{tabular}[c]{@{}c@{}}SAM \\ Sharpness\end{tabular}} & \textbf{\begin{tabular}[c]{@{}c@{}}Fisher-Rao \\ norm\end{tabular}} & \textbf{\begin{tabular}[c]{@{}c@{}}Relative \\ Flatness\end{tabular}} \\ \hline
Baseline                                                             & 28.050 $\pm{0.175}$                                                         & 0.720 $\pm{0.002}$                                                     & 0.186 $\pm{0.001}$                                                & 58.614 $\pm{0.201}$                                                          & 0.282 $\pm{0.001}$                                                               & 1.366E-05 $\pm{1.206E-06}$                                             & 0.032 $\pm{0.001}$                                                       & 34.607 $\pm{0.757}$                                                        \\ \hline
\begin{tabular}[c]{@{}c@{}}Baseline\\ + SAM\end{tabular}             & 20.588 $\pm{0.125}$                                                         & 0.794 $\pm{0.001}$                                                     & 0.108 $\pm{0.001}$                                                & 66.342 $\pm{0.164}$                                                          & 0.168 $\pm{0.000}$                                                               & 5.823E-05 $\pm{9.056E-06}$                                             & 0.107 $\pm{0.006}$                                                       & 75.093 $\pm{1.693}$                                                        \\ \hline
Augmentation                                                         & 10.399 $\pm{0.067}$                                                         & 0.886 $\pm{0.001}$                                                     & 0.077 $\pm{0.001}$                                                & 68.755 $\pm{0.219}$                                                          & 0.121 $\pm{0.001}$                                                               & 1.905E-01 $\pm{2.203E-02}$                                             & 3.940 $\pm{0.207}$                                                       & 2903.220 $\pm{89.243}$                                                     \\ \hline
\begin{tabular}[c]{@{}c@{}}Augmentation\\ + SAM\end{tabular}         & \textbf{6.864 $\pm{0.038}$}                                                 & \textbf{0.908 $\pm{0.000}$}                                            & \textbf{0.014 $\pm{0.001}$}                                       & \textbf{71.419 $\pm{0.283}$}                                                 & \textbf{0.069 $\pm{0.000}$}                                                      & 1.303E-01 $\pm{1.547E-02}$                                             & 5.571 $\pm{0.035}$                                                       & 4970.972 $\pm{30.139}$                                                     \\ \hline
Weight Decay                                                         & 27.942 $\pm{0.196}$                                                         & 0.721 $\pm{0.002}$                                                     & 0.174 $\pm{0.002}$                                                & 58.562 $\pm{0.227}$                                                          & 0.281 $\pm{0.001}$                                                               & 3.391E-05 $\pm{4.494E-06}$                                             & 0.065 $\pm{0.004}$                                                       & 59.767 $\pm{3.009}$                                                        \\ \hline
\begin{tabular}[c]{@{}c@{}}Weight Decay\\ + SAM\end{tabular}                                                   & 19.788 $\pm{0.149}$                                                         & 0.802 $\pm{0.001}$                                                     & 0.096 $\pm{0.001}$                                                & 67.079 $\pm{0.117}$                                                          & 0.162 $\pm{0.001}$                                                               & 8.733E-05 $\pm{1.430E-05}$                                             & 0.127 $\pm{0.006}$                                                       & 88.807 $\pm{2.336}$                                                        \\ \hline
\end{tabular}}
\caption{Results for ResNet18 trained on CIFAR10. \textbf{Bolded} values indicate the best performance per metric. For sharpness metrics, lower values correspond to flatter models.}
\label{tab:c10_resnet18}
\end{table}

\begin{table}[H]
\resizebox{\textwidth}{!}{
\begin{tabular}{|c|c|c|c|c|c|c|c|c|}
\hline
\textbf{\begin{tabular}[c]{@{}c@{}}Control\\ Condition\end{tabular}} & \textbf{\begin{tabular}[c]{@{}c@{}}Generalisation \\ Gap\end{tabular}} & \textbf{\begin{tabular}[c]{@{}c@{}}Test \\ Accuracy\end{tabular}} & \textbf{\begin{tabular}[c]{@{}c@{}}Test \\ ECE\end{tabular}} & \textbf{\begin{tabular}[c]{@{}c@{}}Corruption \\ Accuracy\end{tabular}} & \textbf{\begin{tabular}[c]{@{}c@{}}Prediction \\ Disagreement\end{tabular}} & \textbf{\begin{tabular}[c]{@{}c@{}}SAM \\ Sharpness\end{tabular}} & \textbf{\begin{tabular}[c]{@{}c@{}}Fisher-Rao \\ norm\end{tabular}} & \textbf{\begin{tabular}[c]{@{}c@{}}Relative \\ Flatness\end{tabular}} \\ \hline
\begin{tabular}[c]{@{}c@{}}Baseline \\ + SAM\end{tabular}      & \cmark~(1.0) & \cmark~(1.0) & \cmark~(1.0) & \cmark~(1.0) & \cmark~(1.0) & \cmark~(1.0)  & \cmark~(1.0) & \cmark~(1.0) \\ \hline
Augmentation                                                   & \cmark~(1.0) & \cmark~(1.0) & \cmark~(1.0) & \cmark~(1.0) & \cmark~(1.0) & \cmark~(1.0)  & \cmark~(1.0) & \cmark~(1.0) \\ \hline
\begin{tabular}[c]{@{}c@{}}Augmentation  \\ + SAM\end{tabular} & \cmark~(1.0) & \cmark~(1.0) & \cmark~(1.0) & \cmark~(1.0) & \cmark~(1.0) & \cmark~(1.0)  & \cmark~(1.0) & \cmark~(1.0) \\ \hline
Weight Decay                                                   & \xmark~(N/A)  & \xmark~(N/A) & \cmark~(1.0) & \xmark~(N/A) & \xmark~(N/A)  & \cmark~(0.96) & \cmark~(1.0) & \cmark~(1.0) \\ \hline
\begin{tabular}[c]{@{}c@{}}Weight Decay \\ + SAM\end{tabular}  & \cmark~(1.0) & \cmark~(1.0) & \cmark~(1.0) & \cmark~(1.0) & \cmark~(1.0) & \cmark~(1.0)  & \cmark~(1.0) & \cmark~(1.0) \\ \hline
\end{tabular}}
\caption{Significance results for the control conditions against the baseline for ResNet18 trained on CIFAR10.~\cmark~indicates significant difference compared to the baseline; \xmark~indicates no significance, the value in the brackets represents the  effect sizes using the absolute RBC.}
\label{tab:res-sign-c10}
\end{table}

\textit{SAM Can Improve Performance While Increasing Sharpness}.
Contrary to claims that SAM necessarily finds flatter solutions~\citep{foret2021sharpnessaware,cha2021swad}, our results show that SAM often increases sharpness across metrics and conditions (Tables~\ref{tab:c10_resnet18},~\ref{tab:c100_resnet18} and~\ref{tab:TiN_resnet18}, as well as Appendix Sections~\ref{app:VGG} and~\ref{app:ViT}). Notably, augmentation + SAM achieves the best performance across evaluations while also being among the sharpest models. There are limited exceptions; for example, SAM-Sharpness decreases for augmentation + SAM on CIFAR10 and CIFAR100 (Tables~\ref{tab:c10_resnet18}, ~\ref{tab:c100_resnet18}), but this behaviour is not consistent across metrics. On more complex data sets (TinyImageNet; Table~\ref{tab:TiN_resnet18}), SAM can sometimes lead to flatter solutions, though again not uniformly across measures. Largely, when considering significance, we find that there is statistical support that SAM increases the sharpness of models in Tables~\ref{tab:res-sign-c10},~\ref{tab:res-sign-c100} and~\ref{tab:res-sign-TiN}, with a few exceptions. This finding is robust across hyperparameter settings and architectures.

Overall, these findings suggest that SAM can support the learning of higher-performing functions that reside in sharper regions of the loss landscape. In Section~\ref{sec:rho_SAM_seep}, we further show that modifying the SAM hyperparameter, $\rho$, which controls the radius of the exploration space, directly alters  sharpness of the minima found at the end of training. The results again show that the strongest-performing models can occupy sharper minima than the baseline as the perturbation radius grows, although there is a critical perturbation-radius threshold at which performance begins to deteriorate.

\textit{Reconciling SAM's Objective with Increased Sharpness.} While SAM is commonly understood as a flatness-promoting optimisation~\citep{foret2021sharpnessaware}, its objective encourages local robustness rather than global flatness. Specifically, SAM minimises the loss at the worst-case perturbation within a small neighbourhood around the current weights, thereby promoting low curvature in that vicinity. However, this does not guarantee low values under all global or reparameterisation-invariant sharpness metrics. Our findings -- where SAM often increases Fisher-Rao norm, Relative Flatness, and SAM-sharpness -- highlight that sharper solutions can still emerge, especially when the model learns more complex (more structured or tightly constrained) functions. This suggests that SAM may improve generalisation and reliability metrics not solely by flattening the landscape, but by guiding the model toward solutions that are robust in important local directions, even if they remain globally sharp under broader measures.

\begin{table}[H]
\centering
\begin{tabular}{|c|c|c|c|}
\hline
                     & \textbf{Baseline} & \textbf{Weight Decay} & \textbf{Augmentation} \\ \hline
\textbf{\begin{tabular}[c]{@{}c@{}}Without \\ SAM\end{tabular}} &  \includegraphics[width=0.25\textwidth]{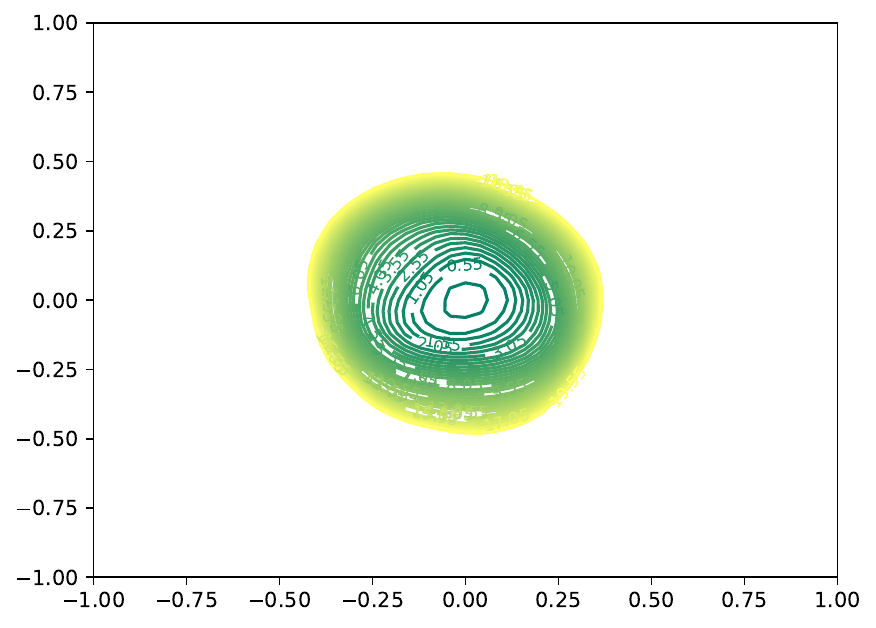}                  & \includegraphics[width=0.25\textwidth]{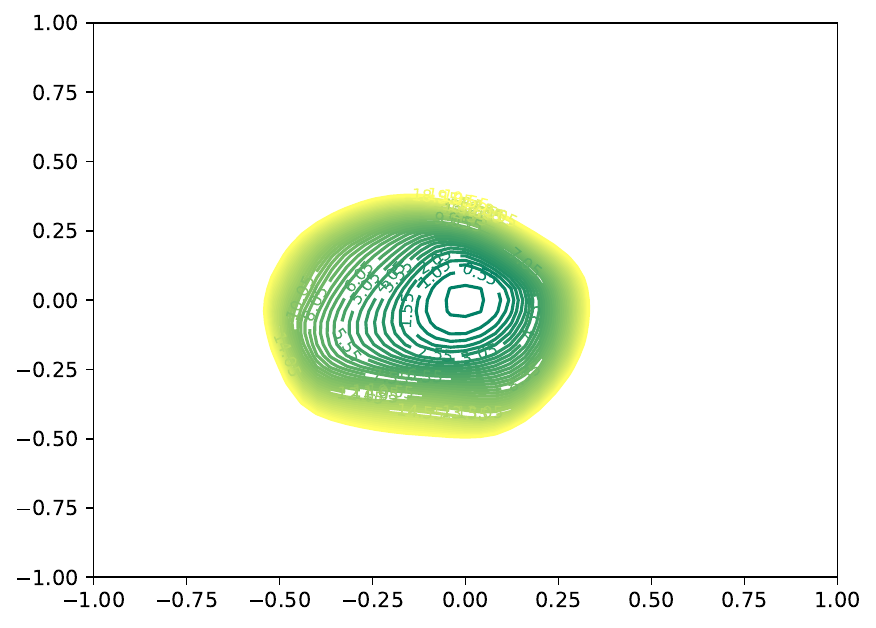}                          &  \includegraphics[width=0.25\textwidth]{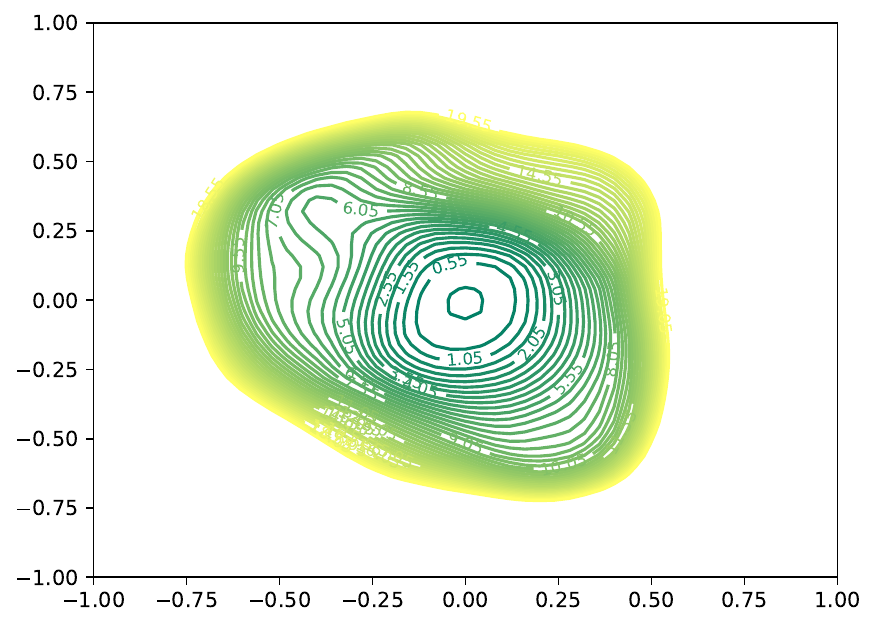}                       \\ \hline
\textbf{\begin{tabular}[c]{@{}c@{}}With \\ SAM\end{tabular}}    & \includegraphics[width=0.25\textwidth]{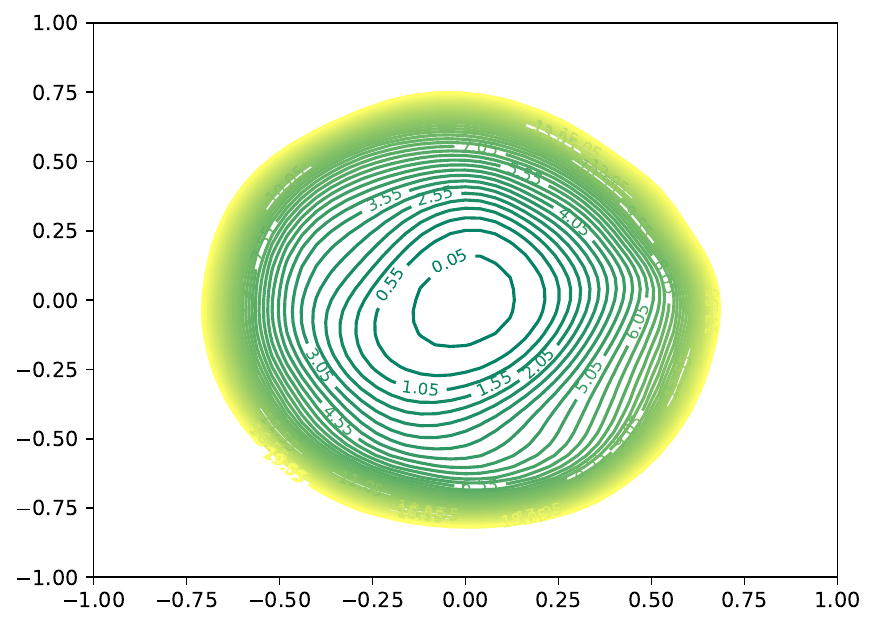}                    & \includegraphics[width=0.25\textwidth]{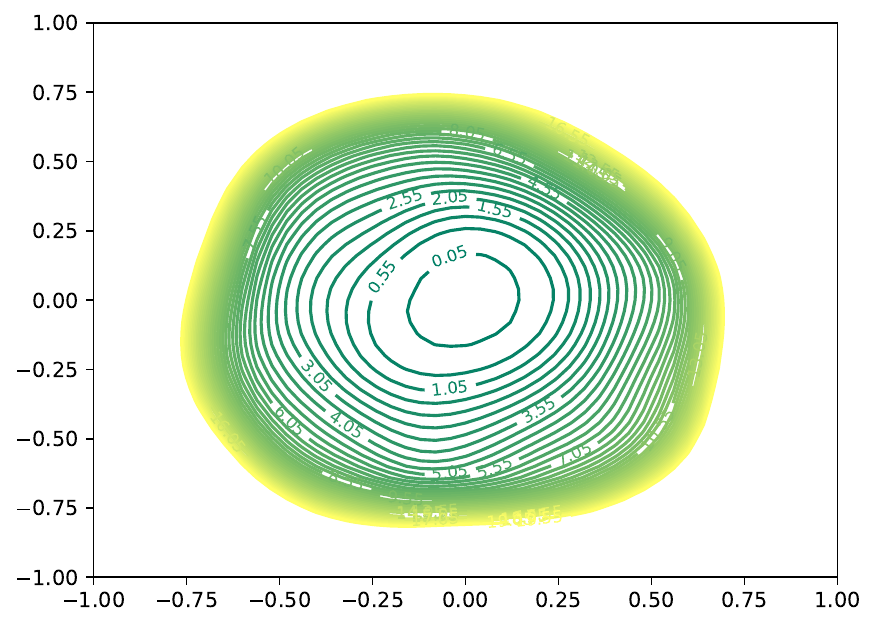}                        & \includegraphics[width=0.25\textwidth]{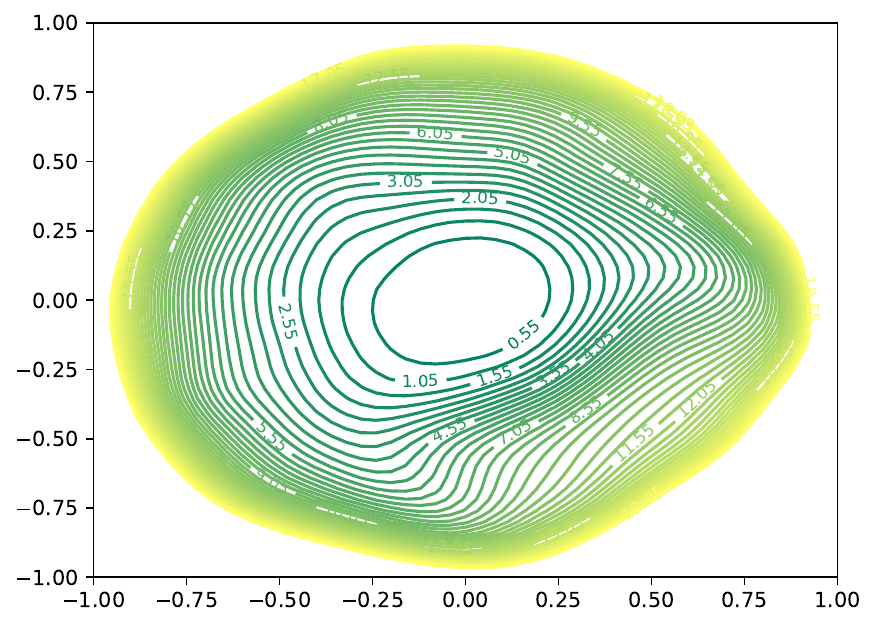}                          \\ \hline
\end{tabular}
\caption{Loss landscape visualisation~\cite{li2018visualizing} of ResNet18 landscape on CIFAR10, exploring the loss in the domain of the perturbations $[-1,1]^2$ with 51 steps in both directions.}
\label{fig:c10_res_vis}
\end{table}
\textit{Sharper Minima Can Coincide with Stronger Reliability-Related Properties.} 
Across all CIFAR data sets, we consistently observe that the baseline control yields the flattest solutions, yet performs worst on reliability-related evaluations. In contrast, the controls that achieve the best performance on these metrics are always sharper than the baseline under most of the sharpness measures considered. The relationship between regularisation, increased sharpness and improved reliability performance is largely statistically supported in Tables~\ref{tab:res-sign-c10},~\ref{tab:res-sign-c100} and~\ref{tab:res-sign-TiN}. These results suggest that sharper minima can coincide with improved reliability properties, indicating that sharpness may in some settings be an important factor in the learning of more reliable models. Notably, there are instances, although limited in number, where a regularisation technique which sharpens the minima found at the end of training does not improve reliability and generalisation performance against the baseline. This shows that sharpness can be beneficial in some settings but does not by itself necessitate improved performance. However, from the results in this paper a sharper solution has a higher probability of increasing performance against flatter minima and therefore may provide an improved bias for some learning problems. 

A possible explanation for the relationship we observe between generalisation, reliability and sharpness in high-dimensional learning problems is suggested by our decision-boundary experiments in Section~\ref{sec:decision_boundaries}. There we show that sharper minima correspond not only to memorisation, but also to well-generalising models with tighter decision boundaries, which may be beneficial in certain tasks. This provides a useful lens through which to interpret our findings: improved reliability-related performance does not require flatness, and may in some cases arise alongside sharper solutions. 

\textit{We Find No Geometric Goldilocks Zone for Sharpness.}
Although we report statistical support for sharper solutions performing better across generalisation and reliability-related metrics on the CIFAR data sets, the sharpest model is not always the best overall. Still, the top-performing model is typically sharper than the baseline, suggesting that a learning task may require a level of sharpness beyond what is induced by the architecture's implicit bias alone. This supports the view that neither extreme flatness nor extreme sharpness is universally optimal. This insight offers further refinement of the perspective developed from Section~\ref{sec:decision_boundaries} where we saw that extreme sharpness could indicate memorisation while intermediate sharpness showed a strong relationship with tightening decision boundaries. Instead, the ``right" level of sharpness appears to be task- and architecture-dependent.
More broadly, this highlights the risk of misleading conclusions when aggregating sharpness trends across heterogeneous architectures: we observe that apparent general trends can invert under such aggregation, consistent with Simpson's Paradox~\citep{simpson1951interpretation}. Careful control over architecture-specific inductive biases is therefore essential when studying geometry--function relationships.

\begin{table}[H]
\centering
\resizebox{\textwidth}{!}{
\begin{tabular}{|c|c|c|c|c|c|c|c|c|}
\hline
\textbf{\begin{tabular}[c]{@{}c@{}}Control\\ Condition\end{tabular}} & \textbf{\begin{tabular}[c]{@{}c@{}}Generalisation \\ Gap\end{tabular}} & \textbf{\begin{tabular}[c]{@{}c@{}}Test \\ Accuracy\end{tabular}} & \textbf{\begin{tabular}[c]{@{}c@{}}Test \\ ECE\end{tabular}} & \textbf{\begin{tabular}[c]{@{}c@{}}Corruption \\ Accuracy\end{tabular}} & \textbf{\begin{tabular}[c]{@{}c@{}}Prediction \\ Disagreement\end{tabular}} & \textbf{\begin{tabular}[c]{@{}c@{}}SAM \\ Sharpness\end{tabular}} & \textbf{\begin{tabular}[c]{@{}c@{}}Fisher-Rao \\ norm\end{tabular}} & \textbf{\begin{tabular}[c]{@{}c@{}}Relative \\ Flatness\end{tabular}} \\ \hline
Baseline                                                             & 47.010 $\pm{0.166}$                                                         & 0.530 $\pm{0.002}$                                                     & 0.220 $\pm{0.001}$                                                & 38.760 $\pm{0.085}$                                                          & 0.452 $\pm{0.000}$                                                               & 2.607E-04 $\pm{3.147E-05}$                                             & 0.294 $\pm{0.028}$                                                       & 32.085 $\pm{0.313}$                                                        \\ \hline
\begin{tabular}[c]{@{}c@{}}Baseline\\ + SAM\end{tabular}             & 44.421 $\pm{0.168}$                                                         & 0.556 $\pm{0.002}$                                                     & 0.191 $\pm{0.002}$                                                & 41.888 $\pm{0.098}$                                                          & 0.410 $\pm{0.000}$                                                               & 4.231E-04 $\pm{4.973E-05}$                                             & 0.399 $\pm{0.014}$                                                       & 123.791 $\pm{4.185}$                                                       \\ \hline
Augmentation                                                         & 29.642 $\pm{0.133}$                                                         & 0.697 $\pm{0.002}$                                                     & 0.185 $\pm{0.001}$                                                & 44.613 $\pm{0.169}$                                                          & 0.288 $\pm{0.001}$                                                               & 1.110E-01 $\pm{9.173E-03}$                                             & 3.587 $\pm{0.150}$                                                       & 2766.925 $\pm{178.669}$                                                    \\ \hline
\begin{tabular}[c]{@{}c@{}}Augmentation\\ + SAM\end{tabular}         & \textbf{28.999 $\pm{0.092}$}                                                & \textbf{0.705 $\pm{0.001}$}                                            & 0.145 $\pm{0.001}$                                                & \textbf{45.428 $\pm{0.217}$}                                                 & \textbf{0.269 $\pm{0.000}$}                                                      & 1.081E-01 $\pm{1.636E-02}$                                             & 4.179 $\pm{0.032}$                                                       & 4196.832 $\pm{52.606}$                                                     \\ \hline
Weight Decay                                                         & 47.838 $\pm{0.301}$                                                         & 0.521 $\pm{0.003}$                                                     & \textbf{0.099 $\pm{0.005}$}                                       & 37.868 $\pm{0.265}$                                                          & 0.474 $\pm{0.001}$                                                               & 5.192E-04 $\pm{8.009E-05}$                                             & 0.861 $\pm{0.116}$                                                       & 136.969 $\pm{7.484}$                                                       \\ \hline
\begin{tabular}[c]{@{}c@{}}Weight Decay\\ + SAM\end{tabular}         & 45.644 $\pm{0.117}$                                                         & 0.543 $\pm{0.001}$                                                     & 0.106 $\pm{0.002}$                                                & 40.604 $\pm{0.222}$                                                          & 0.444 $\pm{0.001}$                                                               & 1.528E-03 $\pm{1.427E-04}$                                             & 1.788 $\pm{0.069}$                                                       & 360.271 $\pm{16.190}$                                                      \\ \hline
\end{tabular}}
\caption{Results for ResNet18 trained on CIFAR100. \textbf{Bolded} values indicate the best performance per metric. For sharpness metrics, lower values correspond to flatter models.}
\label{tab:c100_resnet18}
\end{table}

\begin{table}[H]
\resizebox{\textwidth}{!}{
\begin{tabular}{|c|c|c|c|c|c|c|c|c|}
\hline
\textbf{\begin{tabular}[c]{@{}c@{}}Control\\ Condition\end{tabular}} & \textbf{\begin{tabular}[c]{@{}c@{}}Generalisation \\ Gap\end{tabular}} & \textbf{\begin{tabular}[c]{@{}c@{}}Test \\ Accuracy\end{tabular}} & \textbf{\begin{tabular}[c]{@{}c@{}}Test \\ ECE\end{tabular}} & \textbf{\begin{tabular}[c]{@{}c@{}}Corruption \\ Accuracy\end{tabular}} & \textbf{\begin{tabular}[c]{@{}c@{}}Prediction \\ Disagreement\end{tabular}} & \textbf{\begin{tabular}[c]{@{}c@{}}SAM \\ Sharpness\end{tabular}} & \textbf{\begin{tabular}[c]{@{}c@{}}Fisher-Rao \\ norm\end{tabular}} & \textbf{\begin{tabular}[c]{@{}c@{}}Relative \\ Flatness\end{tabular}} \\ \hline
\begin{tabular}[c]{@{}c@{}}Baseline \\ + SAM\end{tabular}            & \cmark~(1.0)  & \cmark~(1.0)  & \cmark~(1.0) & \cmark~(1.0) & \cmark~(1.0)  & \cmark~(0.93) & \cmark~(0.89) & \cmark~(1.0) \\ \hline
Augmentation                                                         & \cmark~(1.0)  & \cmark~(1.0)  & \cmark~(1.0) & \cmark~(1.0) & \cmark~(1.0)  & \cmark~(1.0)  & \cmark~(1.0)  & \cmark~(1.0) \\ \hline
\begin{tabular}[c]{@{}c@{}}Augmentation \\ + SAM\end{tabular}        & \cmark~(1.0)  & \cmark~(1.0)  & \cmark~(1.0) & \cmark~(1.0) & \cmark~(1.0)  & \cmark~(1.0)  & \cmark~(1.0)  & \cmark~(1.0) \\ \hline
Weight Decay                                                         & \xmark~(N/A)   & \xmark~(N/A)  & \cmark~(1.0) & \xmark~(N/A) & \xmark~(N/A)   & \cmark~(1.0)  & \cmark~(1.0)  & \cmark~(1.0) \\ \hline
\begin{tabular}[c]{@{}c@{}}Weight Decay \\ + SAM\end{tabular}        & \cmark~(0.96) & \cmark~(0.96) & \cmark~(1.0) & \cmark~(1.0) & \cmark~(0.92) & \cmark~(1.0)  & \cmark~(1.0)  & \cmark~(1.0) \\ \hline
\end{tabular}}
\caption{Significance results for the control conditions against the baseline for ResNet18 trained on CIFAR100.~\cmark~indicates significant difference compared to the baseline; \xmark~indicates no significance, the value in the brackets represents the  effect sizes using the absolute RBC}.
\label{tab:res-sign-c100}
\end{table}
\begin{table}[H]
\centering
\begin{tabular}{|c|c|c|c|}
\hline
                     & \textbf{Baseline} & \textbf{Weight Decay} & \textbf{Augmentation} \\ \hline
\textbf{\begin{tabular}[c]{@{}c@{}}Without \\ SAM\end{tabular}} &  \includegraphics[width=0.25\textwidth]{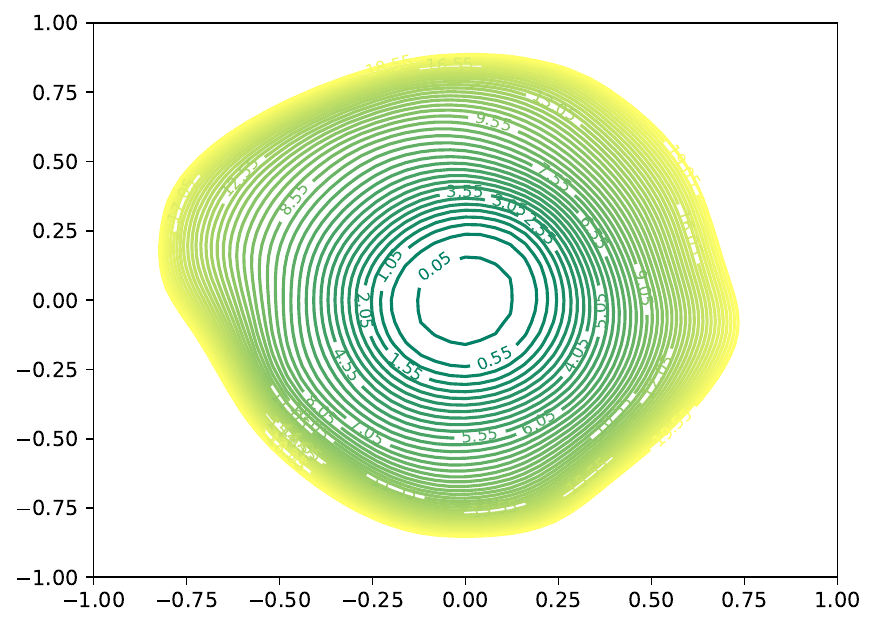}                  & \includegraphics[width=0.25\textwidth]{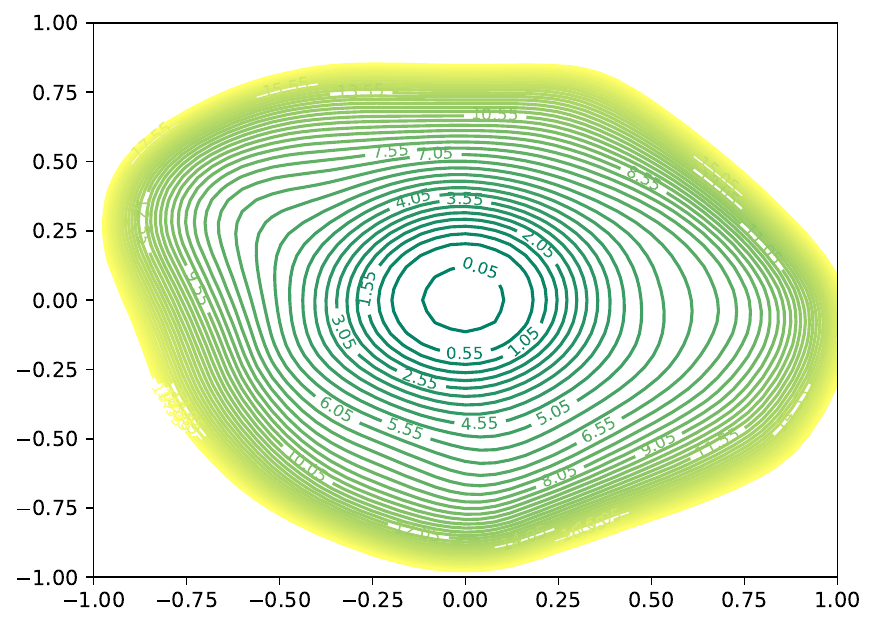}                          &  \includegraphics[width=0.25\textwidth]{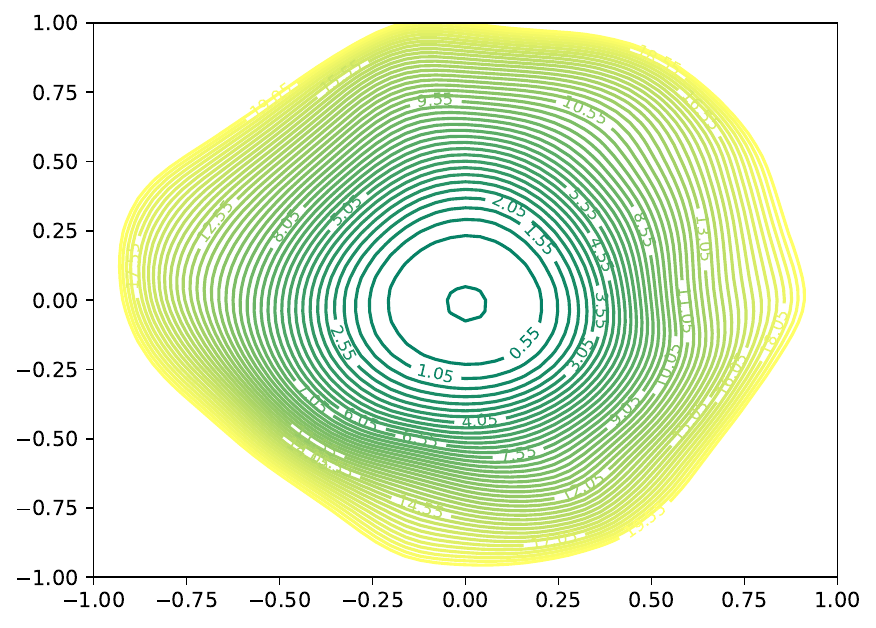}                       \\ \hline
\textbf{\begin{tabular}[c]{@{}c@{}}With \\ SAM\end{tabular}}    & \includegraphics[width=0.25\textwidth]{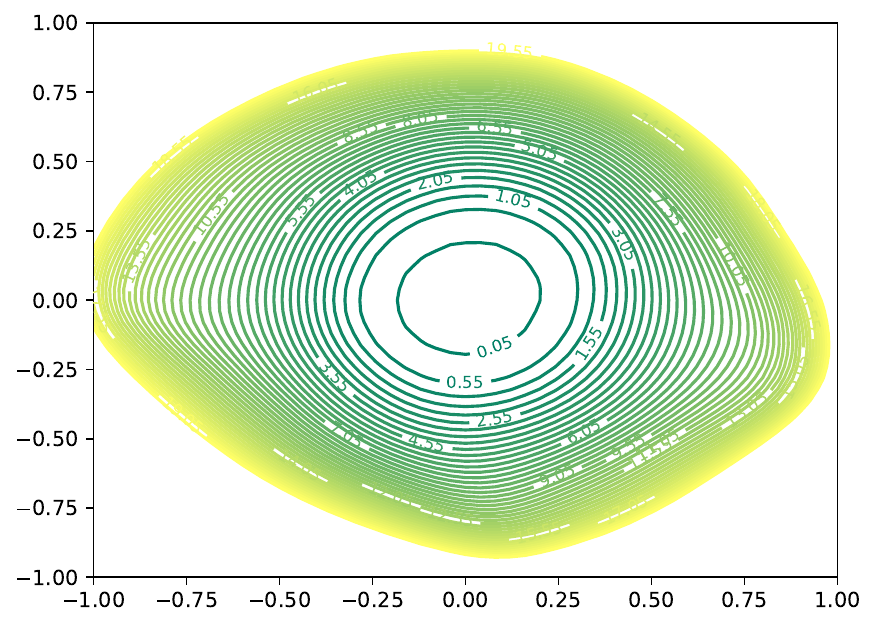}                    & \includegraphics[width=0.25\textwidth]{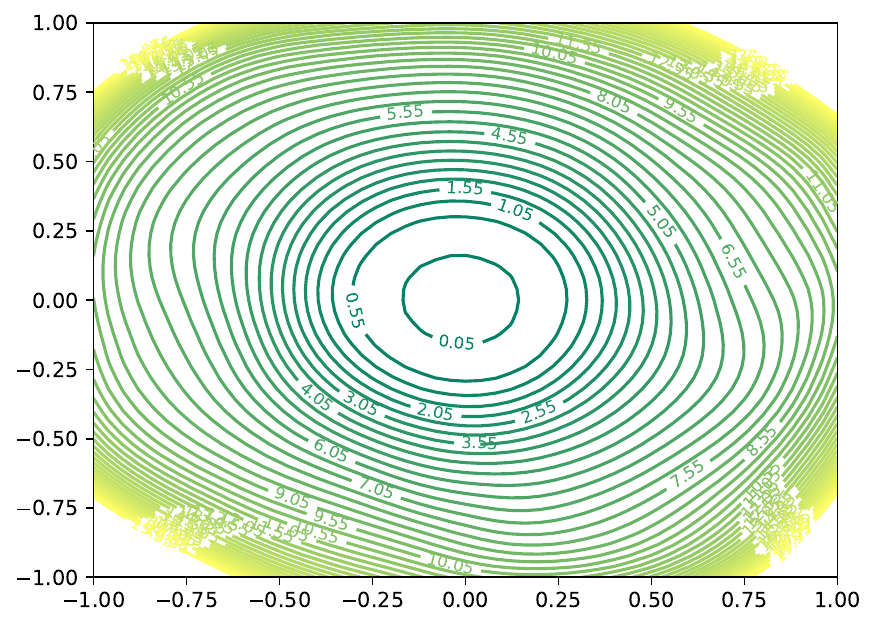}                        & \includegraphics[width=0.25\textwidth]{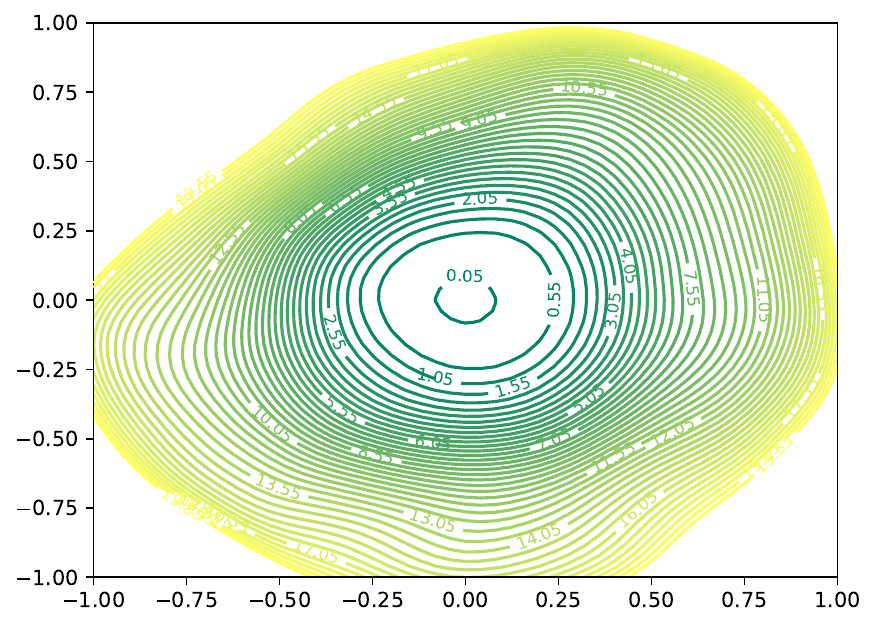}                          \\ \hline
\end{tabular}
\caption{Loss landscape visualisation~\cite{li2018visualizing} of ResNet18 landscape on CIFAR100, exploring the loss in the domain of the perturbations $[-1,1]^2$ with 51 steps in both directions.}
\label{tab:c100_vis}
\end{table}
\textit{Limitations of Loss Landscape Visualisations.}
Loss landscape visualisations (Figures~\ref{fig:c10_res_vis}, \ref{tab:c100_vis}), produced using the method of~\citet{li2018visualizing}, have been used to support the benefits of flatness generally~\citep{foret2021sharpnessaware,pmlr-v137-huang20a} and qualitatively illustrate that regularisation, especially SAM, alters the geometry of the solution. However, these plots often show that applying regularisation results in broader minima in some directions, even when reparametrisation-invariant sharpness metrics increase in sharpness in a statistically supported manner. This apparent mismatch underscores the limitations of low-dimensional loss surface plots, which capture only 2D projections of high-dimensional landscapes. In contrast, sharpness metrics reflect geometric properties beyond local projections. While visualisations can help convey functional changes, we argue that metric-based evaluations provide a more consistent and interpretable picture of sharpness.

\begin{table}[H]
\centering
\resizebox{\textwidth}{!}{
\begin{tabular}{|c|c|c|c|c|c|c|c|}
\hline
\textbf{\begin{tabular}[c]{@{}c@{}}Control\\ Condition\end{tabular}} & \textbf{\begin{tabular}[c]{@{}c@{}}Generalisation \\ Gap\end{tabular}} & \textbf{\begin{tabular}[c]{@{}c@{}}Test \\ Accuracy\end{tabular}} & \textbf{\begin{tabular}[c]{@{}c@{}}Test \\ ECE\end{tabular}} & \textbf{\begin{tabular}[c]{@{}c@{}}Corruption \\ Accuracy\end{tabular}} & \textbf{\begin{tabular}[c]{@{}c@{}}Prediction \\ Disagreement\end{tabular}} & \textbf{\begin{tabular}[c]{@{}c@{}}SAM \\ Sharpness\end{tabular}} & \textbf{\begin{tabular}[c]{@{}c@{}}Fisher-Rao \\ norm\end{tabular}} \\ \hline
Baseline                                                             & 49.643 $\pm{0.103}$                                                         & 0.503 $\pm{0.001}$                                                     & 0.257 $\pm{0.001}$                                                & 28.347 $\pm{0.945}$                                                          & 0.385 $\pm{0.000}$                                                               & 3.202E-04 $\pm{9.872E-06}$                                             & 0.479 $\pm{0.002}$                                                       \\ \hline
\begin{tabular}[c]{@{}c@{}}Baseline\\ + SAM\end{tabular}             & 46.255 $\pm{0.128}$                                                         & 0.537 $\pm{0.001}$                                                     & 0.223 $\pm{0.001}$                                                & 28.984 $\pm{0.865}$                                                          & 0.344 $\pm{0.000}$                                                               & 3.080E-04 $\pm{8.424E-06}$                                             & 0.427 $\pm{0.004}$                                                       \\ \hline
Augmentation                                                         & 19.993 $\pm{0.091}$                                                         & 0.508 $\pm{0.001}$                                                     & 0.102 $\pm{0.001}$                                                & 34.457 $\pm{0.875}$                                                          & 0.544 $\pm{0.000}$                                                               & 1.680E+00 $\pm{8.776E-02}$                                             & 25.887 $\pm{0.098}$                                                      \\ \hline
\begin{tabular}[c]{@{}c@{}}Augmentation\\ + SAM\end{tabular}         & \textbf{16.777 $\pm{0.084}$}                                                & 0.520 $\pm{0.001}$                                                     & \textbf{0.044 $\pm{0.001}$}                                       & \textbf{35.324 $\pm{0.865}$}                                                 & 0.514 $\pm{0.000}$                                                               & 1.446E+00 $\pm{6.332E-02}$                                             & 25.193 $\pm{0.034}$                                                      \\ \hline
Weight Decay                                                         & 49.689 $\pm{0.092}$                                                         & 0.503 $\pm{0.001}$                                                     & 0.202 $\pm{0.001}$                                                & 28.997 $\pm{0.959}$                                                          & 0.384 $\pm{0.000}$                                                               & 2.297E-04 $\pm{9.718E-06}$                                             & 0.998 $\pm{0.002}$                                                       \\ \hline
\begin{tabular}[c]{@{}c@{}}Weight Decay\\ + SAM\end{tabular}         & 46.061 $\pm{0.111}$                                                         & \textbf{0.539 $\pm{0.001}$}                                            & 0.177 $\pm{0.001}$                                                & 34.549 $\pm{0.057}$                                                          & \textbf{0.339 $\pm{0.000}$}                                                      & 3.784E-04 $\pm{9.996E-06}$                                             & 0.736 $\pm{0.004}$                                                       \\ \hline
\end{tabular}}
\caption{Results for ResNet18 (Pre-Trained) on TinyImageNet. \textbf{Bolded} values indicate the best performance per metric. For sharpness metrics, lower values correspond to flatter models.}
\label{tab:TiN_resnet18}
\end{table}

\begin{table}[H]
\resizebox{\textwidth}{!}{
\begin{tabular}{|c|c|c|c|c|c|c|c|}
\hline
\textbf{\begin{tabular}[c]{@{}c@{}}Control\\ Condition\end{tabular}} & \textbf{\begin{tabular}[c]{@{}c@{}}Generalisation \\ Gap\end{tabular}} & \textbf{\begin{tabular}[c]{@{}c@{}}Test \\ Accuracy\end{tabular}} & \textbf{\begin{tabular}[c]{@{}c@{}}Test \\ ECE\end{tabular}} & \textbf{\begin{tabular}[c]{@{}c@{}}Corruption \\ Accuracy\end{tabular}} & \textbf{\begin{tabular}[c]{@{}c@{}}Prediction \\ Disagreement\end{tabular}} & \textbf{\begin{tabular}[c]{@{}c@{}}SAM \\ Sharpness\end{tabular}} & \textbf{\begin{tabular}[c]{@{}c@{}}Fisher-Rao \\ norm\end{tabular}} \\ \hline
\begin{tabular}[c]{@{}c@{}}Baseline \\ + SAM\end{tabular}            & \cmark~(1.0) & \cmark~(1.0) & \cmark~(1.0) & \xmark~(N/A) & \cmark~(1.0)  & \xmark~(N/A) & \xmark~(N/A) \\ \hline
Augmentation                                                         & \cmark~(1.0) & \cmark~(0.8) & \cmark~(1.0) & \cmark~(1.0) & \xmark~(N/A)   & \cmark~(1.0) & \cmark~(1.0) \\ \hline
\begin{tabular}[c]{@{}c@{}}Augmentation \\ + SAM\end{tabular}        & \cmark~(1.0) & \cmark~(1.0) & \cmark~(1.0) & \cmark~(1.0) & \xmark~(N/A)   & \cmark~(1.0) & \cmark~(1.0) \\ \hline
Weight Decay                                                         & \xmark~(N/A)  & \xmark~(N/A) & \cmark~(1.0) & \xmark~(N/A) & \cmark~(0.51) & \xmark~(N/A) & \cmark~(1.0) \\ \hline
\begin{tabular}[c]{@{}c@{}}Weight Decay \\ + SAM\end{tabular}        & \cmark~(1.0) & \cmark~(1.0) & \cmark~(1.0) & \cmark~(1.0) & \cmark~(1.0)  & \cmark~(1.0) & \cmark~(1.0) \\ \hline
\end{tabular}}
\caption{Significance results for the control conditions against the baseline for ResNet18 trained on TinyImageNet.~\cmark~indicates significant difference compared to the baseline; \xmark~indicates no significance, the value in the brackets represents the  effect sizes using the absolute RBC}.
\label{tab:res-sign-TiN}
\end{table}
\textit{Geometric Studies of Loss Landscapes Should Consider The Simpson's Paradox}. We observe that under different optimisation parameters seen in learning rate (1e-2 and 1e-3) and batch sizes (128 and 256), on the ResNet18 and VGG19 trained on CIFAR10, that optimisation settings can bias models to flatter solutions. For example, for the ResNet18 when we use a batch size of 128 and a learning rate of 1e-2 (see Table~\ref{tab:c10_resnet18_128_0.01})  models are flatter than in other batch size and learning rate settings (Tables~\ref{tab:c10_resnet18},~\ref{tab:c10_resnet18_256_0.01} and~\ref{tab:128-1e3-c10}). As a result, this highlights that for some optimisation conditions a Simpson's paradox could emerge, where a general trend for flatness could be preferred. However, despite this, what remains consistent across all hyperparameters and training settings explored in this work is that, regardless of a specific optimisation configuration and geometric bias, regularisation strategies consistently increase the sharpness of the model in matched initialisation and optimisation and training order conditions. As such, the identification of Simpson's paradox for minima geometry actively supports the position that caution should be exercised when attempting to compare the geometric properties of models with different architectures or models of the same architecture that are trained under different optimisation hyperparameters, as this could unintentionally bias the geometry of the models analysed and lead to incorrect findings. We argue that analysing minima geometry requires strict experimental controls, such as those that we have demonstrated in our matched initialisation, data order, training hyperparameters, and multiple-seed analysis of neural network training dynamics and minima properties. 

\subsection{Understanding SAM's Increased Sharpness Through the \texorpdfstring{$\boldsymbol{\rho}$}{ρ} Hyperparameter}
\label{sec:rho_SAM_seep}
\begin{figure}[H]
    \centering
    \subfigure[Fisher-Rao norm and Loss]{\includegraphics[width=0.32\linewidth] {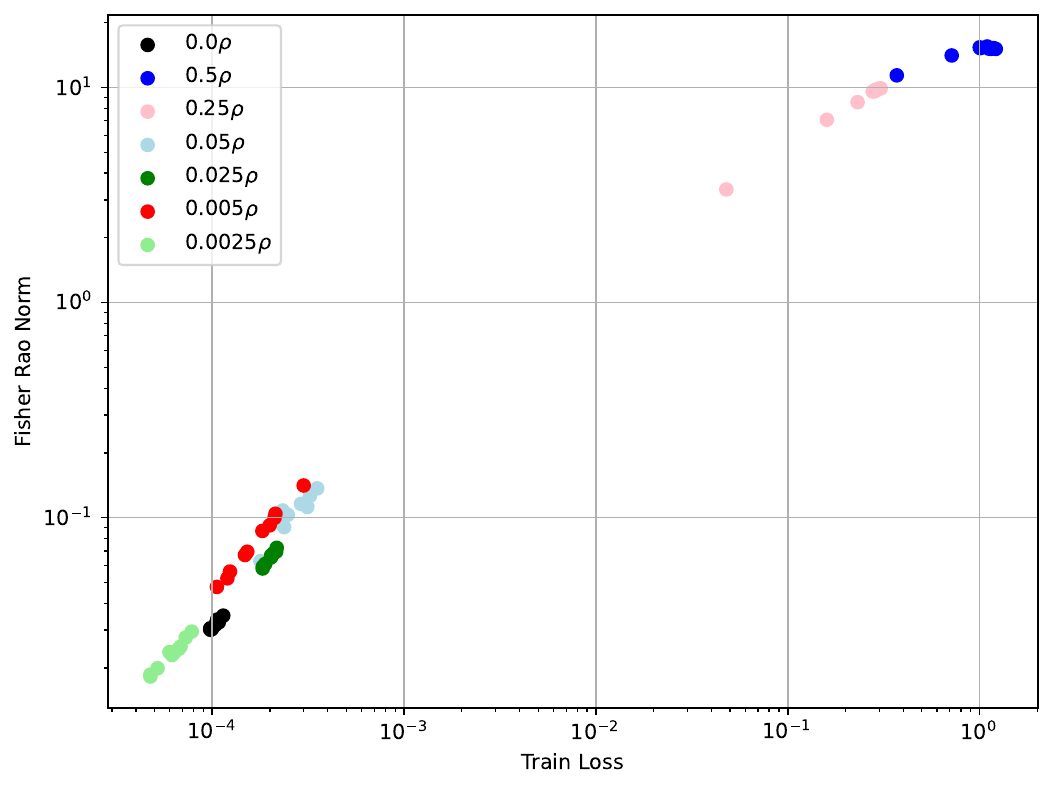}}
    \subfigure[Relative Flatness and Loss]{\includegraphics[width=0.315\linewidth] {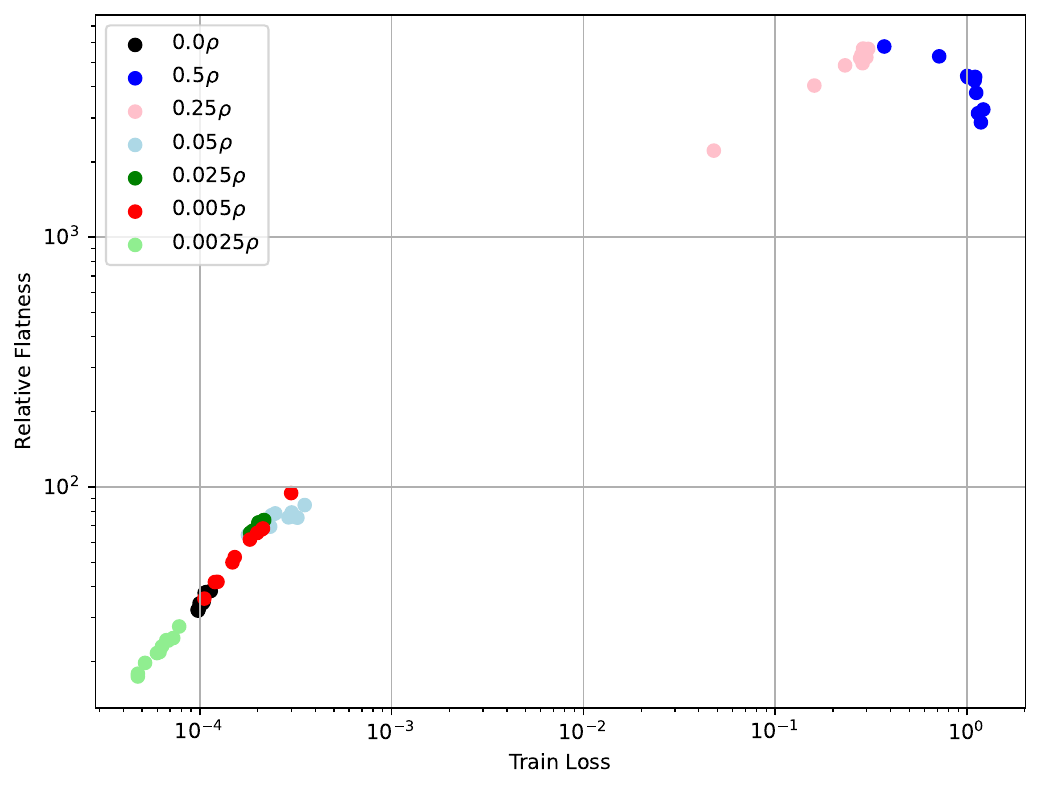}}
    \subfigure[Generalisation Gap and Loss]{\includegraphics[width=0.32\linewidth] {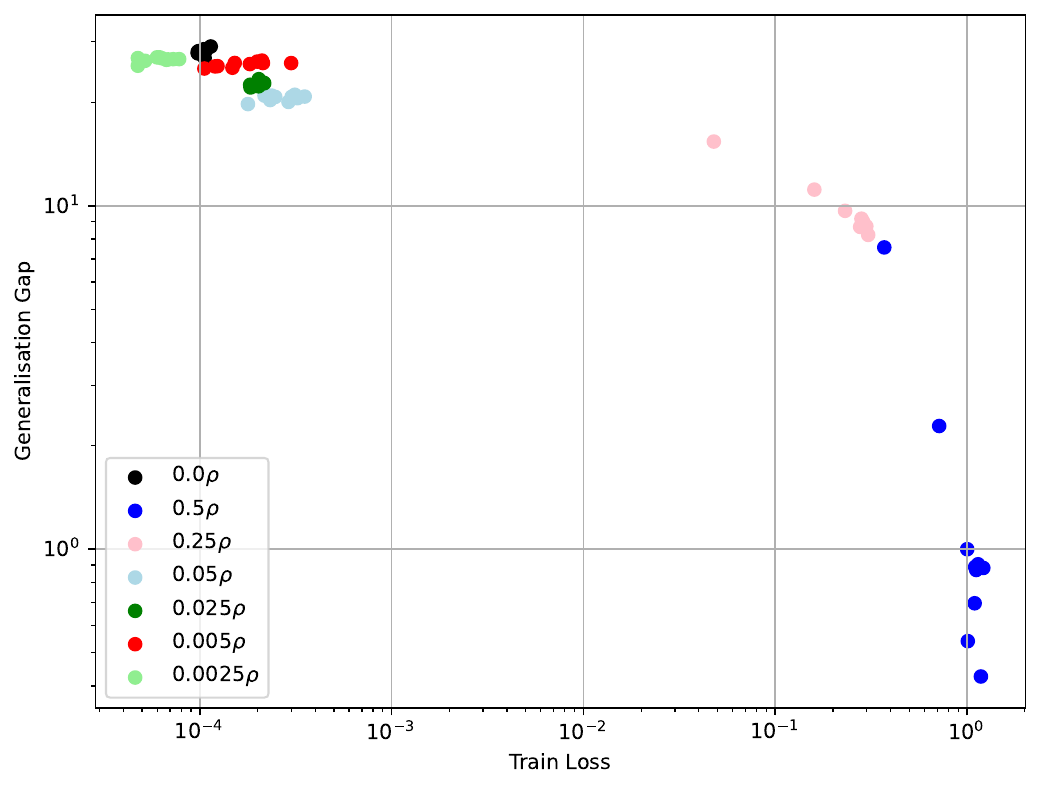}}
    \caption{Scatter plots of 60 minima for ResNet-18 on CIFAR-10 varying the SAM ($\rho$) hyperparameter (${0.5, 0.25 ,0.05 ,0.025, 0.005, 0.0025}$), using batch size ${256}$ and learning rate ${1^{-3}}$: (a) Fisher–Rao norm vs. train loss, (b) Relative Flatness vs. train loss, and (c) generalisation gap vs. train loss (log scale).}
   \label{fig:sharpness_gen_gap_plot_app}
\end{figure}

In this subsection we show that our finding of increased sharpness under SAM is robust to perturbations of the $\rho$ hyperparameter. We vary $\rho$ across the values ${0.5, 0.25, 0.05, 0.025, 0.005, 0.0025}$, training ResNet-18 with a \textbf{batch size 256 and a learning rate $\mathbf{1e^{-3}}$}. As shown in Figure~\ref{fig:sharpness_gen_gap_plot_app}, increasing $\rho$  increases the sharpness of the minima found at the end of training, which often coincides with a reduced generalisation gap. We find statistical support for the majority of $\rho$ values, showing that most values of $\rho$ coincide with increased reliability performance and sharpness in Table~\ref{tab:sig-rho} against the baseline.
Table~\ref{tab:rho_table} shows that when $\rho=0.25$, we obtain the best test accuracy, calibration, corruption robustness, and functional consistency. Notably, this condition is substantially sharper than the lower-$\rho$ values, again suggesting a relationship between increased sharpness and improved generalisation properties. 

\begin{table}[H]
\centering
\resizebox{\textwidth}{!}{
\begin{tabular}{|c|c|c|c|c|c|c|c|c|}
\hline
\textbf{\begin{tabular}[c]{@{}c@{}}$\rho$ \\ Radius \end{tabular}} & \textbf{\begin{tabular}[c]{@{}c@{}}Generalisation \\ Gap\end{tabular}} & \textbf{\begin{tabular}[c]{@{}c@{}}Test \\ Accuracy\end{tabular}} & \textbf{\begin{tabular}[c]{@{}c@{}}Test \\ ECE\end{tabular}} & \textbf{\begin{tabular}[c]{@{}c@{}}Corruption \\ Accuracy\end{tabular}} & \textbf{\begin{tabular}[c]{@{}c@{}}Prediction \\ Disagreement\end{tabular}} & \textbf{\begin{tabular}[c]{@{}c@{}}SAM \\ Sharpness\end{tabular}} & \textbf{\begin{tabular}[c]{@{}c@{}}Fisher-Rao \\ norm\end{tabular}} & \textbf{\begin{tabular}[c]{@{}c@{}}Relative \\ Flatness\end{tabular}} \\ \hline
0.000              & 28.050 $\pm{0.175}$                                                         & 0.720 $\pm{0.002}$                                                     & 0.186 $\pm{0.001}$                                                & 58.614 $\pm{0.201}$                                                          & 0.282 $\pm{0.001}$                                                               & 1.366E-05 $\pm{1.206E-06}$                                             & 0.032 $\pm{0.001}$                                                       & 34.607 $\pm{0.757}$                                                        \\ \hline
0.5000             & \textbf{1.605 $\pm{0.646}$}                                                 & 0.629 $\pm{0.024}$                                                     & 0.079 $\pm{0.007}$                                                & 52.847 $\pm{1.667}$                                                          & 0.221 $\pm{0.009}$                                                               & 6.814E-02 $\pm{7.326E-03}$                                             & 14.767 $\pm{0.375}$                                                      & 4156.344 $\pm{279.557}$                                                    \\ \hline
0.2500             & 9.751 $\pm{0.640}$                                                          & \textbf{0.835 $\pm{0.002}$}                                            & \textbf{0.026 $\pm{0.003}$}                                       & \textbf{68.479 $\pm{0.302}$}                                                 & \textbf{0.089 $\pm{0.002}$}                                                      & 3.884E-02 $\pm{4.981E-03}$                                             & 8.712 $\pm{0.623}$                                                       & 4876.348 $\pm{314.164}$                                                    \\ \hline
0.0500             & 20.588 $\pm{0.125}$                                                         & 0.794 $\pm{0.001}$                                                     & 0.108 $\pm{0.001}$                                                & 66.342 $\pm{0.164}$                                                          & 0.168 $\pm{0.000}$                                                               & 5.823E-05 $\pm{9.056E-06}$                                             & 0.107 $\pm{0.006}$                                                       & 75.093 $\pm{1.693}$                                                        \\ \hline
0.0250             & 22.602 $\pm{0.109}$                                                         & 0.774 $\pm{0.001}$                                                     & 0.124 $\pm{0.001}$                                                & 64.224 $\pm{0.154}$                                                          & 0.195 $\pm{0.000}$                                                               & 2.587E-05 $\pm{1.987E-06}$                                             & 0.065 $\pm{0.001}$                                                       & 70.223 $\pm{0.941}$                                                        \\ \hline
0.0050             & 25.793 $\pm{0.137}$                                                         & 0.742 $\pm{0.001}$                                                     & 0.167 $\pm{0.001}$                                                & 60.985 $\pm{0.280}$                                                          & 0.250 $\pm{0.000}$                                                               & 4.861E-05 $\pm{7.166E-06}$                                             & 0.082 $\pm{0.009}$                                                       & 57.886 $\pm{5.223}$                                                        \\ \hline
0.0025             & 26.654 $\pm{0.130}$                                                         & 0.733 $\pm{0.001}$                                                     & 0.176 $\pm{0.001}$                                                & 60.107 $\pm{0.226}$                                                          & 0.262 $\pm{0.001}$                                                               & 8.624E-06 $\pm{7.512E-07}$                                             & 0.023 $\pm{0.001}$                                                       & 22.262 $\pm{0.969}$                                                        \\ \hline
\end{tabular}}
\caption{Results for ResNet-18 trained on CIFAR10 with \textbf{batch size 256 and a learning rate of $\mathbf{1e^{-3}}$}, while varying the SAM $\rho$ hyperperameter (0.5, 0.25 ,0.05 ,0.025, 0.005, 0.0025). Numbers in bold indicate best scores per metric. For sharpness metrics lower values represent flatter models.}
\label{tab:rho_table}
\end{table}

\begin{table}[H]
\centering
\resizebox{\textwidth}{!}{
\begin{tabular}{|c|c|c|c|c|c|c|c|c|}
\hline
\textbf{\begin{tabular}[c]{@{}c@{}}$\rho$ \\ Radius\end{tabular}} & \textbf{\begin{tabular}[c]{@{}c@{}}Generalisation \\ Gap\end{tabular}} & \textbf{\begin{tabular}[c]{@{}c@{}}Test\\ Accuracy\end{tabular}} & \textbf{\begin{tabular}[c]{@{}c@{}}Test \\ ECE\end{tabular}} & \textbf{\begin{tabular}[c]{@{}c@{}}Corruption \\ Accuracy\end{tabular}} & \textbf{\begin{tabular}[c]{@{}c@{}}Prediction \\ Disagreement\end{tabular}} & \textbf{\begin{tabular}[c]{@{}c@{}}SAM \\ Sharpness\end{tabular}} & \textbf{\begin{tabular}[c]{@{}c@{}}Fisher-Rao \\ norm\end{tabular}} & \textbf{\begin{tabular}[c]{@{}c@{}}Relative \\ Flatness\end{tabular}} \\ \hline
0.5000                                                            & \cmark~(1.0) & \xmark~(N/A) & \cmark~(1.0) & \xmark~(N/A) & \cmark~(0.69) & \cmark~(1.0) & \cmark~(1.0) & \cmark~(1.0)  \\ \hline
0.2500                                                            & \cmark~(1.0) & \cmark~(1.0) & \cmark~(1.0) & \cmark~(1.0) & \cmark~(1.0)  & \cmark~(1.0) & \cmark~(1.0) & \cmark~(1.0)  \\ \hline
0.0500                                                            & \cmark~(1.0) & \cmark~(1.0) & \cmark~(1.0) & \cmark~(1.0) & \cmark~(1.0)  & \cmark~(1.0) & \cmark~(1.0) & \cmark~(1.0)  \\ \hline
0.0250                                                            & \cmark~(1.0) & \cmark~(1.0) & \cmark~(1.0) & \cmark~(1.0) & \cmark~(1.0)  & \cmark~(1.0) & \cmark~(1.0) & \cmark~(1.0)  \\ \hline
0.0050                                                            & \cmark~(1.0) & \cmark~(1.0) & \cmark~(1.0) & \cmark~(1.0) & \cmark~(1.0)  & \cmark~(1.0) & \cmark~(1.0) & \cmark~(0.96) \\ \hline
0.0025                                                            & \cmark~(1.0) & \cmark~(1.0) & \cmark~(1.0) & \cmark~(1.0) & \cmark~(1.0)  & \xmark~(N/A) & \xmark~(N/A) & \xmark~(N/A)  \\ \hline
\end{tabular}}
\caption{Significance Table for ResNet-18 trained on CIFAR10 with \textbf{batch size 256 and a learning rate of} $\mathbf{1e^{-3}}$, while varying the SAM $\rho$ hyperperameter (0.5, 0.25 ,0.05 ,0.025, 0.005, 0.0025).~\cmark~indicates significant difference compared to the baseline ($\rho$ = 0.000); \xmark~indicates significance, the value in the brackets represents the  effect sizes using the absolute RBC.}
\label{tab:sig-rho}
\end{table}
Although, the empirical evidence presented raises an important caveat. The sharpest condition overall, found at $\rho=0.50$, is not the best-performing model. As a result, this reinforces our broader interpretation that the sharpness required to fit a function and the resulting performance is highly dependent on the learning problem itself as there appears to be no universal Goldilocks zone of sharpness that is sufficient across tasks in our settings. We also record an instance in this setting at a very low value of $\rho={0.0025}$ where the resulting model is statistically flatter than the baseline and outperforms on generalisation and reliability metrics, but is generally not the best model compared to sharper models under an increased $\rho$ radius. 

\subsection{Explicitly Increasing Function Complexity on High-Dimensional Data}
\label{Sec:Increasing_func_complexity}
To further demonstrate the relationship between sharp minima and increased function complexity, suggested already by the toy-setting results in Section~\ref{sec:single_objective_optimisation} and ~\ref{sec:decision_boundaries}, we conduct an additional experiment in the classification setting on high-dimensional data. Here, we artificially increase the difficulty of the learning task and record the sharpness of the minima at the end of training. \textbf{Importantly, we calculate the sharpness of all models at the end of training on the same unaltered training data set}. To do this, we randomise the labels in the CIFAR10 training data set in 20\% intervals and show that the resulting model trained to minimise loss on this data set reaches a sharper minima than a model trained on the standard training data (0\% randomised data). The training setup matches that of the ResNet18 baseline models, with a batch size 256 and a learning rate 0.001, and results are averaged over 5 seeds (0--4) for each condition.

The increased complexity of the learning task is also reflected in the training loss, which is higher for the randomised-data conditions despite all models being trained under the same setup, as shown in Table~\ref{tab:random_data_experiment}. It is important to note that all models nevertheless achieve 100\% training accuracy.
\begin{figure}[htb]
    \centering
    \subfigure[Train Accuracy]{\includegraphics[width=0.32\linewidth] {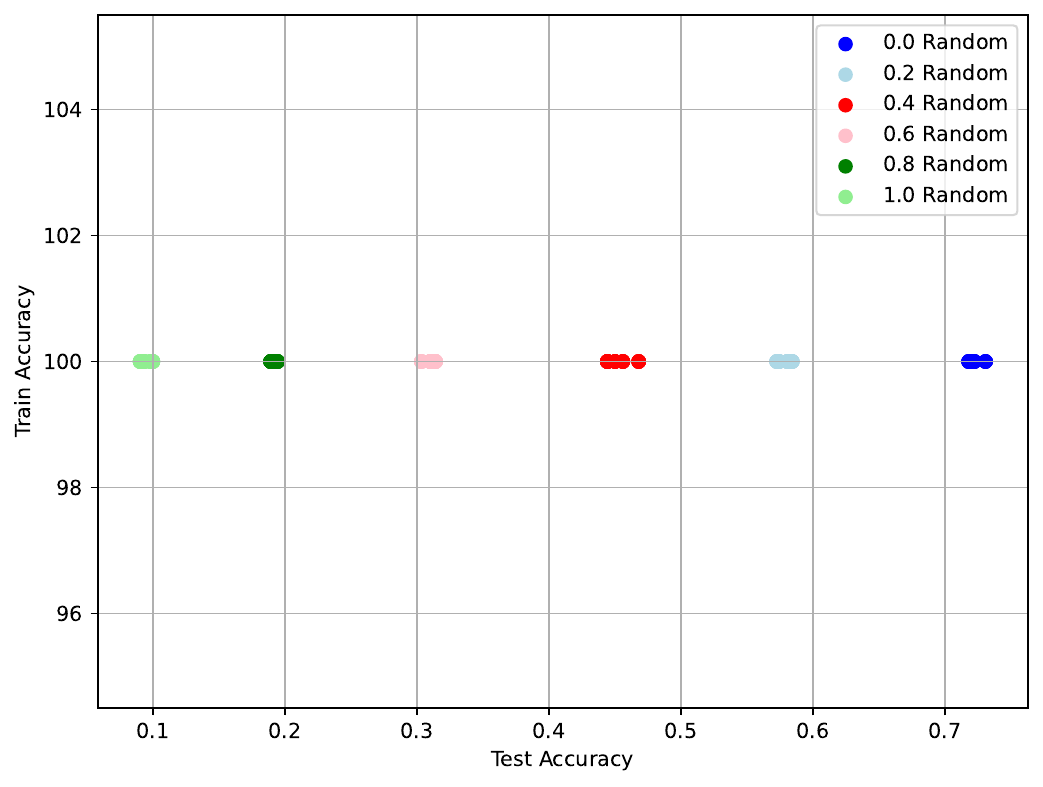}}
    \subfigure[Fisher Rao norm]{\includegraphics[width=0.32\linewidth] {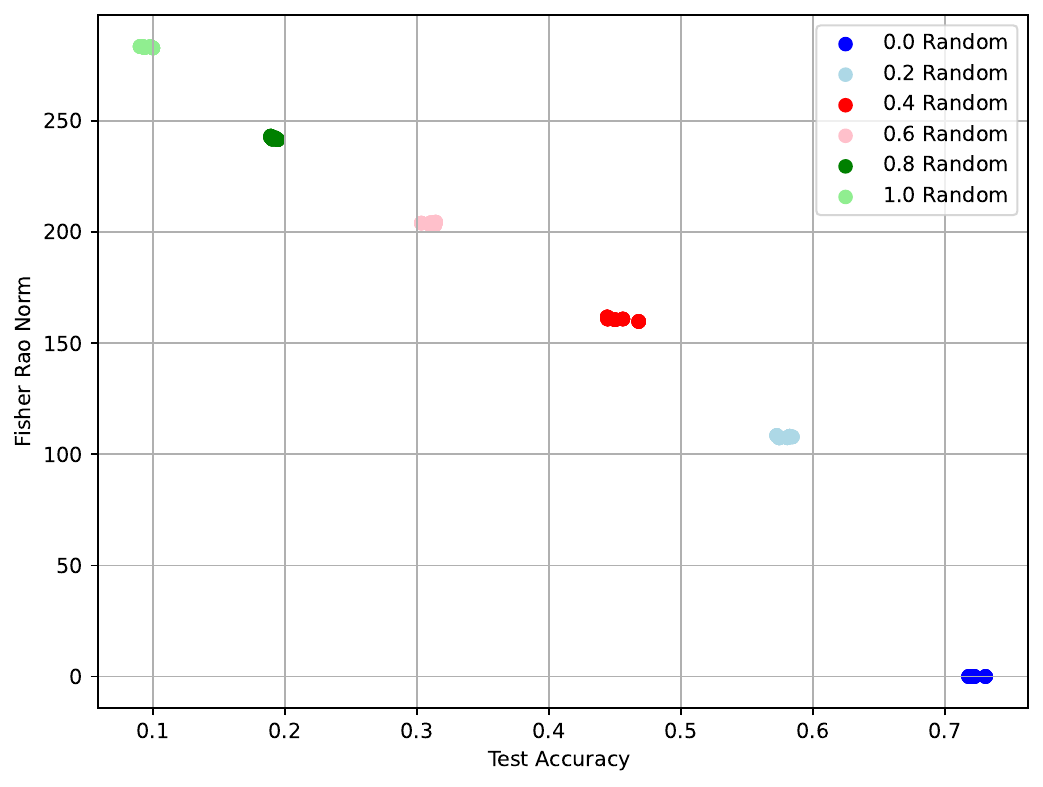}}
    \subfigure[Relative Flatness]{\includegraphics[width=0.32\linewidth] {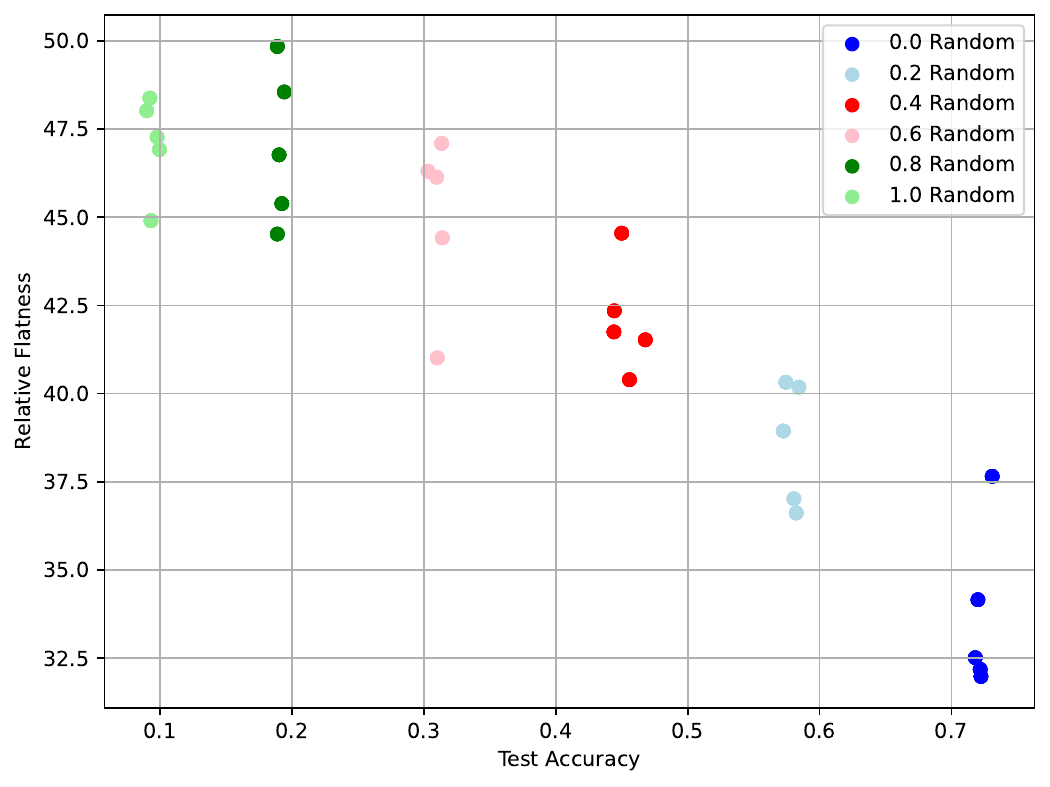}}
    \caption{Increasing the portion of random training data on CIFAR10 with Resnet18 from 0.0 to 1.0: plotting train accuracy and sharpness metrics against test accuracy.}
   \label{fig:random_experiemnt}
\end{figure}
\begin{table}[H]
\centering
\resizebox{\textwidth}{!}{
\begin{tabular}{|c|c|c|c|c|c|}
\hline
\textbf{\begin{tabular}[c]{@{}c@{}}Percentage of\\ Random Data\end{tabular}} & \textbf{\begin{tabular}[c]{@{}c@{}}Train \\ Accuracy\end{tabular}} & \textbf{\begin{tabular}[c]{@{}c@{}}Train \\ Loss\end{tabular}} & \textbf{\begin{tabular}[c]{@{}c@{}}SAM \\ Sharpness\end{tabular}} & \textbf{\begin{tabular}[c]{@{}c@{}}Fisher-Rao \\ norm\end{tabular}} & \textbf{\begin{tabular}[c]{@{}c@{}}Relative \\ Flatness\end{tabular}} \\ \hline
0\% (Baseline)                                                      & 1.000 $\pm{0.000}$                                                      & 9.997E-05 $\pm{1.381E-06}$                                          & 1.344E-05 $\pm{1.968E-06}$                                             & 0.031 $\pm{0.001}$                                                       & 33.700 $\pm{0.949}$                                                        \\ \hline
20\%                                                                & 1.000 $\pm{0.000}$                                                      & 1.039E-04 $\pm{1.714E-06}$                                          & 2.594E-01 $\pm{2.349E-03}$                                             & 107.830 $\pm{0.183}$                                                     & 38.615 $\pm{0.693}$                                                        \\ \hline
40\%                                                                & 1.000 $\pm{0.000}$                                                      & 1.071E-04 $\pm{1.759E-06}$                                          & 4.180E-01 $\pm{2.405E-02}$                                             & 160.742 $\pm{0.292}$                                                     & 42.111 $\pm{0.613}$                                                        \\ \hline
60\%                                                                & 1.000 $\pm{0.000}$                                                      & 1.085E-04 $\pm{2.294E-06}$                                          & 8.927E-01 $\pm{1.848E-02}$                                             & 203.816 $\pm{0.212}$                                                     & 44.990 $\pm{0.972}$                                                        \\ \hline
80\%                                                                & 1.000 $\pm{0.000}$                                                      & 1.107E-04 $\pm{2.535E-06}$                                          & 4.984E-01 $\pm{3.913E-02}$                                             & 242.238 $\pm{0.260}$                                                     & 47.011 $\pm{0.878}$                                                        \\ \hline
100\%                                                               & 1.000 $\pm{0.000}$                                                      & 1.094E-04 $\pm{1.490E-06}$                                          & 3.450E-01 $\pm{3.319E-02}$                                             & 283.181 $\pm{0.129}$                                                     & 47.095 $\pm{0.543}$                                                        \\ \hline
\end{tabular}}
\caption{Results for ResNet-18 trained on CIFAR10 with \textbf{increasingly randomised data}. For sharpness metrics lower values represent flatter models.}
\label{tab:random_data_experiment}
\end{table}

The results in Table~\ref{tab:random_data_experiment} and Figure~\ref{fig:random_experiemnt} show that as the percentage of randomised examples increases, and the class samples are increasingly disjoint through increased randomisation, 
there is a strong increase in the sharpness of the resulting minima (under Fisher-Rao norm and Relative Flatness) found at the end of training. In this setting, the experiment can be interpreted as simulating a function of increasing complexity, insofar as the class structure becomes progressively more disjoint and the effective decision-boundary geometry becomes more complex (irregular and/or tightly constrained).   
While the learned function in this case may not be useful from the standpoint of generalisation, the results suggest that more complex relationships in the training data can lead to sharper minima at the end of training, directly associating function complexity and minima geometry. 

While we increase complexity in this experiment by increasing the number of disjoint examples the model must fit for each class, the regularisation experiments suggest there exist other sources of increased complexity in learning. These include, for example, a larger number of classes, or greater variability within each class, as introduced by augmentation. Across these scenarios, we observe that the model can navigate to sharper regions of the loss landscape in order to fit more complex data structure, but, crucially these tighter decision boundaries are harder for the model to fit in the same training budget. 

\section{Conclusion}
Our paper revisits the relationship between geometry and generalisation in deep learning, extending it to include reliability-related evaluations such as calibration, robustness to corruptions, and functional consistency, and the interaction of regularisation techniques. Rather than focusing solely on accuracy, we study how sharpness relates to broader (reliability) properties of model behaviour and optimisation settings. Across diverse architectures and data sets, we find that standard training controls such as weight decay, data augmentation, and SAM often lead to sharper solutions that also achieve stronger performance on reliability-related metrics. These results call into question the conventional assumption that flatter minima are inherently preferable, and instead support a function-centric view in which sharper minima can correspond to more constrained, well-generalising functions. We further reconcile SAM's behaviour by  that it promotes local robustness rather than global flatness, helping to explain why improved generalisation can coincide with increased sharpness. More broadly, our findings suggest that sharpness is not universally harmful; rather its meaning depends on the learned function, the inductive bias of the model, and the training control used to obtain the solution. 
We posit that the geometry of learned solutions is shaped by task-specific demands, including the need to represent functions of greater complexity or tighter decision-boundary structure, which we demonstrate in toy-settings and on high-dimensional tasks. 
Overall, this work calls for a re-evaluation of geometric intuitions in deep learning, and underscores the importance of connecting training controls, solution geometry, and functional reliability.

\section{Limitations}
We survey a large range of architectures, single-objective optimisation problems, binary classification problems and high-dimensional learning tasks using a range of hyperparameters, which shows a strong relationship between regularisation, increased sharpness and improved generalisation and reliability performance. However, we do not explore how specific optimisers impact learning beyond the settings considered here. As a result, it would be interesting to explore how our results may change under optimisers such as Adam~\citep{kingma2017adammethodstochasticoptimization}, AdamW~\citep{loshchilov2018decoupled} and RAdam \citep{Liu2020On}. Another extension of this work would be to develop a more formal operationalisation of function complexity. Function complexity can be considered in many ways, some of which might be easy to formalise and others do not yet have sufficient mathematical formalisms. For example, the number of classes to be modelled can represent a form of complexity. Typically, as the number of classes present in a dataset increases a neural network requires more capacity to represent these distinct classes in a separated manner. Additionally, the number of disjoint samples in a dataset can increase the complexity of the learning task; this can be understood through augmentation, which increases the number of disjoint features across samples, meaning that a model within the same computational budget takes longer to minimise loss compared to a non-augmented dataset. Other forms of feature complexity can arise from the proximity of decision boundaries regardless of the number of classes, which we observe in the case of synthetic tasks and data memorisation in deep neural networks. These examples only represent a couple of dimensions from which function complexity can arise. As a result, to precisely define all of these forms of complexity is challenging, and is an interesting and important direction for future work. Finally, the conclusion supported by our results is that sharpness is not a universal indicator of poor generalisation under the settings and measures considered here. The causal relationship between training control, learned function, and minima geometry remains unresolved.
\section{Acknowledgements}
\textbf{1}: Calculations were performed using the Sulis Tier 2 HPC platform hosted by the Scientific Computing Research Technology Platform at the University of Warwick. Sulis is funded by EPSRC Grant EP/T022108/1 and the HPC Midlands+ consortium. \textbf{2}:  This research utilised Queen Mary's Apocrita HPC facility, supported by QMUL Research-IT \citep{qmul_hpc}. \textbf{3}: Calculations were performed using the King’s College London HPC \citep{kingshpc}. \textbf{4}: This work was supported by UK Research and Innovation grant number EP/S023356/1, in the UKRI Centre for Doctoral Training in Safe and Trusted Artificial Intelligence\footnote{UKRI Centre for Doctoral Training in Safe and Trusted Artificial: \url{www.safeandtrustedai.org}.}.

\bibliography{main}
\bibliographystyle{tmlr}

\appendix

\section{Single-Objective Optimisation Functions}
\label{app:Single-Objective-Optimisation-Functions}
The Sphere, Rosenbrock, Rastrigin, Beale, Booth, Three Hump Camel and the Himmelblaus functions are defined in equations \ref{eq:sphere} to \ref{eq:himmelblaus} respectively.
\begin{equation}
    f({\boldsymbol {x, y}})=(x^2 + y^2)
    \label{eq:sphere}
\end{equation}

\begin{equation}
    {\displaystyle f({\boldsymbol {x,y}})= (a - x)^2 + b(y - x^2)^2, \, \text{where}\, a=1\, \text{and}\, b=100}
    \label{eq:rosenbrock}
\end{equation}

\begin{equation}
    {\displaystyle f(\boldsymbol {x,y} )=2a + x^2 - a\cos(2x\pi) + y^2 - a\cos(2y\pi), \, \text{where}\, a=10}
    \label{eq:rastrigin}
\end{equation}

\begin{equation}
    {\displaystyle f(\boldsymbol {x,y} )=(1.5 - x + xy)^2 + (2.25 - x + xy^2)^2 + (2.625 - x + xy^3)^2}
    \label{eq:beale}
\end{equation}

\begin{equation}
    {\displaystyle f(\boldsymbol {x,y} )= (x + 2y -7)^2 + (2x + y - 5)^2}
    \label{eq:booth}
\end{equation}

\begin{equation}
    {\displaystyle f(\boldsymbol {x,y} )= 2x^2 - 1.05x^4 + \frac{x^6}{6} + xy + y^2}
    \label{eq:thee-hump-camel}
\end{equation}

\begin{equation}
    {\displaystyle f(\boldsymbol {x,y} )= (x^2 + y - 11)^2 + (x + y^2 - 7)^2}
    \label{eq:himmelblaus}
\end{equation}

\subsection{Training Details}
We trained a 3 layer ReLU MLP with an input width of two, a hidden width of 64 and output width of 1 with the Adam Optimizer and a learning rate of 1e-3. The train and test data set consisted of 10,000 input pairs $(X,Y)$ generated by independently sampling $X$ and $Y$ from a uniform distribution~$\mathcal{U}(-3.5, 3.5)$. For each of the seven functions, every input pair was evaluated using that specific function, yielding a target output $T$ for each function such that $F(X,Y)=T$. This procedure resulted in seven distinct data sets with identical input distributions but unique output transformations determined by their respective functions allowing for a clear assessment and comparison of the model’s capacity to learn each target function under controlled input conditions.  For Figure \ref{fig:single-objective-optimization} the model was trained 10 times with the same initialisation with ten different data sets for the respective function for $10^6$ epochs, where the mean SAM sharpness based on the training data and train and test loss where recorded for initialisation and epochs $10^0, 10^1, 10^2, 10^3, 10^4, 10^5, 10^6$. For Figures \ref{fig:target_loss_300}-\ref{fig:target_loss_1}, the model was trained 10 times with the same initialisation with ten different data sets for the respective function until the model reached the training target loss of $300, 150, 100, 10,$ and $1$. For the Beale function, the model was unable to achieve a train loss of than 150 and lower within $10^6$ epochs, and for Rosenbrock the model was unable to achieve a train loss of 100 and lower within $10^6$ epochs. 

\subsection{Training to Equivalent Loss}
\label{sec:equlivaent_loss}

\begin{figure}[H]
    \centering
    \includegraphics[width=\linewidth]{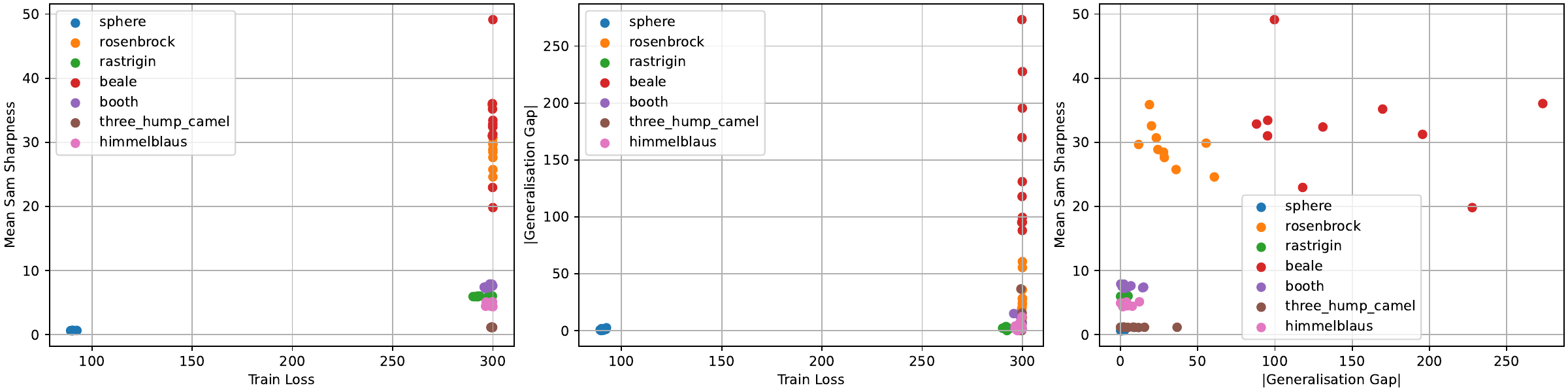}
    \caption{Scatter plots an MLP trained on the sphere, rosenbrock, rastrigin, beale, booth, three-hump camel, and himmelblaus functions for 10 different data sets till reaching a target train loss of 300: (left) mean SAM sharpness vs. train loss, (centre) $|$ generalisation gap $|$ vs. train loss, and (right) $|$generalisation gap$|$ vs. mean SAM sharpness.}
    \label{fig:target_loss_300}
\end{figure}

\begin{figure}[H]
    \centering
    \includegraphics[width=\linewidth]{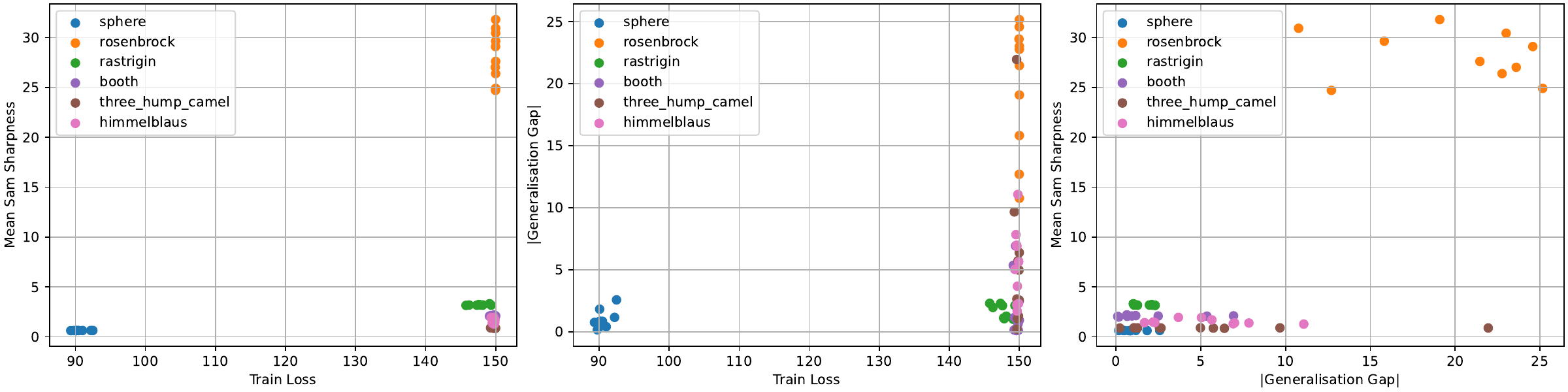}
    \caption{Scatter plots an MLP trained on the sphere, rosenbrock, rastrigin, booth, three-hump camel, and himmelblaus functions for 10 different data sets till reaching a target train loss of 150: (left) mean SAM sharpness vs. train loss, (centre) $|$ generalisation gap $|$ vs. train loss, and (right) $|$generalisation gap$|$ vs. mean SAM sharpness.}
    \label{fig:target_loss_150}
\end{figure}

\begin{figure}[H]
    \centering
    \includegraphics[width=\linewidth]{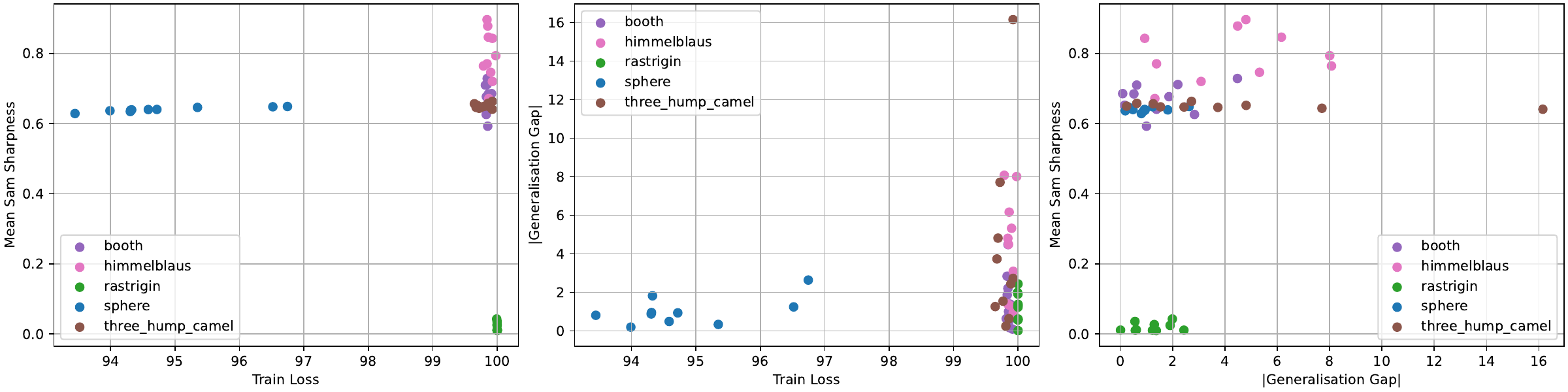}
    \caption{Scatter plots an MLP trained on the sphere, rastrigin, booth, three-hump camel, and himmelblaus functions for 10 different data sets till reaching a target train loss of 100: (left) mean SAM sharpness vs. train loss, (centre) $|$ generalisation gap $|$ vs. train loss, and (right) $|$generalisation gap$|$ vs. mean SAM sharpness.}
    \label{fig:target_loss_100}
\end{figure}

\begin{figure}[H]
    \centering
    \includegraphics[width=\linewidth]{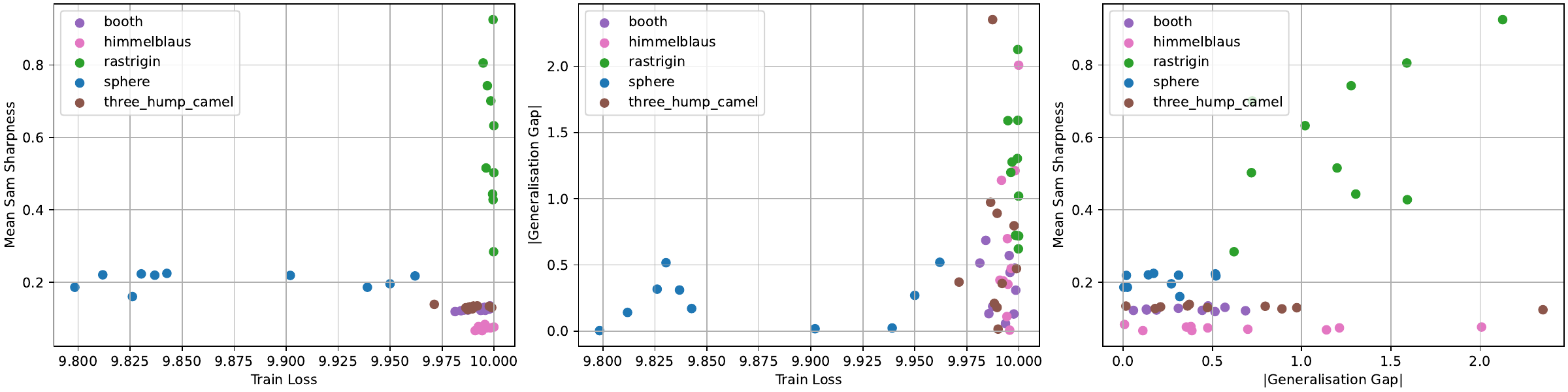}
    \caption{Scatter plots an MLP trained on the sphere, booth, three-hump camel, and himmelblaus functions for 10 different data sets till reaching a target train loss of 10: (left) mean SAM sharpness vs. train loss, (centre) $|$ generalisation gap $|$ vs. train loss, and (right) $|$generalisation gap$|$ vs. mean SAM sharpness.}
    \label{fig:target_loss_10}
\end{figure}

\begin{figure}[H]
    \centering
    \includegraphics[width=\linewidth]{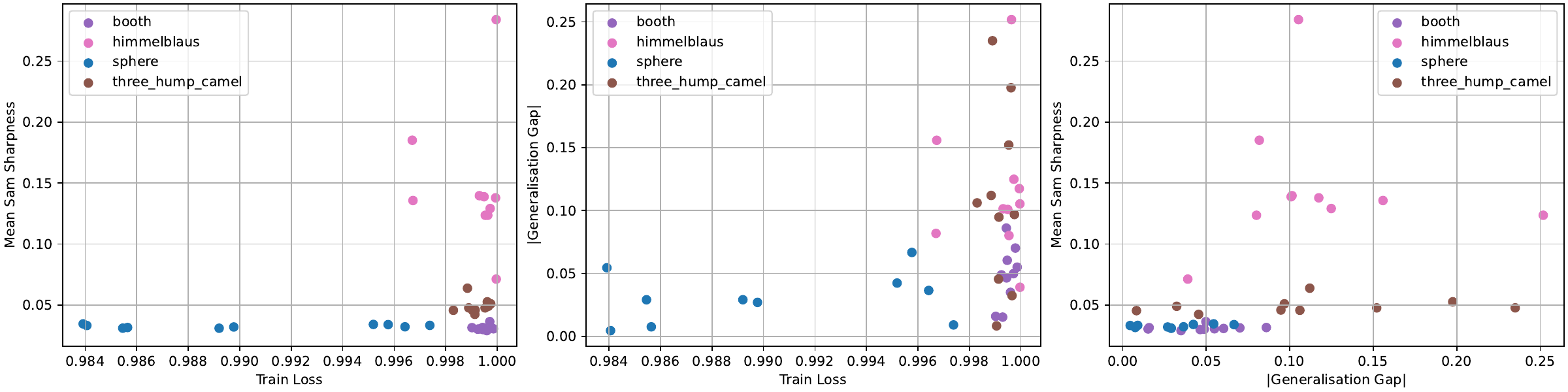}
    \caption{Scatter plots an MLP trained on the sphere, booth, three-hump camel, and himmelblaus functions for 10 different data sets till reaching a target train loss of 1: (left) mean SAM sharpness vs. train loss, (centre) $|$ generalisation gap $|$ vs. train loss, and (right) $|$generalisation gap$|$ vs. mean SAM sharpness.}
    \label{fig:target_loss_1}
\end{figure}

As the model achieves different final losses after training for $[10^0, 10^1, 10^2, 10^3, 10^4, 10^5,$ $ 10^6]$ epochs, we control for training duration by fixing a target train loss. We then investigate how reaching an approximate target train loss influences model sharpness and the generalisation gap. To illustrate these findings, we compare the mean SAM sharpness achieved at a train loss of 300 (Figure \ref{fig:target_loss_300}) and observe clear patterns. The model trained on the Rosenbrock and Beale tasks has sharpness values between 20 and 50. In contrast, when trained on Rastrigin, Booth, and Himmelblau tasks, sharpness values range between 5 and 10, while the Sphere and Three-Hump Camel tasks produce the flattest results. At this fixed loss, Figure \ref{fig:target_loss_300}(centre) highlights that the model trained on different tasks shows varied generalisation gaps. Notably, Figure \ref{fig:target_loss_300}(right) reveals several tasks (Sphere, Rastrigin, Booth, Three-Hump Camel, Himmelblaus) yielding similar generalisation gaps but differing sharpness values. Interestingly, the Rosenbrock task produces significantly higher sharpness while overlapping in generalisation gap with the Three-Hump Camel task. These observations collectively underscore that sharpness reflects the learning task rather than dictating model generalisation. Recognising that measuring at a train loss of 300 is arbitrary, we also examine target losses of 150, 100, 10, and 1 (Figures \ref{fig:target_loss_150}-\ref{fig:target_loss_1}). Across these settings, we again find that the models can achieve similar train losses but differ in sharpness depending on the learned function, which further supports our initial claims.


Notably, some functions drop off as we reach particular target losses. This happens because functions with more complicated landscapes, such as Beale and Rosenbrock, cannot exceed a train loss of 150 MSE. Less complex functions, such as the Sphere, can surpass this threshold. This supports our understanding that function complexity and solution geometry impact how easily a function can be fit. Less complex functions are more easily fit and tend to record lower sharpness values than complex functions, even when they achieve the same relative loss and generalisation gaps. Therefore, to improve optimisation outcomes for complex functions, it may be necessary to design or adopt inductive biases specifically tailored to their landscapes, rather than relying solely on traditional initialisation strategies. 

\section{Experimental Settings}
\label{app:settings}

All models are trained using NVIDIA A100 GPU's and each sharpness metric is calculated using the same GPU setup - as models output layer becomes larger for transitions between CIFAR10, CIFAR100 and TinyImageNet the computational cost of the calculation of sharpness metrics increases (by an order of magnitude  between CIFAR10 and CIFAR100). It should be noted that while Fisher-Rao norm is computationally inexpensive to calculate, SAM sharpness takes a factor of time longer and Relative Flatness is the most computationally expensive measure from a time and memory perspective. All models are trained such that they converge on the training data set or approximately converge in the case of augmentation conditions - meaning that they achieve 100\% training accuracy. It is important to note that all models (for the computer vision experiments) are given \textbf{100 epochs to reduce loss on the training} set to make comparisons fair. As a result, the test error is appropriate for assessing the generalisation gap as a high test accuracy is indicative of a small generalisation gap. 

\label{sec:Training_Settings}
\textit{CIFAR10 Training:} To train the \textbf{baseline} architectures on the CIFAR10 data set we use the following settings: SGD with the momentum hyperparameter at 0.9 to minimise cross entropy loss for 100 epochs, using a batch size of 256 and a learning rate of 0.001. For all architectures in the \textbf{SAM condition} we use the same settings as above but with SAM an extra optimization step occurs. We use SAM with the hyperparameter $\rho$ at the standard value of 0.05. For the \textbf{augmentation condition} we use the baseline conditions with the augmentations Random Crop with a padding of 4 and a fill of 128 alongside a Random Horizontal Flip with a probability of 0.5. Finally for the \textbf{weight decay condition} we use the same setup as the baseline condition but with the addition of the weight decay value set at $5e^{-4}$.

\textit{CIFAR10 Sharpness:} For all sharpness metrics on CIFAR10 we used the entire training data set to calculate sharpness across Fisher-Rao norm, SAM sharpness and Relative Flatness. For the augmentation condition, the training data set is the augmentation data used to train the model. We show in Sections~\ref{sec:resnet_sharpness-equivalence} and~\ref{sec:vgg_sharpness-equivalence} that calculating sharpness on the augmented training data set for the models in the augmentation condition is approximately equivalent to calculating with the original training data set without augmentation, thus preserving the same trends of increased sharpness for models trained with augmentation.

\textit{CIFAR100 Training:} To train the \textbf{baseline} architectures on the CIFAR100 data set we use the following settings: SGD with the momentum hyperparameter at 0.9 to minimise cross entropy loss for 100 epochs, using a batch size of 256 a learning rate of $1e^{-2}$, we also use a Pytorch's~\citep{NEURIPS2019_bdbca288} Cosine Annealing learning rate scheduler with a Maximum number of iterations of 100. For all architectures in the \textbf{SAM condition} we use the same settings as above but with SAM as an extra optimization step occurs and for this we use SAM with the hyperparameter $\rho$ at the standard value of 0.05. For the \textbf{augmentation condition} we use the baseline conditions with the augmentations Random Crop with a padding of 4 and a fill of 128 alongside a Random Horizontal Flip with a probability of 0.5. Finally for the \textbf{weight decay condition} we use the same setup as the baseline condition but with the addition of the weight decay value set at $5e^{-4}$.

\textit{CIFAR100 Sharpness:} For both the Fisher-Rao norm and SAM sharpness metrics on CIFAR100 we used the entire training data set to calculate sharpness. However, due to the computational burden of calculating Relative Flatness, we only employ 20\% of the training data set to calculate sharpness for this metric. Once again, for the augmentation condition, the training data set is the augmentations data used to train the model.

\textit{TinyImageNet Training:} On the TinyImageNet data set we use use pre-trained weights provided for the ResNet18\footnote{Pytorch ResNet18 ImageNet1K Pretrained Model:~\url{https://docs.pytorch.org/vision/main/models/generated/torchvision.models.resnet18.html}}and VGG19BN\footnote{Pytorch VGG19BN ImageNet1K Pretrained Model:~\url{https://docs.pytorch.org/vision/main/models/generated/torchvision.models.vgg19_bn.html}} by Pytorch - we modify these architectures by removing the existing final layer and replacing it with a final layer with a 200 output classification layer initialised with seeds 0-9 for the respective run. 
\\
\\
To train the \textbf{baseline} condition on these architectures we employ the following settings: SGD with the momentum hyperparameter at 0.9 to minimise cross entropy loss for 100 epochs, using a batch size of 256 and a learning rate of 0.001. For all architectures in the \textbf{SAM condition} we use the same settings as above but with SAM as an extra optimization step occurs and for this we use SAM with the hyperparameter $\rho$ at the standard value of 0.05. For the \textbf{augmentation condition} we use the baseline conditions with the augmentations Random Resized Crop to the size of 64 and a Random Horizontal Flip with a probability of 0.5. Finally, for the \textbf{weight decay condition} we use the same setup as the baseline condition but with the addition of the weight decay value set at $5e^{-4}$. For corruptions, the TinyImageNet-C data set\footnote{Official download link for the TinyImageNet-C data set:~\url{https://zenodo.org/records/2536630}} only consists of 182/200 test classes, as a result we created a TinyImageNet-C data set that contains all 200 classes using the methodology from~\cite{hendrycks2019benchmarking} to ensure the corruption accuracy results presented in the paper accurately estimate the variability of the models under corruption for the TinyImageNet data set. 

\textit{TinyImageNet Sharpness:} For the Fisher-Rao norm sharpness metric on TinyImageNet we used the entire training data set to calculate sharpness. However, due to the computational burden of calculating SAM sharpness, we only employ 20\% of the training data set to calculate sharpness for this metric. Additionally, due to memory constraints on the A100 GPU's we were unable to calculate Relative Flatness for any size of the training data set on this architecture. Once again, for the augmentation condition, the training data set is the augmentations data used to train the model.
\section{Augmented or Standard Training Data Sharpness Calculation?}

We argue that the standard data set is a subset of the augmented training data set, which leads to similar sharpness trends for both data sets. Our results indicate that calculating sharpness using augmented data produces nearly identical outcomes to those from the standard data set for Fisher-Rao norm, Sam Sharpness, Relative Flatness, and loss landscape visualisations. Specifically, we present these findings for the ResNet18 and VGG19 architectures in Subsections~\ref{sec:resnet_sharpness-equivalence} and~\ref{sec:vgg_sharpness-equivalence}.
\subsection{ResNet18}
\label{sec:resnet_sharpness-equivalence}

\textit{Loss Landscape Visualisations}.
Table~\ref{ResNet_aug_no_aug} shows that using augmented or standard training data has little effect on the resulting loss landscape visualisations. This confirms that calculating sharpness for models trained with augmented data is valid. The data set used does not qualitatively affect landscape sharpness or sharpness values, furthering our perspective that the standard data set can be considered a subset of the augmented data. Sharpness calculations depend more on model weights than on data and should represent any data set given identical weight permutations.

\begin{table}[H]
\centering
\begin{tabular}{|c|c|c|c|}
\hline
                     & \textbf{\begin{tabular}[c]{@{}c@{}}Augmentation  \\ Augmented Training Data\end{tabular} } & \textbf{\begin{tabular}[c]{@{}c@{}}Augmentation  \\  Standard Training Data\end{tabular}} \\ \hline
\textbf{\begin{tabular}[c]{@{}c@{}}Without \\ SAM\end{tabular}}                 & \includegraphics[width=0.35\textwidth]{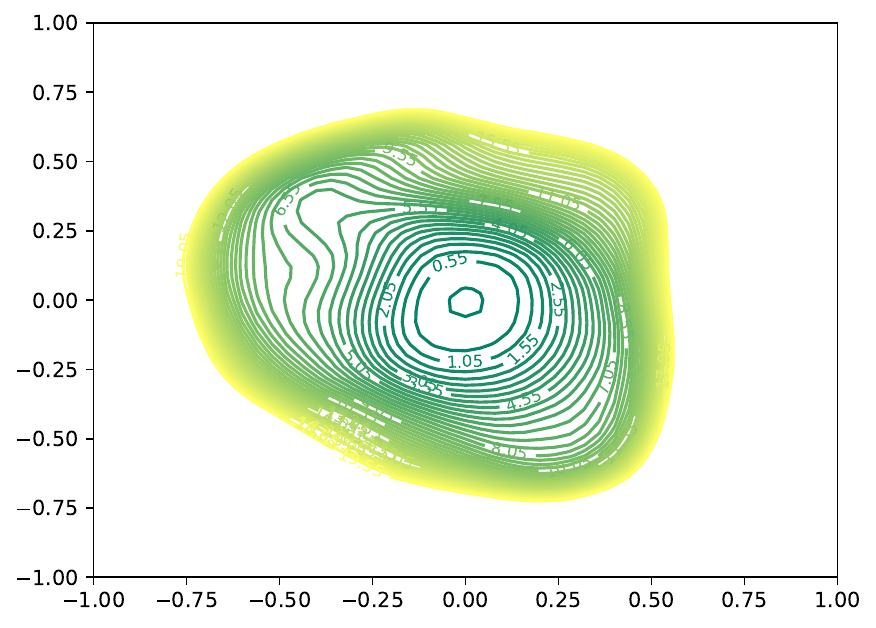}                          &  \includegraphics[width=0.35\textwidth]{figures/loss_landscapes/ResNet/ResNet18_aug.pth_weights_xignore=biasbn_xnorm=filter_yignore=biasbn_ynorm=filter.h5_-1.0,1.0,51x-1.0,1.0,51.h5_train_loss_2dcontour.pdf}                       \\ \hline
\textbf{\begin{tabular}[c]{@{}c@{}}With \\ SAM\end{tabular}}                     & \includegraphics[width=0.35\textwidth]{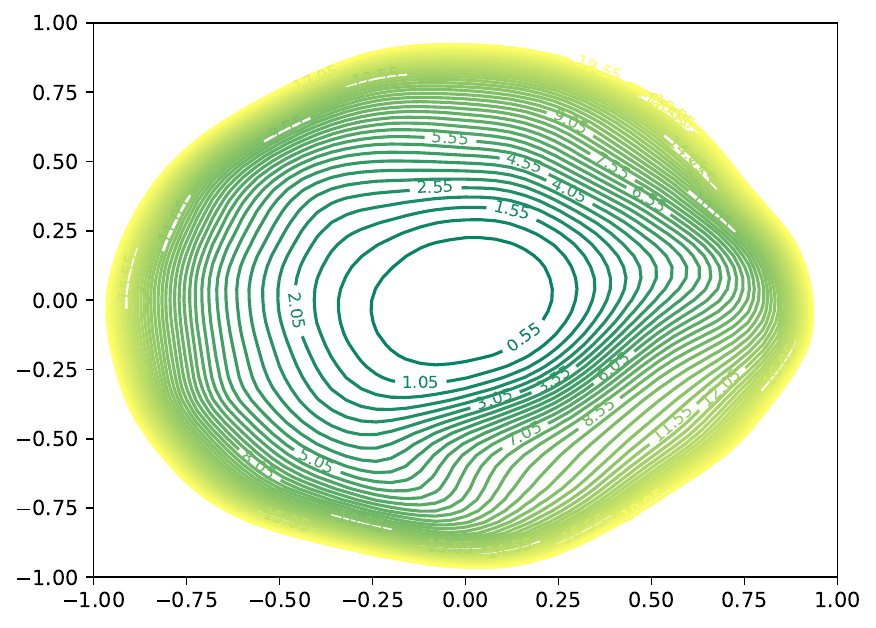}                        & \includegraphics[width=0.35\textwidth]{figures/loss_landscapes/ResNet/ResNet18_aug_SAM.pth_weights_xignore=biasbn_xnorm=filter_yignore=biasbn_ynorm=filter.h5_-1.0,1.0,51x-1.0,1.0,51.h5_train_loss_2dcontour.pdf}                          \\ \hline
\end{tabular}
\caption{Loss landscape visualisation \citep{li2018visualizing}of ResNet18 landscape on CIFAR10 exploring
the loss in the domain of the perturbations $[-1,1]^2$ with 51 steps in both directions on models trained with augmentation visualising landscape with standard training data and augmented training data.} 
\label{ResNet_aug_no_aug}
\end{table}
\textit{Sharpness Metrics}. When calculating the sharpness metrics, it can be seen that the difference between using augmented training data, in Table~\ref{tab:sharp_aug_training_data}, or standard training data, in Table~\ref{tab:sharp_standard_training_data}, for each of the metrics provides no difference for the trends of results observed. 
\begin{table}[H]
\centering
\begin{tabular}{|c|c|c|c|}
\hline
\textbf{\begin{tabular}[c]{@{}c@{}}Control \\ Condition\end{tabular}} & \textbf{\begin{tabular}[c]{@{}c@{}}SAM \\ Sharpness\end{tabular}} & \textbf{\begin{tabular}[c]{@{}c@{}}Fisher-Rao \\ norm\end{tabular}} & \textbf{\begin{tabular}[c]{@{}c@{}}Relative\\  Flatness\end{tabular}} \\ \hline
Augmentation                                                          & 1.905E-01 $\pm{2.203E-02}$                                                        & 3.940 $\pm{0.207}$                                             & 2903.220 $\pm{89.243}$                                                     \\ \hline
\begin{tabular}[c]{@{}c@{}}Augmentation\\ + SAM\end{tabular}          & 1.303E-01 $\pm{1.547E-02}$                                                      &  5.571 $\pm{0.035}$                                              & 4970.972 $\pm{30.139}$                                                     \\ \hline
\end{tabular}
\caption{Sharpness Calculation for ResNet18 landscape on CIFAR10 trained with batch size of 256 and learning rate of 0.001 using augmented training data for sharpness calculations.}
\label{tab:sharp_aug_training_data}
\end{table}
\begin{table}[H]
\centering
\begin{tabular}{|c|c|c|c|}
\hline
\textbf{\begin{tabular}[c]{@{}c@{}}Control \\ Condition\end{tabular}} & \textbf{\begin{tabular}[c]{@{}c@{}}SAM \\ Sharpness\end{tabular}}  & \textbf{\begin{tabular}[c]{@{}c@{}}Fisher-Rao \\ norm\end{tabular}} & \textbf{\begin{tabular}[c]{@{}c@{}}Relative \\ Flatness\end{tabular}} \\ \hline
Augmentation                                                          & 1.591E-02 $\pm{1.609E-03}$                                                        &  3.962 $\pm{0.292}$                                            & 2972.554 $\pm{137.079}$                                                    \\ \hline
\begin{tabular}[c]{@{}c@{}}Augmentation\\ + SAM\end{tabular}          & 2.035E-02 $\pm{1.203E-03}$                                                     & 5.084 $\pm{0.032}$                                              & 5105.327 $\pm{43.058}$                                                     \\ \hline
\end{tabular}
\caption{Sharpness Calculation for ResNet18 landscape on CIFAR10 trained with batch size of 256 and learning rate of 0.001 using standard training data for sharpness calculations.}
\label{tab:sharp_standard_training_data}
\end{table}

\subsection{VGG19}
\label{sec:vgg_sharpness-equivalence}
\textit{Loss Landscape Visualisations}. In Table~\ref{VGG_aug_no_aug}, we show that the use of augmented or standard training data has little impact on the resulting loss landscape visualisations. This reaffirms that it is valid to calculate sharpness for models using augmented training data. The data set used does not qualitatively impact the sharpness of the landscape or the resulting sharpness values. 

\begin{table}[H]
\centering
\begin{tabular}{|c|c|c|c|}
\hline
                     & \textbf{\begin{tabular}[c]{@{}c@{}}Augmentation  \\ Augmented Training Data\end{tabular} } & \textbf{\begin{tabular}[c]{@{}c@{}}Augmentation  \\  Standard Training Data\end{tabular}} \\ \hline
\textbf{\begin{tabular}[c]{@{}c@{}}Without \\ SAM\end{tabular}}                 & \includegraphics[width=0.35\textwidth]{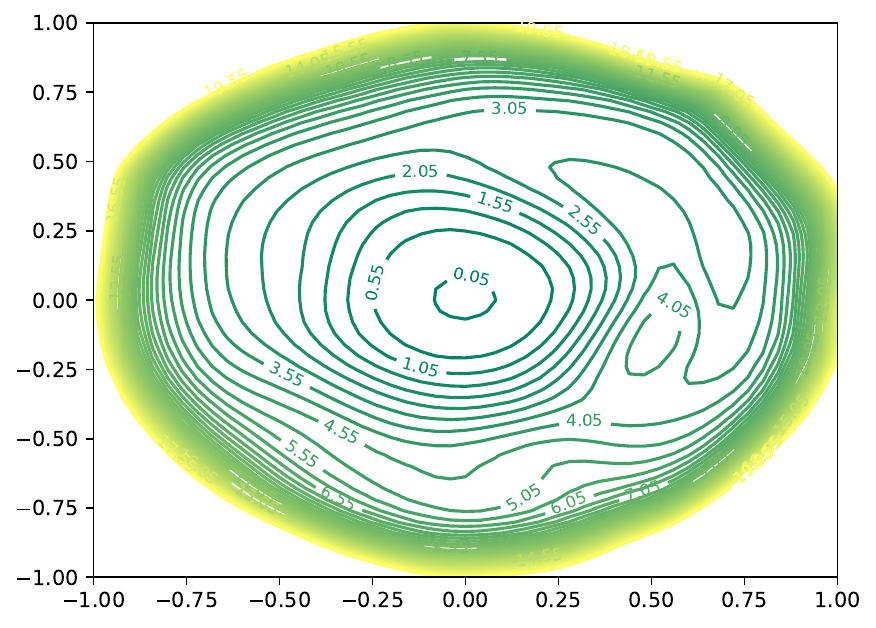}                          &  \includegraphics[width=0.35\textwidth]{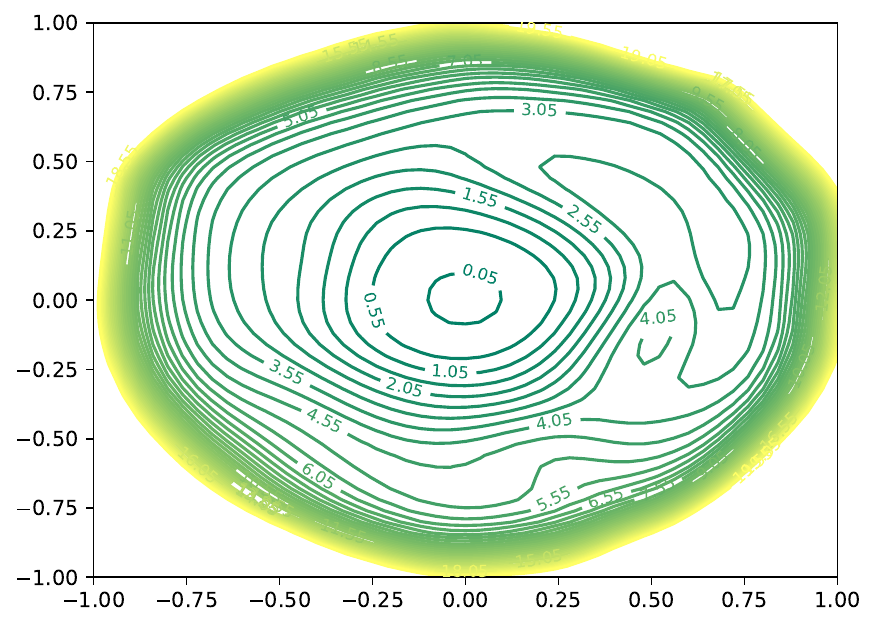}                       \\ \hline
\textbf{\begin{tabular}[c]{@{}c@{}}With \\ SAM\end{tabular}}                     & \includegraphics[width=0.35\textwidth]{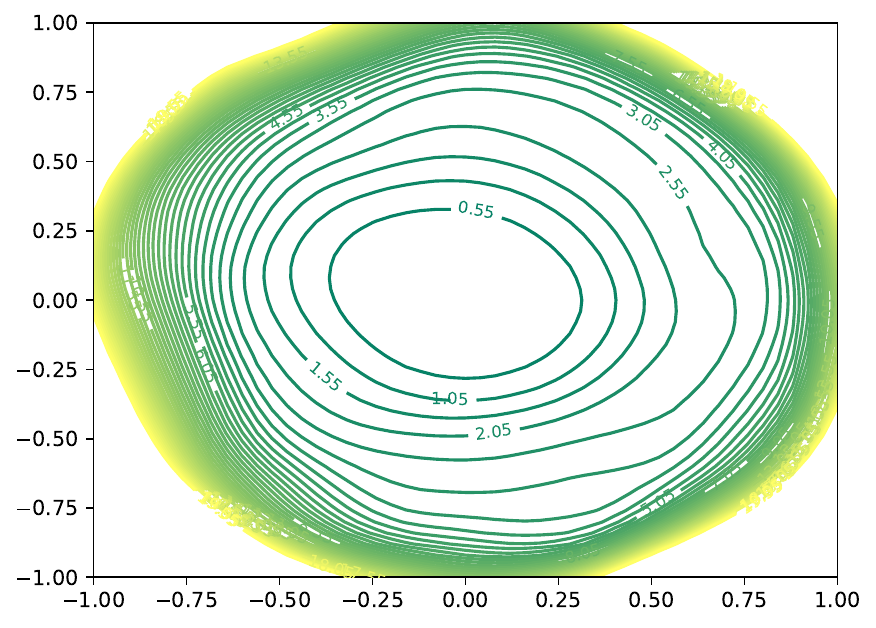}                        & \includegraphics[width=0.35\textwidth]{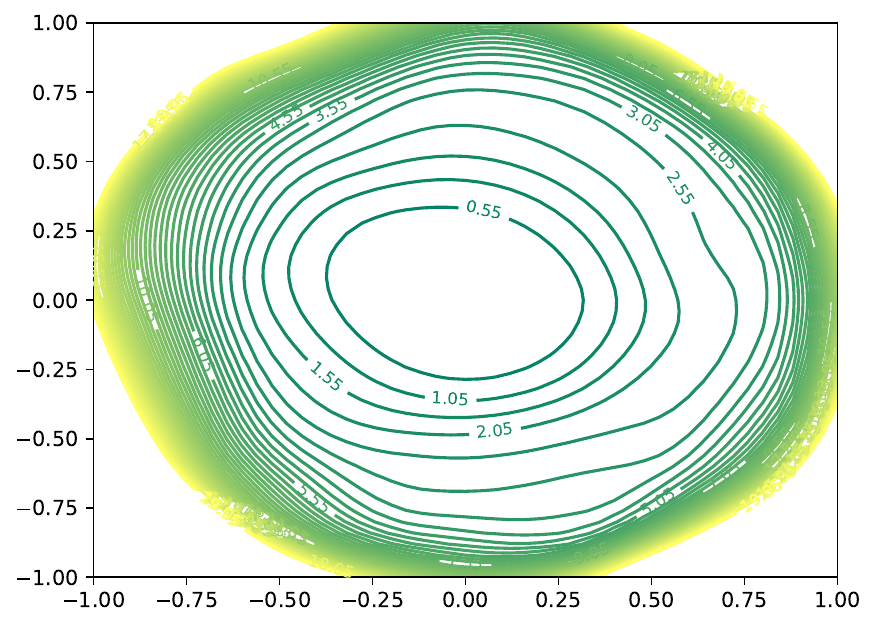}                          \\ \hline
\end{tabular}
\caption{Loss landscape visualisation \citep{li2018visualizing} of VGG19 landscape on CIFAR10 exploring the loss in the domain of the perturbations  $[-1,1]^2$ with 51 steps in both directions on models trained with augmentation visualising landscape with standard training data and augmented training data.}
\label{VGG_aug_no_aug}
\end{table}
\textit{Sharpness Metrics}. When calculating the sharpness metrics, comparing the results from augmented training data (Table~\ref{tab:vgg_sharp_aug_training_data}) and standard training data (Table~\ref{tab:vgg_sharp_standard_training_data}) shows that both present consistent trends. In both cases, training with augmentation or augmentation plus SAM leads to a minima that is substantially sharper than the baseline model, as evidenced by the values in the respective tables. 

\begin{table}[H]
\centering
\begin{tabular}{|c|c|c|c|}
\hline
\textbf{\begin{tabular}[c]{@{}c@{}}Control \\ Condition\end{tabular}} & \textbf{\begin{tabular}[c]{@{}c@{}}SAM \\ Sharpness\end{tabular}} & \textbf{\begin{tabular}[c]{@{}c@{}}Fisher-Rao \\ norm\end{tabular}}  & \textbf{\begin{tabular}[c]{@{}c@{}}Relative \\ Flatness\end{tabular}} \\ \hline
Augmentation                                                          & 1.967E-01 $\pm{}$2.298E-02                                                      & 3.505 $\pm{}$0.155                                               & 688.897 $\pm{}$26.348                                                      \\ \hline
\begin{tabular}[c]{@{}c@{}}Augmentation\\ + SAM\end{tabular}          &  9.777E-02 $\pm{}$1.126E-02                                                      & 4.278 $\pm{}$0.027                                              & 1609.212 $\pm{}$22.719                                                     \\ \hline

\end{tabular}
\caption{Sharpness Calculation for VGG19 on CIFAR10 trained with batch size of 256 and learning rate of 0.001 using augmented training data for sharpness calculations.}
\label{tab:vgg_sharp_aug_training_data}
\end{table}
\begin{table}[H]
\centering
\begin{tabular}{|c|c|c|c|}
\hline
\textbf{\begin{tabular}[c]{@{}c@{}}Control\\ Condition\end{tabular}} & \textbf{\begin{tabular}[c]{@{}c@{}}SAM \\ Sharpness\end{tabular}} &  \textbf{\begin{tabular}[c]{@{}c@{}}Fisher-Rao \\ norm\end{tabular}}  & \textbf{\begin{tabular}[c]{@{}c@{}}Relative \\ Flatness\end{tabular}} \\ \hline
Augmentation                                                         & 9.589E-03 $\pm{}$1.438E-03                                                       & 2.322 $\pm{}$0.125                                              & 481.312 $\pm{}$23.358                                                      \\ \hline
\begin{tabular}[c]{@{}c@{}}Augmentation \\ + SAM\end{tabular}        &  1.497E-02 $\pm{}$7.363E-04                                                      & 3.756 $\pm{}$0.030                                              & 1413.712 $\pm{}$22.212                                                     \\ \hline
\end{tabular}
\caption{Sharpness Calculation for VGG19 on CIFAR10 trained with batch size of 256 and learning rate of 0.001 using standard training data for sharpness calculations.}
\label{tab:vgg_sharp_standard_training_data}
\end{table}

\section{ResNet-18 Further Results}
Here we observe how two core hyperparameters, batch size and learning rate, impact the general finding that models under the use of training regularisation navigate to sharper points and thus tighter decision  boundaries than baseline models without their application. In line with the findings in the main paper, we observe that the best-performing models in each condition are those that are sharper than the baseline models for each respective experiment.
\subsection{ResNet-18 Batch Size and Learning Rate Hyperparameter Sweep}
\label{sec:ResNet_param_sweep}

We observe that baseline sharpness values are impacted by a change in learning rate and batch size. For example, a learning rate of $1e^{-2}$ results in far flatter baseline models for both batch sizes. The flatness observed in these cases does correspond to better-performing baseline models. However, we still observe that in all cases when regularisation is applied, the models become sharper than the respective baseline, and it still remains that these regularised models that are sharper outperform the baseline models across all hyperparameter settings. 

\begin{table}[H]
\centering
\resizebox{\textwidth}{!}{
\begin{tabular}{|c|c|c|c|c|c|c|c|c|}
\hline
\textbf{\begin{tabular}[c]{@{}c@{}}Control\\ Condition\end{tabular}} & \textbf{\begin{tabular}[c]{@{}c@{}}Generalisation \\ Gap\end{tabular}} & \textbf{\begin{tabular}[c]{@{}c@{}}Test \\ Accuracy\end{tabular}} & \textbf{\begin{tabular}[c]{@{}c@{}}Test \\ ECE\end{tabular}} & \textbf{\begin{tabular}[c]{@{}c@{}}Corruption \\ Accuracy\end{tabular}} & \textbf{\begin{tabular}[c]{@{}c@{}}Prediction \\ Disagreement\end{tabular}} & \textbf{\begin{tabular}[c]{@{}c@{}}SAM \\ Sharpness\end{tabular}} & \textbf{\begin{tabular}[c]{@{}c@{}}Fisher-Rao \\ norm\end{tabular}} & \textbf{\begin{tabular}[c]{@{}c@{}}Relative \\ Flatness\end{tabular}} \\ \hline
Baseline                                                             & 28.050 $\pm{0.175}$                                                         & 0.720 $\pm{0.002}$                                                     & 0.186 $\pm{0.001}$                                                & 58.614 $\pm{0.201}$                                                          & 0.282 $\pm{0.001}$                                                               & 1.366E-05 $\pm{1.206E-06}$                                             & 0.032 $\pm{0.001}$                                                       & 34.607 $\pm{0.757}$                                                        \\ \hline
\begin{tabular}[c]{@{}c@{}}Baseline\\ + SAM\end{tabular}             & 20.588 $\pm{0.125}$                                                         & 0.794 $\pm{0.001}$                                                     & 0.108 $\pm{0.001}$                                                & 66.342 $\pm{0.164}$                                                          & 0.168 $\pm{0.000}$                                                               & 5.823E-05 $\pm{9.056E-06}$                                             & 0.107 $\pm{0.006}$                                                       & 75.093 $\pm{1.693}$                                                        \\ \hline
Augmentation                                                         & 10.399 $\pm{0.067}$                                                         & 0.886 $\pm{0.001}$                                                     & 0.077 $\pm{0.001}$                                                & 68.755 $\pm{0.219}$                                                          & 0.121 $\pm{0.001}$                                                               & 1.905E-01 $\pm{2.203E-02}$                                             & 3.940 $\pm{0.207}$                                                       & 2903.220 $\pm{89.243}$                                                     \\ \hline
\begin{tabular}[c]{@{}c@{}}Augmentation\\ + SAM\end{tabular}         & \textbf{6.864 $\pm{0.038}$}                                                 & \textbf{0.908 $\pm{0.000}$}                                            & \textbf{0.014 $\pm{0.001}$}                                       & \textbf{71.419 $\pm{0.283}$}                                                 & \textbf{0.069 $\pm{0.000}$}                                                      & 1.303E-01 $\pm{1.547E-02}$                                             & 5.571 $\pm{0.035}$                                                       & 4970.972 $\pm{30.139}$                                                     \\ \hline
Weight Decay                                                         & 27.942 $\pm{0.196}$                                                         & 0.721 $\pm{0.002}$                                                     & 0.174 $\pm{0.002}$                                                & 58.562 $\pm{0.227}$                                                          & 0.281 $\pm{0.001}$                                                               & 3.391E-05 $\pm{4.494E-06}$                                             & 0.065 $\pm{0.004}$                                                       & 59.767 $\pm{3.009}$                                                        \\ \hline
\begin{tabular}[c]{@{}c@{}}Weight Decay\\ + SAM\end{tabular}                                                   & 19.788 $\pm{0.149}$                                                         & 0.802 $\pm{0.001}$                                                     & 0.096 $\pm{0.001}$                                                & 67.079 $\pm{0.117}$                                                          & 0.162 $\pm{0.001}$                                                               & 8.733E-05 $\pm{1.430E-05}$                                             & 0.127 $\pm{0.006}$                                                       & 88.807 $\pm{2.336}$                                                        \\ \hline
\end{tabular}}
\caption{Results for ResNet-18 trained on CIFAR10 with \textbf{batch size 256 and a learning rate of $\mathbf{1e^{-3}}$}. Numbers in bold indicate best scores for metrics. For sharpness metrics lower values represent flatter models.}
\end{table}

\begin{table}[H]
\centering
\resizebox{\textwidth}{!}{
\begin{tabular}{|c|c|c|c|c|c|c|c|c|}
\hline
\textbf{\begin{tabular}[c]{@{}c@{}}Control\\ Condition\end{tabular}} & \textbf{\begin{tabular}[c]{@{}c@{}}Generalisation \\ Gap\end{tabular}} & \textbf{\begin{tabular}[c]{@{}c@{}}Test \\ Accuracy\end{tabular}} & \textbf{\begin{tabular}[c]{@{}c@{}}Test \\ ECE\end{tabular}} & \textbf{\begin{tabular}[c]{@{}c@{}}Corruption \\ Accuracy\end{tabular}} & \textbf{\begin{tabular}[c]{@{}c@{}}Prediction \\ Disagreement\end{tabular}} & \textbf{\begin{tabular}[c]{@{}c@{}}SAM \\ Sharpness\end{tabular}} & \textbf{\begin{tabular}[c]{@{}c@{}}Fisher-Rao \\ norm\end{tabular}} & \textbf{\begin{tabular}[c]{@{}c@{}}Relative \\ Flatness\end{tabular}} \\ \hline
Baseline                                                             & 16.203 $\pm{0.266}$                                                         & 0.838 $\pm{0.003}$                                                     & 0.109 $\pm{0.004}$                                                & 70.814 $\pm{0.390}$                                                          & 0.138 $\pm{0.001}$                                                               & 7.648E-06 $\pm{3.794E-06}$                                             & 0.015 $\pm{0.006}$                                                       & 16.641 $\pm{5.960}$                                                        \\ \hline
\begin{tabular}[c]{@{}c@{}}Baseline\\ + SAM\end{tabular}             & 14.549 $\pm{0.059}$                                                         & 0.855 $\pm{0.001}$                                                     & 0.084 $\pm{0.001}$                                                & 72.618 $\pm{0.161}$                                                          & 0.110 $\pm{0.000}$                                                               & 2.019E-05 $\pm{2.683E-06}$                                             & 0.042 $\pm{0.002}$                                                       & 49.022 $\pm{1.901}$                                                        \\ \hline
Augmentation                                                         & 7.593 $\pm{0.092}$                                                          & 0.921 $\pm{0.001}$                                                     & 0.056 $\pm{0.003}$                                                & 72.923 $\pm{0.223}$                                                          & 0.078 $\pm{0.001}$                                                               & 1.091E-01 $\pm{1.958E-02}$                                             & 2.390 $\pm{0.268}$                                                       & 1604.778 $\pm{103.972}$                                                    \\ \hline
\begin{tabular}[c]{@{}c@{}}Augmentation\\ + SAM\end{tabular}         & \textbf{6.920 $\pm{0.056}$}                                                 & \textbf{0.931 $\pm{0.001}$}                                            & \textbf{0.037 $\pm{0.001}$}                                       & \textbf{73.483 $\pm{0.212}$}                                                 & \textbf{0.058 $\pm{0.000}$}                                                      & 2.267E-02 $\pm{2.070E-03}$                                             & 1.165 $\pm{0.014}$                                                       & 1173.090 $\pm{15.607}$                                                     \\ \hline
Weight Decay                                                         & 16.791 $\pm{0.122}$                                                         & 0.832 $\pm{0.001}$                                                     & 0.071 $\pm{0.001}$                                                & 68.538 $\pm{0.159}$                                                          & 0.157 $\pm{0.000}$                                                               & 3.673E-05 $\pm{4.662E-06}$                                             & 0.097 $\pm{0.001}$                                                       & 99.041 $\pm{0.736}$                                                        \\ \hline
\begin{tabular}[c]{@{}c@{}}Weight Decay\\ + SAM\end{tabular}         & 14.022 $\pm{0.089}$                                                         & 0.860 $\pm{0.001}$                                                     & 0.050 $\pm{0.001}$                                                & 73.100 $\pm{0.137}$                                                          & 0.116 $\pm{0.000}$                                                               & 3.238E-04 $\pm{6.384E-05}$                                             & 0.446 $\pm{0.013}$                                                       & 178.169 $\pm{4.103}$                                                       \\ \hline
\end{tabular}}
\caption{Results for ResNet-18 trained on CIFAR10 with \textbf{batch size 256 and a learning rate of $\mathbf{1e^{-2}}$}. Numbers in bold indicate best scores for metrics. For sharpness metrics lower values represent flatter models.}
\label{tab:c10_resnet18_256_0.01}
\end{table}

\begin{table}[H]
\resizebox{\textwidth}{!}{
\begin{tabular}{|c|c|c|c|c|c|c|c|c|}
\hline
\textbf{\begin{tabular}[c]{@{}c@{}}Control\\ Condition\end{tabular}} & \textbf{\begin{tabular}[c]{@{}c@{}}Generalisation \\ Gap\end{tabular}} & \textbf{\begin{tabular}[c]{@{}c@{}}Test \\ Accuracy\end{tabular}} & \textbf{\begin{tabular}[c]{@{}c@{}}Test \\ ECE\end{tabular}} & \textbf{\begin{tabular}[c]{@{}c@{}}Corruption \\ Accuracy\end{tabular}} & \textbf{\begin{tabular}[c]{@{}c@{}}Prediction\\ Disagreement\end{tabular}} & \textbf{\begin{tabular}[c]{@{}c@{}}SAM \\ Sharpness\end{tabular}} & \textbf{\begin{tabular}[c]{@{}c@{}}Fisher-Rao \\ norm\end{tabular}} & \textbf{\begin{tabular}[c]{@{}c@{}}Relative\\ Flatness\end{tabular}} \\ \hline
\begin{tabular}[c]{@{}c@{}}Baseline \\ + SAM\end{tabular}            & \cmark~(0.96) & \cmark~(0.96) & \cmark~(0.93) & \cmark~(0.96) & \cmark~(0.99) & \cmark~(0.75) & \cmark~(0.96) & \cmark~(1.0) \\ \hline
Augmentation                                                         & \cmark~(1.0)  & \cmark~(1.0)  & \cmark~(1.0)  & \cmark~(0.90) & \cmark~(1.0)                & \cmark~(1.0)  & \cmark~(1.0)  & \cmark~(1.0) \\ \hline
\begin{tabular}[c]{@{}c@{}}Augmentation  \\ + SAM\end{tabular}       & \cmark~(1.0)  & \cmark~(1.0)  & \cmark~(1.0)  & \cmark~(0.96) & \cmark~(1.0)                & \cmark~(1.0)  & \cmark~(1.0)  & \cmark~(1.0) \\ \hline
Weight Decay                                                         & \xmark~(N/A)   & \xmark~(N/A)  & \cmark~(1.0)  & \xmark~(N/A)  & \xmark~(N/A)                 & \cmark~(0.96) & \cmark~(1.0)  & \cmark~(1.0) \\ \hline
\begin{tabular}[c]{@{}c@{}}Weight Decay \\ + SAM\end{tabular}        & \cmark~(1.0)  & \cmark~(1.0)  & \cmark~(1.0)  & \cmark~(0.96) & \cmark~(0.99)               & \cmark~(1.0)  & \cmark~(1.0)  & \cmark~(1.0) \\ \hline
\end{tabular}}
\caption{Significance results for the control conditions against the baseline for ResNet18 trained on CIFAR10 with \textbf{batch size 256 and a learning rate of $\mathbf{1e^{-2}}$}.~\cmark~indicates significant difference compared to the baseline; \xmark~indicates significance, the value in the brackets represents the  effect sizes using the absolute RBC.}
\label{tab:sig-256-1e2-c10}
\end{table}

\begin{table}[H]
\resizebox{\textwidth}{!}{
\begin{tabular}{|c|c|c|c|c|c|c|c|c|}
\hline
\textbf{\begin{tabular}[c]{@{}c@{}}Control\\ Condition\end{tabular}} & \textbf{\begin{tabular}[c]{@{}c@{}}Generalisation \\ Gap\end{tabular}} & \textbf{\begin{tabular}[c]{@{}c@{}}Test \\ Accuracy\end{tabular}} & \textbf{\begin{tabular}[c]{@{}c@{}}Test \\ ECE\end{tabular}} & \textbf{\begin{tabular}[c]{@{}c@{}}Corruption \\ Accuracy\end{tabular}} & \textbf{\begin{tabular}[c]{@{}c@{}}Prediction \\ Disagreement\end{tabular}} & \textbf{\begin{tabular}[c]{@{}c@{}}SAM \\ Sharpness\end{tabular}} & \textbf{\begin{tabular}[c]{@{}c@{}}Fisher-Rao \\ norm\end{tabular}} & \textbf{\begin{tabular}[c]{@{}c@{}}Relative \\ Flatness\end{tabular}} \\ \hline
Baseline                                                             & 23.325 $\pm{0.140}$                                                         & 0.767 $\pm{0.001}$                                                     & 0.154 $\pm{0.001}$                                                & 63.035 $\pm{0.204}$                                                          & 0.227 $\pm{0.000}$                                                               & 5.181E-06 $\pm{5.856E-07}$                                             & 0.013 $\pm{0.000}$                                                       & 27.916 $\pm{0.340}$                                                        \\ \hline
\begin{tabular}[c]{@{}c@{}}Baseline\\ + SAM\end{tabular}             & 16.714 $\pm{0.125}$                                                         & 0.833 $\pm{0.001}$                                                     & 0.083 $\pm{0.001}$                                                & 69.769 $\pm{0.108}$                                                          & 0.126 $\pm{0.000}$                                                               & 2.640E-05 $\pm{4.345E-06}$                                             & 0.072 $\pm{0.006}$                                                       & 139.589 $\pm{2.679}$                                                       \\ \hline
Augmentation                                                         & 9.110 $\pm{0.079}$                                                          & 0.905 $\pm{0.001}$                                                     & 0.065 $\pm{0.001}$                                                & 71.516 $\pm{0.308}$                                                          & 0.099 $\pm{0.000}$                                                               & 9.266E-02 $\pm{5.276E-03}$                                             & 2.465 $\pm{0.105}$                                                       & 3735.018 $\pm{173.247}$                                                    \\ \hline
\begin{tabular}[c]{@{}c@{}}Augmentation\\ + SAM\end{tabular}         & \textbf{6.869 $\pm{0.022}$}                                                 & \textbf{0.921 $\pm{0.000}$}                                            & \textbf{0.013 $\pm{0.000}$}                                       & \textbf{72.870 $\pm{0.207}$}                                                 & \textbf{0.058 $\pm{0.000}$}                                                      & 8.913E-02 $\pm{8.054E-03}$                                             & 4.070 $\pm{0.027}$                                                       & 7532.582 $\pm{69.191}$                                                     \\ \hline
Weight Decay                                                         & 23.504 $\pm{0.136}$                                                         & 0.765 $\pm{0.001}$                                                     & 0.137 $\pm{0.001}$                                                & 62.879 $\pm{0.214}$                                                          & 0.231 $\pm{0.000}$                                                               & 2.241E-05 $\pm{3.426E-06}$                                             & 0.047 $\pm{0.000}$                                                       & 80.599 $\pm{0.548}$                                                        \\ \hline
\begin{tabular}[c]{@{}c@{}}Weight Decay\\ + SAM\end{tabular}         & 16.433 $\pm{0.096}$                                                         & 0.836 $\pm{0.001}$                                                     & 0.072 $\pm{0.001}$                                                & 70.226 $\pm{0.158}$                                                          & 0.124 $\pm{0.000}$                                                               & 4.797E-05 $\pm{7.285E-06}$                                             & 0.110 $\pm{0.004}$                                                       & 194.034 $\pm{2.840}$                                                       \\ \hline
\end{tabular}}
\caption{Results for ResNet-18 trained on CIFAR10 with \textbf{batch size 128 and a learning rate of $\mathbf{1e^{-3}}$}. Numbers in bold indicate best scores for metrics. For sharpness metrics lower values represent flatter models.}
\label{tab:128-1e3-c10}
\end{table}

\begin{table}[H]
\centering
\resizebox{\textwidth}{!}{
\begin{tabular}{|c|c|c|c|c|c|c|c|c|}
\hline
\textbf{\begin{tabular}[c]{@{}c@{}}Control\\ Condition\end{tabular}} & \textbf{\begin{tabular}[c]{@{}c@{}}Generalisation \\ Gap\end{tabular}} & \textbf{\begin{tabular}[c]{@{}c@{}}Test \\ Accuracy\end{tabular}} & \textbf{\begin{tabular}[c]{@{}c@{}}Test \\ ECE\end{tabular}} & \textbf{\begin{tabular}[c]{@{}c@{}}Corruption \\ Accuracy\end{tabular}} & \textbf{\begin{tabular}[c]{@{}c@{}}Prediction \\ Disagreement\end{tabular}} & \textbf{\begin{tabular}[c]{@{}c@{}}SAM \\ Sharpness\end{tabular}} & \textbf{\begin{tabular}[c]{@{}c@{}}Fisher-Rao \\ norm\end{tabular}} & \textbf{\begin{tabular}[c]{@{}c@{}}Relative \\ Flatness\end{tabular}} \\ \hline
\begin{tabular}[c]{@{}c@{}}Baseline \\ + SAM\end{tabular}            & \cmark~(1.0) & \cmark~(1.0) & \cmark~(1.0) & \cmark~(1.0) & \cmark~(1.0) & \cmark~(1.0) & \cmark~(1.0) & \cmark~(1.0) \\ \hline
Augmentation                                                         & \cmark~(1.0) & \cmark~(1.0) & \cmark~(1.0) & \cmark~(1.0) & \cmark~(1.0) & \cmark~(1.0) & \cmark~(1.0) & \cmark~(1.0) \\ \hline
\begin{tabular}[c]{@{}c@{}}Augmentation  \\ + SAM\end{tabular}       & \cmark~(1.0) & \cmark~(1.0) & \cmark~(1.0) & \cmark~(1.0) & \cmark~(1.0) & \cmark~(1.0) & \cmark~(1.0) & \cmark~(1.0) \\ \hline
Weight Decay                                                         & \xmark~(N/A)  & \xmark~(N/A) & \cmark~(1.0) & \xmark~(N/A) & \xmark~(N/A)  & \cmark~(1.0) & \cmark~(1.0) & \cmark~(1.0) \\ \hline
\begin{tabular}[c]{@{}c@{}}Weight Decay \\ + SAM\end{tabular}        & \cmark~(1.0) & \cmark~(1.0) & \cmark~(1.0) & \cmark~(1.0) & \cmark~(1.0) & \cmark~(1.0) & \cmark~(1.0) & \cmark~(1.0) \\ \hline
\end{tabular}}
\caption{Significance results for the control conditions against the baseline for ResNet18 trained on CIFAR10 with \textbf{batch size 128 and a learning rate of $\mathbf{1e^{-3}}$}.~\cmark~indicates significant difference compared to the baseline; \xmark~indicates significance, the value in the brackets represents the  effect sizes using the absolute RBC.}
\label{tab:sig-128-1e3-c10}
\end{table}

\begin{table}[H]
\centering
\resizebox{\textwidth}{!}{
\begin{tabular}{|c|c|c|c|c|c|c|c|c|}
\hline
\textbf{\begin{tabular}[c]{@{}c@{}}Control\\ Condition\end{tabular}} & \textbf{\begin{tabular}[c]{@{}c@{}}Generalisation \\ Gap\end{tabular}} & \textbf{\begin{tabular}[c]{@{}c@{}}Test \\ Accuracy\end{tabular}} & \textbf{\begin{tabular}[c]{@{}c@{}}Test \\ ECE\end{tabular}} & \textbf{\begin{tabular}[c]{@{}c@{}}Corruption \\ Accuracy\end{tabular}} & \textbf{\begin{tabular}[c]{@{}c@{}}Prediction \\ Disagreement\end{tabular}} & \textbf{\begin{tabular}[c]{@{}c@{}}SAM \\ Sharpness\end{tabular}} & \textbf{\begin{tabular}[c]{@{}c@{}}Fisher-Rao \\ norm\end{tabular}} & \textbf{\begin{tabular}[c]{@{}c@{}}Relative \\ Flatness\end{tabular}} \\ \hline
Baseline                                                             & 15.027 $\pm{0.069}$                                                         & 0.850 $\pm{0.001}$                                                     & 0.109 $\pm{0.001}$                                                & 71.960 $\pm{0.158}$                                                          & 0.125 $\pm{0.000}$                                                               & 1.094E-06 $\pm{9.618E-08}$                                             & 0.003 $\pm{0.000}$                                                       & 8.785 $\pm{0.142}$                                                         \\ \hline
\begin{tabular}[c]{@{}c@{}}Baseline\\ + SAM\end{tabular}             & 13.231 $\pm{0.065}$                                                         & 0.868 $\pm{0.001}$                                                     & 0.081 $\pm{0.001}$                                                & 73.053 $\pm{0.164}$                                                          & 0.099 $\pm{0.000}$                                                               & 1.021E-05 $\pm{5.519E-07}$                                             & 0.024 $\pm{0.001}$                                                       & 70.694 $\pm{2.273}$                                                        \\ \hline
Augmentation                                                         & 7.455 $\pm{0.062}$                                                          & 0.923 $\pm{0.001}$                                                     & 0.057 $\pm{0.001}$                                                & 72.594 $\pm{0.152}$                                                          & 0.076 $\pm{0.000}$                                                               & 8.274E-02 $\pm{6.784E-03}$                                             & 2.086 $\pm{0.140}$                                                       & 2864.657 $\pm{151.088}$                                                    \\ \hline
\begin{tabular}[c]{@{}c@{}}Augmentation\\ + SAM\end{tabular}         & \textbf{6.678 $\pm{0.060}$}                                                 & \textbf{0.933 $\pm{0.001}$}                                            & \textbf{0.036 $\pm{0.001}$}                                       & \textbf{73.565 $\pm{0.245}$}                                                 & \textbf{0.056 $\pm{0.000}$}                                                      & 2.173E-02 $\pm{2.042E-03}$                                             & 1.012 $\pm{0.014}$                                                       & 2354.005 $\pm{38.058}$                                                     \\ \hline
Weight Decay                                                         & 12.695 $\pm{0.072}$                                                         & 0.873 $\pm{0.001}$                                                     & 0.057 $\pm{0.001}$                                                & 70.979 $\pm{0.124}$                                                          & 0.103 $\pm{0.000}$                                                               & 1.265E-04 $\pm{3.140E-06}$                                             & 0.159 $\pm{0.003}$                                                       & 355.345 $\pm{12.866}$                                                      \\ \hline
\begin{tabular}[c]{@{}c@{}}Weight Decay\\ + SAM\end{tabular}         & 12.606 $\pm{0.069}$                                                         & 0.874 $\pm{0.001}$                                                     & 0.036 $\pm{0.001}$                                                & 72.795 $\pm{0.162}$                                                          & 0.107 $\pm{0.000}$                                                               & 5.880E-04 $\pm{6.127E-05}$                                             & 0.745 $\pm{0.017}$                                                       & 439.467 $\pm{8.439}$                                                       \\ \hline
\end{tabular}}
\caption{Results for ResNet-18 trained on CIFAR10 with \textbf{batch size 128 and a learning rate of $\mathbf{1e^{-2}}$}. Numbers in bold indicate best scores for metrics. For sharpness metrics lower values represent flatter models.}
\label{tab:c10_resnet18_128_0.01}
\end{table}

\begin{table}[H]
\resizebox{\textwidth}{!}{
\begin{tabular}{|c|c|c|c|c|c|c|c|c|}
\hline
\textbf{\begin{tabular}[c]{@{}c@{}}Control\\ Condition\end{tabular}} & \textbf{\begin{tabular}[c]{@{}c@{}}Generalisation \\ Gap\end{tabular}} & \textbf{\begin{tabular}[c]{@{}c@{}}Test \\ Accuracy\end{tabular}} & \textbf{\begin{tabular}[c]{@{}c@{}}Test \\ ECE\end{tabular}} & \textbf{\begin{tabular}[c]{@{}c@{}}Corruption \\ Accuracy\end{tabular}} & \textbf{\begin{tabular}[c]{@{}c@{}}Prediction \\ Disagreement\end{tabular}} & \textbf{\begin{tabular}[c]{@{}c@{}}SAM \\ Sharpness\end{tabular}} & \textbf{\begin{tabular}[c]{@{}c@{}}Fisher-Rao \\ norm\end{tabular}} & \textbf{\begin{tabular}[c]{@{}c@{}}Relative \\ Flatness\end{tabular}} \\ \hline
\begin{tabular}[c]{@{}c@{}}Baseline \\ + SAM\end{tabular}            & \cmark~(1.0) & \cmark~(1.0) & \cmark~(1.0) & \cmark~(0.96) & \cmark~(1.0) & \cmark~(1.0) & \cmark~(1.0) & \cmark~(1.0) \\ \hline
Augmentation                                                         & \cmark~(1.0) & \cmark~(1.0) & \cmark~(1.0) & \cmark~(0.82) & \cmark~(1.0) & \cmark~(1.0) & \cmark~(1.0) & \cmark~(1.0) \\ \hline
\begin{tabular}[c]{@{}c@{}}Augmentation  \\ + SAM\end{tabular}       & \cmark~(1.0) & \cmark~(1.0) & \cmark~(1.0) & \cmark~(0.96) & \cmark~(1.0) & \cmark~(1.0) & \cmark~(1.0) & \cmark~(1.0) \\ \hline
Weight Decay                                                         & \cmark~(1.0) & \cmark~(1.0) & \cmark~(1.0) & \xmark~(N/A)  & \cmark~(1.0) & \cmark~(1.0) & \cmark~(1.0) & \cmark~(1.0) \\ \hline
\begin{tabular}[c]{@{}c@{}}Weight Decay \\ + SAM\end{tabular}        & \cmark~(1.0) & \cmark~(1.0) & \cmark~(1.0) & \cmark~(0.93) & \cmark~(1.0) & \cmark~(1.0) & \cmark~(1.0) & \cmark~(1.0) \\ \hline
\end{tabular}}
\caption{Significance results for the control conditions against the baseline for ResNet18 trained on CIFAR10 with \textbf{batch size 128 and a learning rate of $\mathbf{1e^{-2}}$}.~\cmark~indicates significant difference compared to the baseline; \xmark~indicates significance, the value in the brackets represents the  effect sizes using the absolute RBC.}
\label{tab:sig-128-1e2-c10}
\end{table}

We observe three core insights across statistically supported results in Tables~\ref{tab:sig-256-1e2-c10},~\ref{tab:sig-128-1e3-c10}, and~\ref{tab:sig-128-1e2-c10}. \textbf{1}: improved generalisation performance can be found at sharper minima, \textbf{2}: improved reliability mostly coincides with sharper minima, and \textbf{3}: sharper minima often represent the best performing models. However, it should be noted that there are instances where weight decay increases the sharpness of minima against the baseline but does not provide improved reliability or generalisation performance. 

\section{VGG-19}
\label{app:VGG}
\subsection{VGG19 Batch Size and Learning Rate Hyperparameter Sweep}
\label{sec:vgg_param_sweep}
\begin{figure}[H]
    \centering
    \subfigure[Fisher-Rao norm and Loss]{\includegraphics[width=0.32\linewidth] {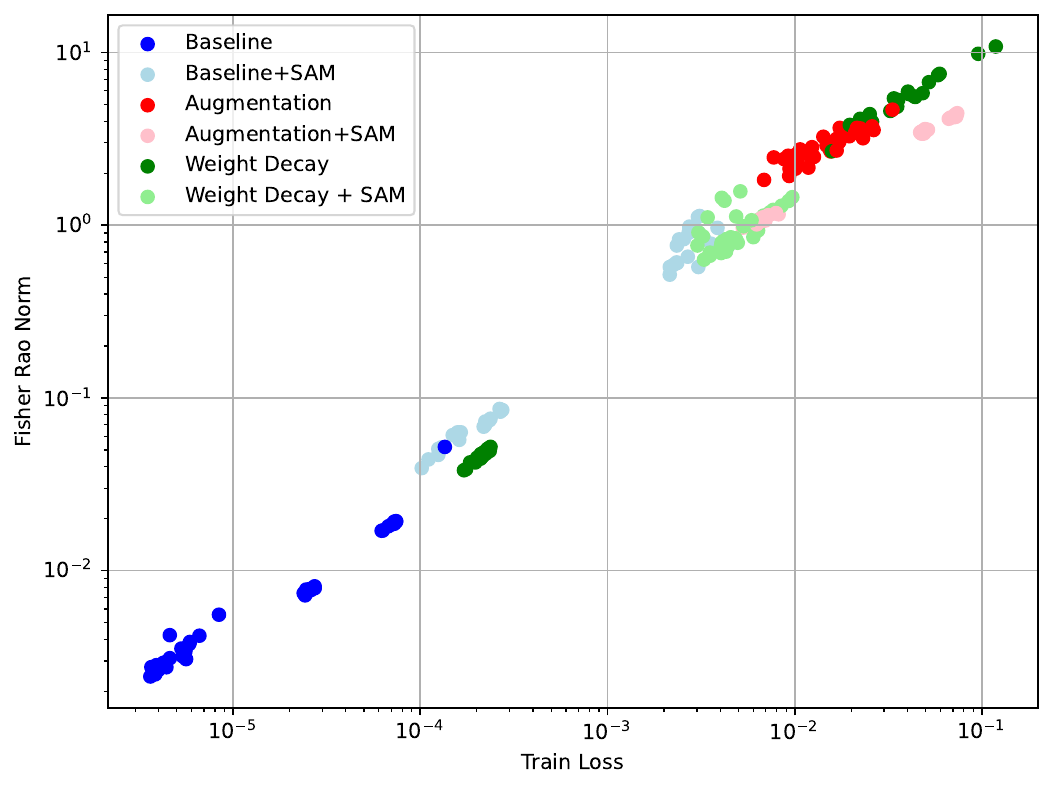}}
    \subfigure[Relative Flatness and Loss]{\includegraphics[width=0.315\linewidth] {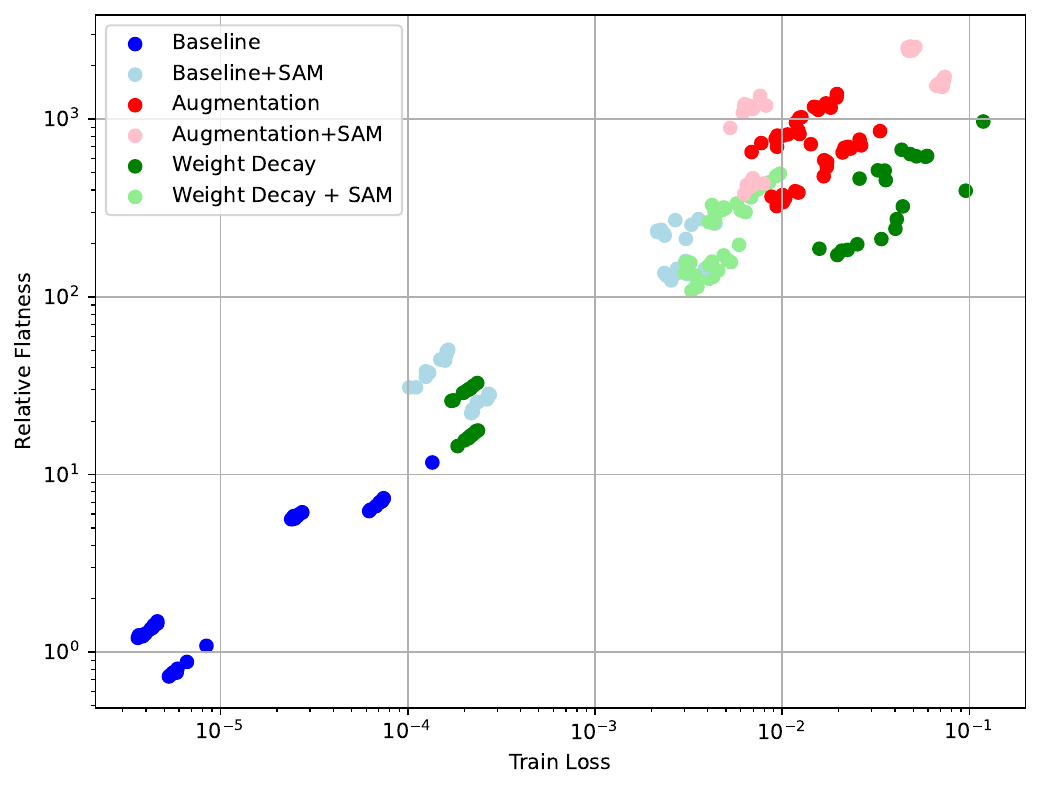}}
    \subfigure[Generalisation Gap and Loss]{\includegraphics[width=0.32\linewidth] {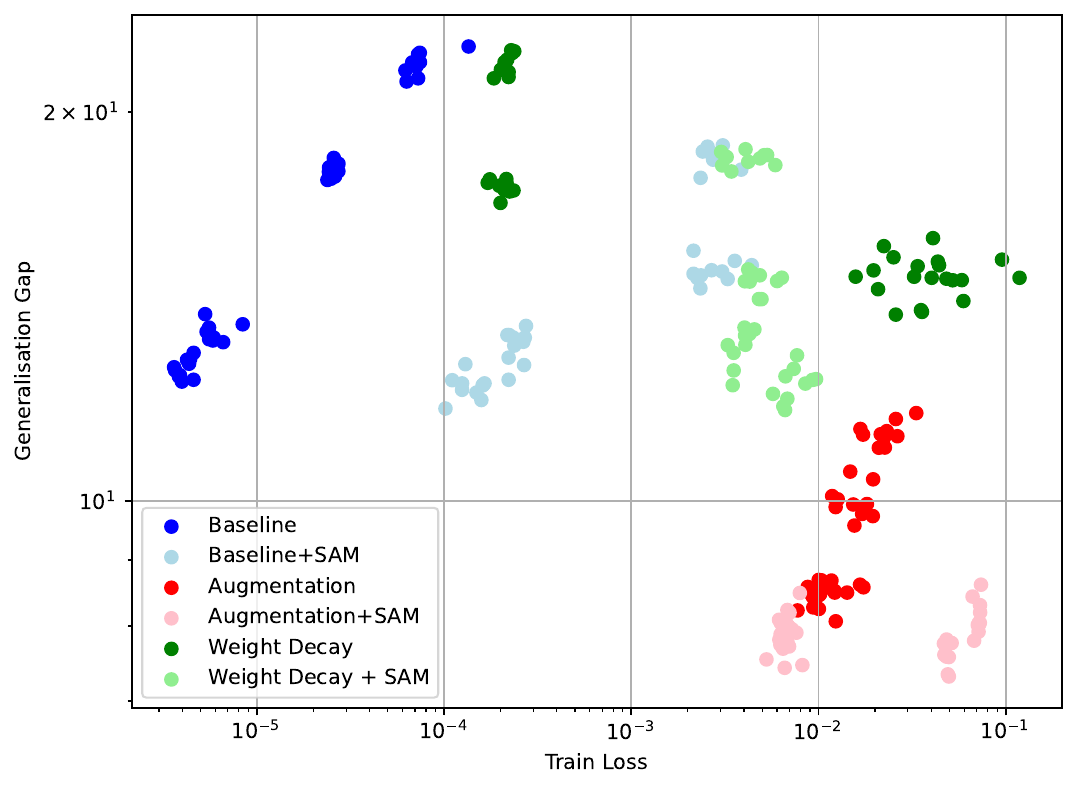}}
    \caption{Plot of 240 minima using reparametrisation invariant sharpness metrics against train loss and generalisation gap against train loss using log-scale for the VGG19 with different training hyperparameters (batch size of 256, 128 and learning rate of 0.001 and $1e^{-2}$) trained on CIFAR10.}
   \label{fig:sharpness_gen_gap_plot_vgg}
\end{figure}
Here we observe how two core hyperparameters, batch size and learning rate, impact the general finding that models using training regularisation navigate to sharper points and thus tighter decision boundaries than base models without regularisation. As found in the main paper, the best-performing models in each condition are those that are sharper than the baseline models for each experiment. However, changing the learning rate and batch size does influence sharpness values in each condition. A larger learning rate typically makes the minima flatter compared to a smaller learning rate. Still, these augmented models navigate to sharper landscapes than the baseline and achieve the best performance across generalisation and reliability evaluations.
\begin{table}[H]
\centering
\resizebox{\textwidth}{!}{
\begin{tabular}{|c|c|c|c|c|c|c|c|c|}
\hline
\textbf{\begin{tabular}[c]{@{}c@{}}Control\\ Condition\end{tabular}} & \textbf{\begin{tabular}[c]{@{}c@{}}Genralisation \\ Gap\end{tabular}} & \textbf{\begin{tabular}[c]{@{}c@{}}Test \\ Accuracy\end{tabular}} & \textbf{\begin{tabular}[c]{@{}c@{}}Test \\ ECE\end{tabular}} & \textbf{\begin{tabular}[c]{@{}c@{}}Corruption \\ Accuracy\end{tabular}} & \textbf{\begin{tabular}[c]{@{}c@{}}Prediction \\ Disagreement\end{tabular}} & \textbf{\begin{tabular}[c]{@{}c@{}}SAM \\ Sharpness\end{tabular}} & \textbf{\begin{tabular}[c]{@{}c@{}}Fisher-Rao \\ norm\end{tabular}} & \textbf{\begin{tabular}[c]{@{}c@{}}Relative \\ Flatness\end{tabular}} \\ \hline
Baseline                                                             & 21.805 $\pm{0.128}$                                                        & 0.782 $\pm{0.001}$                                                     & 0.160 $\pm{0.001}$                                                & 64.316 $\pm{0.193}$                                                          & 0.204 $\pm{0.000}$                                                               & 7.649E-06 $\pm{2.207E-06}$                                             & 0.022 $\pm{0.003}$                                                       & 7.374 $\pm{0.470}$                                                         \\ \hline
\begin{tabular}[c]{@{}c@{}}Baseline\\ + SAM\end{tabular}             & 18.444 $\pm{0.097}$                                                        & 0.815 $\pm{0.001}$                                                     & 0.108 $\pm{0.001}$                                                & 66.655 $\pm{0.296}$                                                          & 0.150 $\pm{0.000}$                                                               & 1.495E-03 $\pm{1.703E-04}$                                             & 0.938 $\pm{0.036}$                                                       & 140.164 $\pm{3.149}$                                                       \\ \hline
Augmentation                                                         & 11.289 $\pm{0.066}$                                                        & 0.879 $\pm{0.001}$                                                     & 0.084 $\pm{0.001}$                                                & 68.497 $\pm{0.199}$                                                          & 0.121 $\pm{0.000}$                                                               & 1.967E-01 $\pm{2.298E-02}$                                             & 3.505 $\pm{0.155}$                                                       & 688.897 $\pm{26.348}$                                                      \\ \hline
\begin{tabular}[c]{@{}c@{}}Augmentation\\ + SAM\end{tabular}         & \textbf{8.139 $\pm{0.074}$}                                                & \textbf{0.903 $\pm{0.001}$}                                            & \textbf{0.019 $\pm{0.001}$}                                       & \textbf{71.268 $\pm{0.196}$}                                                 & \textbf{0.075 $\pm{0.000}$}                                                      & 9.777E-02 $\pm{1.126E-02}$                                             & 4.278 $\pm{0.027}$                                                       & 1609.212 $\pm{22.719}$                                                     \\ \hline
Weight Decay                                                         & 21.801 $\pm{0.121}$                                                        & 0.782 $\pm{0.001}$                                                     & 0.151 $\pm{0.001}$                                                & 64.405 $\pm{0.217}$                                                          & 0.202 $\pm{0.000}$                                                               & 1.315E-05 $\pm{1.143E-06}$                                             & 0.048 $\pm{0.001}$                                                       & 16.494 $\pm{0.292}$                                                        \\ \hline
\begin{tabular}[c]{@{}c@{}}Weight Decay\\ + SAM\end{tabular}         & 18.394 $\pm{0.067}$                                                        & 0.816 $\pm{0.001}$                                                     & 0.104 $\pm{0.001}$                                                & 66.827 $\pm{0.286}$                                                          & 0.151 $\pm{0.000}$                                                               & 3.210E-03 $\pm{6.174E-04}$                                             & 1.121 $\pm{0.080}$                                                       & 157.592 $\pm{5.360}$                                                       \\ \hline
\end{tabular}}
\caption{Results for VGG19 trained on CIFAR10 with \textbf{batch size 256 and a learning rate of $\mathbf{1e^{-3}}$}. Numbers in bold indicate best scores for metrics. For sharpness metrics lower values represent flatter models.}
\label{tab:c10_vgg19_256_0.001}
\end{table}

\begin{table}[H]
\centering
\resizebox{\textwidth}{!}{
\begin{tabular}{|c|c|c|c|c|c|c|c|c|}
\hline
\textbf{\begin{tabular}[c]{@{}c@{}}Control\\ Condition\end{tabular}} & \textbf{\begin{tabular}[c]{@{}c@{}}Genralisation \\ Gap\end{tabular}} & \textbf{\begin{tabular}[c]{@{}c@{}}Test \\ Accuracy\end{tabular}} & \textbf{\begin{tabular}[c]{@{}c@{}}Test \\ ECE\end{tabular}} & \textbf{\begin{tabular}[c]{@{}c@{}}Corruption \\ Accuracy\end{tabular}} & \textbf{\begin{tabular}[c]{@{}c@{}}Prediction \\ Disagreement\end{tabular}} & \textbf{\begin{tabular}[c]{@{}c@{}}SAM \\ Sharpness\end{tabular}} & \textbf{\begin{tabular}[c]{@{}c@{}}Fisher-Rao \\ norm\end{tabular}} & \textbf{\begin{tabular}[c]{@{}c@{}}Relative \\ Flatness\end{tabular}} \\ \hline
\begin{tabular}[c]{@{}c@{}}Baseline \\ + SAM\end{tabular}             & \cmark~(1.0) & \cmark~(1.0) & \cmark~(1.0) & \cmark~(1.0) & \cmark~(1.0)               & \cmark~(1.0) & \cmark~(1.0)                & \cmark~(1.0) \\ \hline
Augmentation                                                          & \cmark~(1.0) & \cmark~(1.0) & \cmark~(1.0) & \cmark~(1.0) & \cmark~(1.0)               & \cmark~(1.0) & \cmark~(1.0)                & \cmark~(1.0) \\ \hline
\begin{tabular}[c]{@{}c@{}}Augmentation \\ + SAM\end{tabular}         & \cmark~(1.0) & \cmark~(1.0) & \cmark~(1.0) & \cmark~(1.0) & \cmark~(1.0)               & \cmark~(1.0) & \cmark~(1.0)                & \cmark~(1.0) \\ \hline
Weight Decay                                                          & \xmark~(N/A)  & \xmark~(N/A) & \cmark~(1.0) & \xmark~(N/A) & \cmark~(0.41) & \xmark~(N/A) & \cmark~(0.96) & \cmark~(1.0) \\ \hline
\begin{tabular}[c]{@{}c@{}}Weight Decay \\ + SAM\end{tabular}         & \cmark~(1.0) & \cmark~(1.0) & \cmark~(1.0) & \cmark~(1.0) & \cmark~(1.0)               & \cmark~(1.0) & \cmark~(1.0)                & \cmark~(1.0) \\ \hline
\end{tabular}}
\caption{Significance results for the control conditions against the baseline for VGG19 trained on CIFAR10 with \textbf{batch size 256 and a learning rate of $\mathbf{1e^{-3}}$}.~\cmark~indicates significant difference compared to the baseline; \xmark~indicates significance, the value in the brackets represents the  effect sizes using the absolute RBC.}
\label{tab:vgg19_256_1e3_sig}
\end{table}

\begin{table}[H]
\centering
\resizebox{\textwidth}{!}{
\begin{tabular}{|c|c|c|c|c|c|c|c|c|}
\hline
\textbf{\begin{tabular}[c]{@{}c@{}}Control\\ Condition\end{tabular}} & \textbf{\begin{tabular}[c]{@{}c@{}}Generalisation \\ Gap\end{tabular}} & \textbf{\begin{tabular}[c]{@{}c@{}}Test \\ Accuracy\end{tabular}} & \textbf{\begin{tabular}[c]{@{}c@{}}Test \\ ECE\end{tabular}} & \textbf{\begin{tabular}[c]{@{}c@{}}Corruption \\ Accuracy\end{tabular}} & \textbf{\begin{tabular}[c]{@{}c@{}}Prediction \\ Disagreement\end{tabular}} & \textbf{\begin{tabular}[c]{@{}c@{}}SAM \\ Sharpness\end{tabular}} & \textbf{\begin{tabular}[c]{@{}c@{}}Fisher-Rao \\ norm\end{tabular}} & \textbf{\begin{tabular}[c]{@{}c@{}}Relative \\ Flatness\end{tabular}} \\ \hline
Baseline                                                             & 13.507 $\pm{0.063}$                                                         & 0.865 $\pm{0.001}$                                                     & 0.105 $\pm{0.001}$                                                & 71.476 $\pm{0.125}$                                                          & 0.119 $\pm{0.000}$                                                               & 1.604E-06 $\pm{1.637E-07}$                                             & 0.004 $\pm{0.000}$                                                       & 0.807 $\pm{0.032}$                                                         \\ \hline
\begin{tabular}[c]{@{}c@{}}Baseline\\ + SAM\end{tabular}             & 13.183 $\pm{0.115}$                                                         & 0.868 $\pm{0.001}$                                                     & 0.081 $\pm{0.001}$                                                & 71.908 $\pm{0.121}$                                                          & 0.103 $\pm{0.000}$                                                               & 4.290E-05 $\pm{5.280E-06}$                                             & 0.077 $\pm{0.002}$                                                       & 25.287 $\pm{0.720}$                                                        \\ \hline
Augmentation                                                         & 8.565 $\pm{0.030}$                                                          & 0.910 $\pm{0.000}$                                                     & 0.065 $\pm{0.000}$                                                & 71.491 $\pm{0.442}$                                                          & 0.092 $\pm{0.000}$                                                               & 1.146E-01 $\pm{9.917E-03}$                                             & 2.555 $\pm{0.154}$                                                       & 396.136 $\pm{22.479}$                                                      \\ \hline
\begin{tabular}[c]{@{}c@{}}Augmentation\\ + SAM\end{tabular}         & \textbf{7.969 $\pm{0.080}$}                                                 & \textbf{0.920 $\pm{0.001}$}                                            & \textbf{0.040 $\pm{0.001}$}                                       & \textbf{73.037 $\pm{0.160}$}                                                 & \textbf{0.071 $\pm{0.000}$}                                                      & 3.122E-02 $\pm{3.347E-03}$                                             & 1.087 $\pm{0.014}$                                                       & 429.679 $\pm{6.763}$                                                       \\ \hline
Weight Decay                                                         & 15.241 $\pm{0.124}$                                                         & 0.836 $\pm{0.003}$                                                     & 0.118 $\pm{0.002}$                                                & 68.294 $\pm{0.390}$                                                          & 0.184 $\pm{0.002}$                                                               & 2.302E-02 $\pm{4.987E-03}$                                             & 5.120 $\pm{0.586}$                                                       & 237.081 $\pm{22.113}$                                                      \\ \hline
\begin{tabular}[c]{@{}c@{}}Weight Decay\\ + SAM\end{tabular}         & 13.188 $\pm{0.130}$                                                         & 0.868 $\pm{0.001}$                                                     & 0.063 $\pm{0.001}$                                                & 71.618 $\pm{0.284}$                                                          & 0.117 $\pm{0.000}$                                                               & 5.764E-04 $\pm{5.989E-05}$                                             & 0.742 $\pm{0.022}$                                                       & 124.537 $\pm{3.152}$                                                       \\ \hline
\end{tabular}}
\caption{Results for VGG19 trained on CIFAR10 with \textbf{batch size 256 and a learning rate of $\mathbf{1e^{-2}}$}. Numbers in bold indicate best scores for metrics. For sharpness metrics lower values represent flatter models.}
\label{tab:c10_vgg19_256_0.01}
\end{table}

\begin{table}[H]
\resizebox{\textwidth}{!}{
\begin{tabular}{|c|c|c|c|c|c|c|c|c|}
\hline
\textbf{\begin{tabular}[c]{@{}c@{}}Control\\ Condition\end{tabular}} & \textbf{\begin{tabular}[c]{@{}c@{}}Generalisation \\ Gap\end{tabular}} & \textbf{\begin{tabular}[c]{@{}c@{}}Test \\ Accuracy\end{tabular}} & \textbf{\begin{tabular}[c]{@{}c@{}}Test \\ ECE\end{tabular}} & \textbf{\begin{tabular}[c]{@{}c@{}}Corruption \\ Accuracy\end{tabular}} & \textbf{\begin{tabular}[c]{@{}c@{}}Prediction \\ Disagreement\end{tabular}} & \textbf{\begin{tabular}[c]{@{}c@{}}SAM \\ Sharpness\end{tabular}} & \textbf{\begin{tabular}[c]{@{}c@{}}Fisher-Rao \\ norm\end{tabular}} & \textbf{\begin{tabular}[c]{@{}c@{}}Relative \\ Flatness\end{tabular}} \\ \hline
\begin{tabular}[c]{@{}c@{}}Baseline \\ + SAM\end{tabular}             & \cmark~(0.75) & \cmark~(0.75) & \cmark~(1.0) & \cmark~(0.78) & \cmark~(1.0)  & \cmark~(1.0) & \cmark~(1.0) & \cmark~(1.0) \\ \hline
Augmentation                                                          & \cmark~(1.0)  & \cmark~(1.0)  & \cmark~(1.0) & \xmark~(N/A)  & \cmark~(1.0)  & \cmark~(1.0) & \cmark~(1.0) & \cmark~(1.0) \\ \hline
\begin{tabular}[c]{@{}c@{}}Augmentation \\ + SAM\end{tabular}         & \cmark~(1.0)  & \cmark~(1.0)  & \cmark~(1.0) & \cmark~(1.0)  & \cmark~(1.0)  & \cmark~(1.0) & \cmark~(1.0) & \cmark~(1.0) \\ \hline
Weight Decay                                                          & \xmark~(N/A)   & \xmark~(N/A)  & \xmark~(N/A)  & \xmark~(N/A)  & \xmark~(N/A)   & \cmark~(1.0) & \cmark~(1.0) & \cmark~(1.0) \\ \hline
\begin{tabular}[c]{@{}c@{}}Weight Decay \\ + SAM\end{tabular}         & \xmark~(N/A)   & \xmark~(N/A)  & \cmark~(1.0) & \xmark~(N/A)  & \cmark~(0.29) & \cmark~(1.0) & \cmark~(1.0) & \cmark~(1.0) \\ \hline
\end{tabular}}
\caption{Significance results for the control conditions against the baseline for VGG19 trained on CIFAR10 with \textbf{batch size 256 and a learning rate of $\mathbf{1e^{-2}}$}.~\cmark~indicates significant difference compared to the baseline; \xmark~indicates significance, the value in the brackets represents the  effect sizes using the absolute RBC.}
\label{tab:vgg19_256_1e2_sig}
\end{table}

\begin{table}[H]
\centering
\resizebox{\textwidth}{!}{
\begin{tabular}{|c|c|c|c|c|c|c|c|c|}
\hline
\textbf{\begin{tabular}[c]{@{}c@{}}Control\\ Condition\end{tabular}} & \textbf{\begin{tabular}[c]{@{}c@{}}Generalisation \\ Gap\end{tabular}} & \textbf{\begin{tabular}[c]{@{}c@{}}Test \\ Accuracy\end{tabular}} & \textbf{\begin{tabular}[c]{@{}c@{}}Test \\ ECE\end{tabular}} & \textbf{\begin{tabular}[c]{@{}c@{}}Corruption \\ Accuracy\end{tabular}} & \textbf{\begin{tabular}[c]{@{}c@{}}Prediction \\ Disagreement\end{tabular}} & \textbf{\begin{tabular}[c]{@{}c@{}}SAM \\ Sharpness\end{tabular}} & \textbf{\begin{tabular}[c]{@{}c@{}}Fisher-Rao \\ norm\end{tabular}} & \textbf{\begin{tabular}[c]{@{}c@{}}Relative \\ Flatness\end{tabular}} \\ \hline
Baseline                                                             & 18.026 $\pm{0.066}$                                                         & 0.820 $\pm{0.001}$                                                     & 0.140 $\pm{0.001}$                                                & 68.325 $\pm{0.110}$                                                          & 0.167 $\pm{0.000}$                                                               & 8.460E-07 $\pm{5.355E-08}$                                             & 0.008 $\pm{0.000}$                                                       & 5.824 $\pm{0.058}$                                                         \\ \hline
\begin{tabular}[c]{@{}c@{}}Baseline\\ + SAM\end{tabular}             & 15.059 $\pm{0.085}$                                                         & 0.849 $\pm{0.001}$                                                     & 0.089 $\pm{0.001}$                                                & 69.791 $\pm{0.174}$                                                          & 0.115 $\pm{0.000}$                                                               & 5.689E-04 $\pm{1.248E-04}$                                             & 0.649 $\pm{0.032}$                                                       & 242.483 $\pm{6.334}$                                                       \\ \hline
Augmentation                                                         & 9.988 $\pm{0.088}$                                                          & 0.895 $\pm{0.001}$                                                     & 0.074 $\pm{0.001}$                                                & 70.671 $\pm{0.288}$                                                          & 0.107 $\pm{0.000}$                                                               & 1.708E-01 $\pm{1.967E-02}$                                             & 2.851 $\pm{0.121}$                                                       & 1158.004 $\pm{40.307}$                                                     \\ \hline
\begin{tabular}[c]{@{}c@{}}Augmentation\\ + SAM\end{tabular}         & \textbf{7.594 $\pm{0.050}$}                                                 & \textbf{0.916 $\pm{0.000}$}                                            & \textbf{0.017 $\pm{0.000}$}                                       & \textbf{72.194 $\pm{0.175}$}                                                 & \textbf{0.066 $\pm{0.000}$}                                                      & 7.664E-02 $\pm{1.149E-02}$                                             & 3.469 $\pm{0.023}$                                                       & 2487.050 $\pm{14.772}$                                                     \\ \hline
Weight Decay                                                         & 17.485 $\pm{0.066}$                                                         & 0.825 $\pm{0.001}$                                                     & 0.127 $\pm{0.001}$                                                & 68.655 $\pm{0.156}$                                                          & 0.158 $\pm{0.000}$                                                               & 4.918E-06 $\pm{6.710E-07}$                                             & 0.044 $\pm{0.001}$                                                       & 29.566 $\pm{0.647}$                                                        \\ \hline
\begin{tabular}[c]{@{}c@{}}Weight Decay\\ + SAM\end{tabular}         & 14.803 $\pm{0.082}$                                                         & 0.851 $\pm{0.001}$                                                     & 0.082 $\pm{0.001}$                                                & 69.699 $\pm{0.181}$                                                          & 0.114 $\pm{0.000}$                                                               & 1.089E-03 $\pm{3.140E-04}$                                             & 0.802 $\pm{0.022}$                                                       & 298.386 $\pm{7.049}$                                                       \\ \hline
\end{tabular}}
\caption{Results for VGG19 trained on CIFAR10 with \textbf{batch size 128 and a learning rate of $\mathbf{1e^{-3}}$}. Numbers in bold indicate best scores for metrics. For sharpness metrics lower values represent flatter models.}
\label{tab:c10_vgg19_128_0.001}
\end{table}

\begin{table}[H]
\centering
\resizebox{\textwidth}{!}{
\begin{tabular}{|c|c|c|c|c|c|c|c|c|}
\hline
\textbf{\begin{tabular}[c]{@{}c@{}}Control\\ Condition\end{tabular}} & \textbf{\begin{tabular}[c]{@{}c@{}}Generalisation \\ Gap\end{tabular}} & \textbf{\begin{tabular}[c]{@{}c@{}}Test \\ Accuracy\end{tabular}} & \textbf{\begin{tabular}[c]{@{}c@{}}Test \\ ECE\end{tabular}} & \textbf{\begin{tabular}[c]{@{}c@{}}Corruption \\ Accuracy\end{tabular}} & \textbf{\begin{tabular}[c]{@{}c@{}}Prediction \\ Disagreement\end{tabular}} & \textbf{\begin{tabular}[c]{@{}c@{}}SAM \\ Sharpness\end{tabular}} & \textbf{\begin{tabular}[c]{@{}c@{}}Fisher-Rao \\ norm\end{tabular}} & \textbf{\begin{tabular}[c]{@{}c@{}}Relative \\ Flatness\end{tabular}} \\ \hline
\begin{tabular}[c]{@{}c@{}}Baseline \\ + SAM\end{tabular}             & \cmark~(1.0) & \cmark~(1.0) & \cmark~(1.0) & \cmark~(1.0) & \cmark~(1.0) & \cmark~(1.0) & \cmark~(1.0) & \cmark~(1.0) \\ \hline
Augmentation                                                          & \cmark~(1.0) & \cmark~(1.0) & \cmark~(1.0) & \cmark~(1.0) & \cmark~(1.0) & \cmark~(1.0) & \cmark~(1.0) & \cmark~(1.0) \\ \hline
\begin{tabular}[c]{@{}c@{}}Augmentation \\ + SAM\end{tabular}         & \cmark~(1.0) & \cmark~(1.0) & \cmark~(1.0) & \cmark~(1.0) & \cmark~(1.0) & \cmark~(1.0) & \cmark~(1.0) & \cmark~(1.0) \\ \hline
Weight Decay                                                          & \cmark~(1.0) & \cmark~(1.0) & \cmark~(1.0) & \xmark~(N/A) & \cmark~(1.0) & \cmark~(1.0) & \cmark~(1.0) & \cmark~(1.0) \\ \hline
\begin{tabular}[c]{@{}c@{}}Weight Decay \\ + SAM\end{tabular}         & \cmark~(1.0) & \cmark~(1.0) & \cmark~(1.0) & \cmark~(1.0) & \cmark~(1.0) & \cmark~(1.0) & \cmark~(1.0) & \cmark~(1.0) \\ \hline
\end{tabular}}
\caption{Significance results for the control conditions against the baseline for VGG19 trained on CIFAR10 with \textbf{batch size 128 and a learning rate of $\mathbf{1e^{-3}}$}.~\cmark~indicates significant difference compared to the baseline; \xmark~indicates significance, the value in the brackets represents the  effect sizes using the absolute RBC.}
\label{tab:vgg19_128_1e3_sig}
\end{table}

\begin{table}[H]
\centering
\resizebox{\textwidth}{!}{
\begin{tabular}{|c|c|c|c|c|c|c|c|c|}
\hline
\textbf{\begin{tabular}[c]{@{}c@{}}Control\\ Condition\end{tabular}} & \textbf{\begin{tabular}[c]{@{}c@{}}Generalisation \\ Gap\end{tabular}} & \textbf{\begin{tabular}[c]{@{}c@{}}Test \\ Accuracy\end{tabular}} & \textbf{\begin{tabular}[c]{@{}c@{}}Test \\ ECE\end{tabular}} & \textbf{\begin{tabular}[c]{@{}c@{}}Corruption \\ Accuracy\end{tabular}} & \textbf{\begin{tabular}[c]{@{}c@{}}Prediction \\ Disagreement\end{tabular}} & \textbf{\begin{tabular}[c]{@{}c@{}}SAM \\ Sharpness\end{tabular}} & \textbf{\begin{tabular}[c]{@{}c@{}}Fisher-Rao \\ norm\end{tabular}} & \textbf{\begin{tabular}[c]{@{}c@{}}Relative \\ Flatness\end{tabular}} \\ \hline
Baseline                                                             & 12.661 $\pm{0.066}$                                                         & 0.873 $\pm{0.001}$                                                     & 0.100 $\pm{0.001}$                                                & 71.689 $\pm{0.113}$                                                          & 0.111 $\pm{0.000}$                                                               & 3.227E-07 $\pm{3.198E-08}$                                             & 0.003 $\pm{0.000}$                                                       & 1.331 $\pm{0.031}$                                                         \\ \hline
\begin{tabular}[c]{@{}c@{}}Baseline\\ + SAM\end{tabular}             & 12.248 $\pm{0.078}$                                                         & 0.878 $\pm{0.001}$                                                     & 0.077 $\pm{0.001}$                                                & 71.838 $\pm{0.256}$                                                          & 0.100 $\pm{0.000}$                                                               & 8.424E-06 $\pm{9.410E-07}$                                             & 0.054 $\pm{0.003}$                                                       & 40.853 $\pm{2.177}$                                                        \\ \hline
Augmentation                                                         & 8.333 $\pm{0.060}$                                                          & 0.913 $\pm{0.001}$                                                     & 0.063 $\pm{0.001}$                                                & 72.114 $\pm{0.259}$                                                          & 0.091 $\pm{0.000}$                                                               & 1.605E-01 $\pm{2.042E-02}$                                             & 2.510 $\pm{0.122}$                                                       & 772.205 $\pm{19.954}$                                                      \\ \hline
\begin{tabular}[c]{@{}c@{}}Augmentation\\ + SAM\end{tabular}         & \textbf{7.791 $\pm{0.072}$}                                                 & \textbf{0.922 $\pm{0.001}$}                                            & \textbf{0.038 $\pm{0.001}$}                                       & \textbf{72.470 $\pm{0.190}$}                                                 & \textbf{0.072 $\pm{0.000}$}                                                      & 3.497E-02 $\pm{4.347E-03}$                                             & 1.076 $\pm{0.019}$                                                       & 1155.243 $\pm{35.334}$                                                     \\ \hline
Weight Decay                                                         & 14.589 $\pm{0.143}$                                                         & 0.839 $\pm{0.003}$                                                     & 0.110 $\pm{0.002}$                                                & 66.692 $\pm{0.603}$                                                          & 0.187 $\pm{0.001}$                                                               & 7.018E-03 $\pm{8.372E-04}$                                             & 6.253 $\pm{0.596}$                                                       & 608.657 $\pm{44.645}$                                                      \\ \hline
\begin{tabular}[c]{@{}c@{}}Weight Decay\\ + SAM\end{tabular}         & 12.295 $\pm{0.112}$                                                         & 0.877 $\pm{0.001}$                                                     & 0.052 $\pm{0.001}$                                                & 71.167 $\pm{0.301}$                                                          & 0.116 $\pm{0.001}$                                                               & 5.407E-04 $\pm{1.109E-04}$                                             & 1.193 $\pm{0.043}$                                                       & 406.163 $\pm{15.541}$                                                      \\ \hline
\end{tabular}}
\caption{Results for VGG19 trained on CIFAR10 with \textbf{batch size 128 and a learning rate of $\mathbf{1e^{-2}}$}. Numbers in bold indicate best scores for metrics. For sharpness metrics lower values represent flatter models.}
\label{tab:c10_vgg19_128_0.01}
\end{table}

\begin{table}[H]
\centering
\resizebox{\textwidth}{!}{
\begin{tabular}{|c|c|c|c|c|c|c|c|c|}
\hline
\textbf{\begin{tabular}[c]{@{}c@{}}Control\\ Condition\end{tabular}} & \textbf{\begin{tabular}[c]{@{}c@{}}Generalisation \\ Gap\end{tabular}} & \textbf{\begin{tabular}[c]{@{}c@{}}Test \\ Accuracy\end{tabular}} & \textbf{\begin{tabular}[c]{@{}c@{}}Test \\ ECE\end{tabular}} & \textbf{\begin{tabular}[c]{@{}c@{}}Corruption \\ Accuracy\end{tabular}} & \textbf{\begin{tabular}[c]{@{}c@{}}Prediction \\ Disagreement\end{tabular}} & \textbf{\begin{tabular}[c]{@{}c@{}}SAM \\ Sharpness\end{tabular}} & \textbf{\begin{tabular}[c]{@{}c@{}}Fisher-Rao \\ norm\end{tabular}} & \textbf{\begin{tabular}[c]{@{}c@{}}Relative \\ Flatness\end{tabular}} \\ \hline
\begin{tabular}[c]{@{}c@{}}Baseline \\ + SAM\end{tabular}            & \cmark~(1.0)                & \cmark~(1.0)                & \cmark~(1.0) & \xmark~(N/A)                & \cmark~(1.0) & \cmark~(1.0) & \cmark~(1.0) & \cmark~(1.0) \\ \hline
Augmentation                                                         & \cmark~(1.0)                & \cmark~(1.0)                & \cmark~(1.0) & \xmark~(N/A)                & \cmark~(1.0) & \cmark~(1.0) & \cmark~(1.0) & \cmark~(1.0) \\ \hline
\begin{tabular}[c]{@{}c@{}}Augmentation \\ + SAM\end{tabular}        & \cmark~(1.0)                & \cmark~(1.0)                & \cmark~(1.0) & \cmark~(0.75) & \cmark~(1.0) & \cmark~(1.0) & \cmark~(1.0) & \cmark~(1.0) \\ \hline
Weight Decay                                                         & \xmark~(N/A)                 & \xmark~(N/A)                & \xmark~(N/A)  & \xmark~(N/A)                & \xmark~(N/A)  & \cmark~(1.0) & \cmark~(1.0) & \cmark~(1.0) \\ \hline
\begin{tabular}[c]{@{}c@{}}Weight Decay \\ + SAM\end{tabular}        & \cmark~(0.89) & \cmark~(0.82) & \cmark~(1.0) & \xmark~(N/A)                & \xmark~(N/A)  & \cmark~(1.0) & \cmark~(1.0) & \cmark~(1.0) \\ \hline
\end{tabular}}
\caption{Significance results for the control conditions against the baseline for VGG19 trained on CIFAR10 with \textbf{batch size 128 and a learning rate of $\mathbf{1e^{-2}}$}.~\cmark~indicates significant difference compared to the baseline; \xmark~indicates significance, the value in the brackets represents the  effect sizes using the absolute RBC.}
\label{tab:vgg19_128_1e2_sig}
\end{table}

\begin{table}[H]
\centering

\begin{tabular}{|c|c|c|c|}
\hline
                     & \textbf{Baseline} & \textbf{Weight Decay} & \textbf{Augmentation} \\ \hline
\textbf{\begin{tabular}[c]{@{}c@{}}Without \\ SAM\end{tabular}} &  \includegraphics[width=0.25\textwidth]{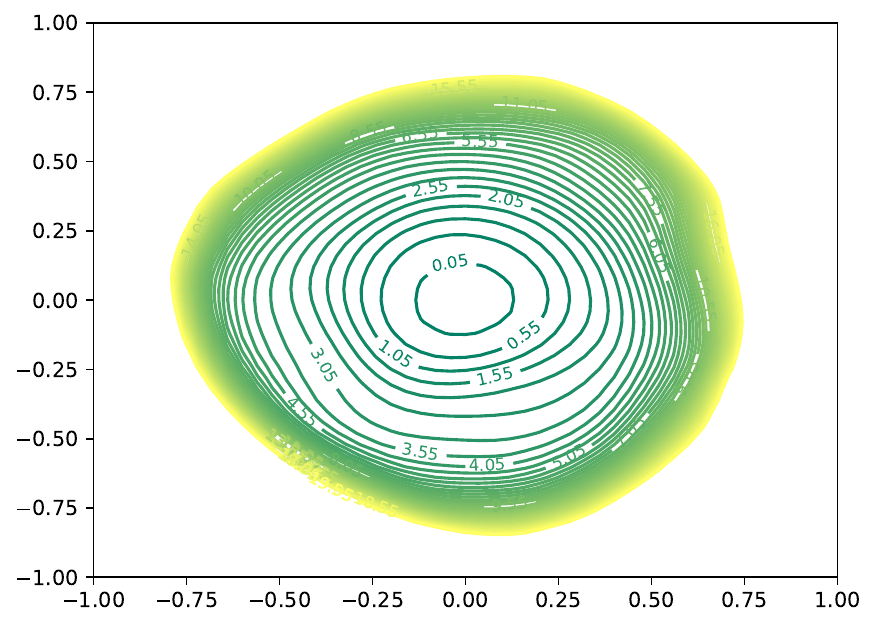}                  & \includegraphics[width=0.25\textwidth]{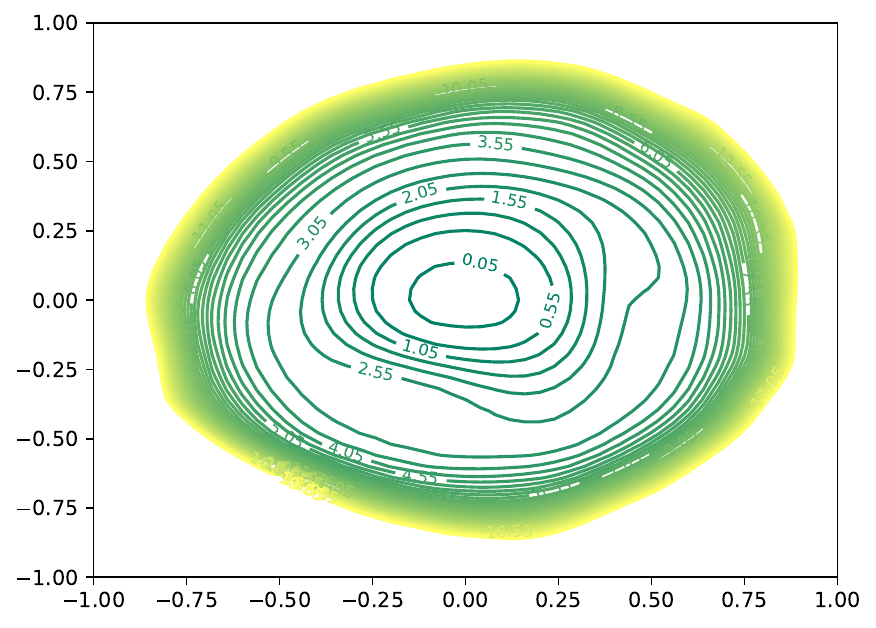}                          &  \includegraphics[width=0.25\textwidth]{figures/loss_landscapes/VGG/VGG19_aug.pth_weights_xignore=biasbn_xnorm=filter_yignore=biasbn_ynorm=filter.h5_-1.0,1.0,51x-1.0,1.0,51.h5_train_loss_2dcontour.pdf}                       \\ \hline
\textbf{\begin{tabular}[c]{@{}c@{}}With \\ SAM\end{tabular}}    & \includegraphics[width=0.25\textwidth]{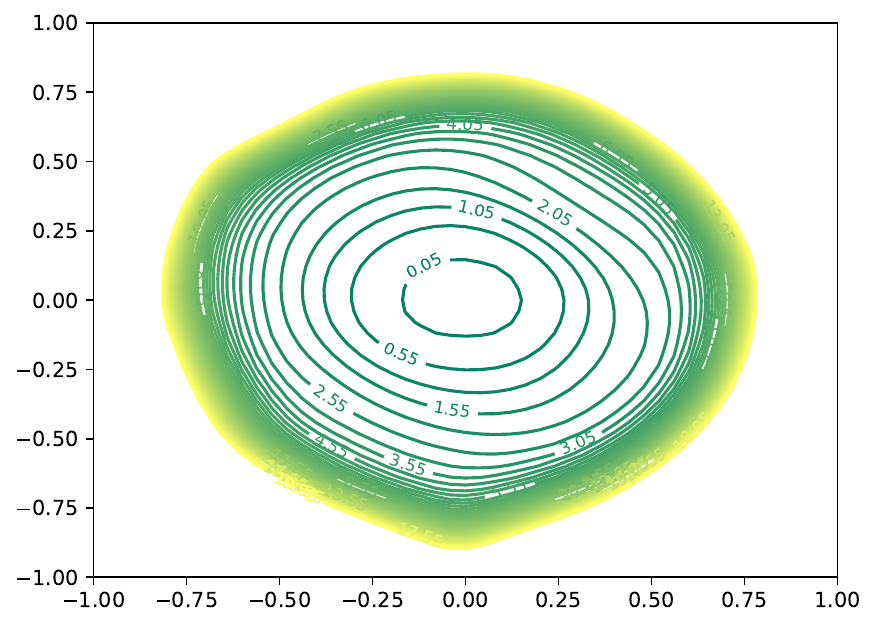}                    & \includegraphics[width=0.25\textwidth]{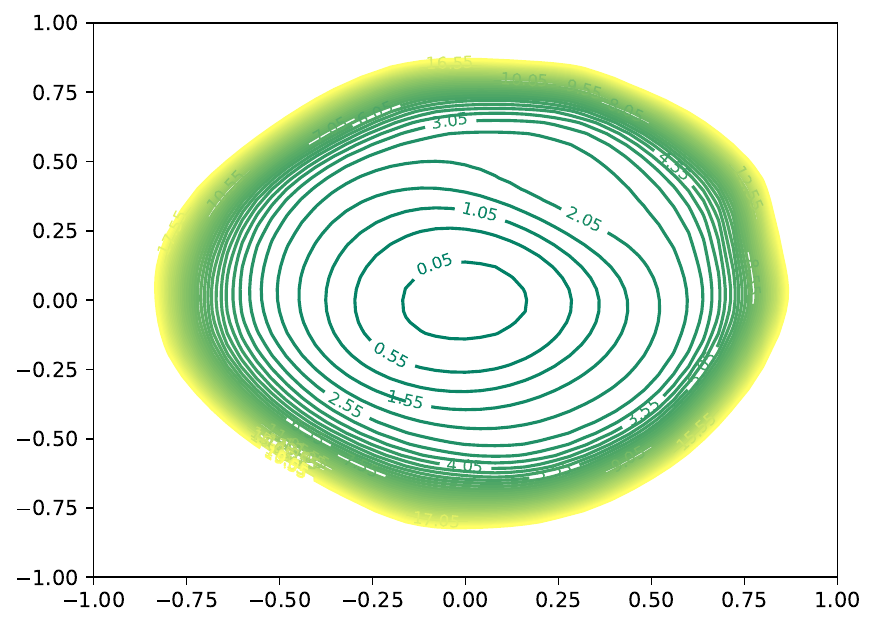}                        & \includegraphics[width=0.25\textwidth]{figures/loss_landscapes/VGG/VGG19_aug_SAM.pth_weights_xignore=biasbn_xnorm=filter_yignore=biasbn_ynorm=filter.h5_-1.0,1.0,51x-1.0,1.0,51.h5_train_loss_2dcontour.pdf}                          \\ \hline
\end{tabular}
\caption{Loss landscape visualisation \citep{li2018visualizing} of VGG19 landscape on CIFAR10 exploring
the loss in the domain of the perturbations  $[-1,1]^2$ with 51 steps in both directions}.
\label{vis_vgg_c10}
\end{table}
In line with the hyperparameter sweeps of the ResNet (Section~\ref{sec:ResNet_param_sweep}), we find that the three main core findings hold for the VGG architecture.  We observe across hyperparameter settings in Tables~\ref{tab:vgg19_128_1e3_sig},~\ref{tab:vgg19_128_1e2_sig},~\ref{tab:128-1e3-c10} and~\ref{tab:vgg19_128_1e2_sig} that the three core insights remain supported. \textbf{1}: improved generalisation performance can be found at sharper minima, \textbf{2}: improved reliability mostly coincides with sharper minima and \textbf{3}: sharper minima often represent the best performing models. Once again, it should be noted that there are instances where weight decay increases the sharpness of minima against the baseline but does not provide improved reliability or generalisation performance. 

\textit{CIFAR10 Landscape Visualisation}. Here we observe that the loss landscapes show that the use of regularisation does change the function learned by the model and that this can often increase in complexity. For example, in Table~\ref{vis_vgg_c10} we can see that the use of weight decay, augmentation and SAM all change the minima that is reached at the end of training, with weight decay and augmentation showing a big increase in complexity compared to the baseline landscape. 

\subsection{CIFAR100:} For the VGG19 on the CIFAR100 data set we observe that augmentation + SAM condition performs the best for test accuracy, corruption accuracy and prediction disagreement. However, for ECE, we see that weight decay is the best condition. Augmentation + SAM are the second sharpest models for Fisher-Rao norm and SAM sharpness, and have the highest value for Relative Flatness. It is important to note that for weight decay, with the lowest ECE, it has higher sharpness values than the baseline condition. 

\begin{table}[H]
\centering
\resizebox{\textwidth}{!}{
\begin{tabular}{|c|c|c|c|c|c|c|c|c|}
\hline
\textbf{\begin{tabular}[c]{@{}c@{}}Control\\ Condition\end{tabular}} & \textbf{\begin{tabular}[c]{@{}c@{}}Generalisation \\ Gap\end{tabular}} & \textbf{\begin{tabular}[c]{@{}c@{}}Test \\ Accuracy\end{tabular}} & \textbf{\begin{tabular}[c]{@{}c@{}}Test \\ ECE\end{tabular}} & \textbf{\begin{tabular}[c]{@{}c@{}}Corruption \\ Accuracy\end{tabular}} & \textbf{\begin{tabular}[c]{@{}c@{}}Prediction \\ Disagreement\end{tabular}} & \textbf{\begin{tabular}[c]{@{}c@{}}SAM \\ Sharpness\end{tabular}} & \textbf{\begin{tabular}[c]{@{}c@{}}Fisher-Rao \\ norm\end{tabular}} & \textbf{\begin{tabular}[c]{@{}c@{}}Relative \\ Flatness\end{tabular}} \\ \hline
Baseline                                                             & 42.454 $\pm{0.092}$                                                         & 0.575 $\pm{0.001}$                                                     & 0.253 $\pm{0.000}$                                                & 40.749 $\pm{0.124}$                                                          & 0.396 $\pm{0.000}$                                                               & 2.123E-04 $\pm{2.649E-05}$                                             & 0.158 $\pm{0.017}$                                                       & 8.384 $\pm{0.151}$                                                         \\ \hline
\begin{tabular}[c]{@{}c@{}}Baseline\\ + SAM\end{tabular}             & 43.815 $\pm{0.224}$                                                         & 0.561 $\pm{0.002}$                                                     & 0.232 $\pm{0.002}$                                                & 39.690 $\pm{0.196}$                                                          & 0.399 $\pm{0.001}$                                                               & 7.520E-04 $\pm{5.791E-05}$                                             & 0.529 $\pm{0.017}$                                                       & 67.485 $\pm{1.802}$                                                        \\ \hline
Augmentation                                                         & 32.519 $\pm{0.156}$                                                         & 0.646 $\pm{0.002}$                                                     & 0.222 $\pm{0.002}$                                                & 40.832 $\pm{0.321}$                                                          & 0.358 $\pm{0.001}$                                                               & 2.835E-01 $\pm{1.439E-02}$                                             & 7.156 $\pm{0.270}$                                                       & 1430.826 $\pm{53.977}$                                                     \\ \hline
\begin{tabular}[c]{@{}c@{}}Augmentation\\ + SAM\end{tabular}         & \textbf{32.008 $\pm{0.099}$}                                                & \textbf{0.656 $\pm{0.001}$}                                            & \textbf{0.157 $\pm{0.001}$}                                       & \textbf{41.276 $\pm{0.089}$}                                                 & \textbf{0.326 $\pm{0.001}$}                                                      & 1.971E-01 $\pm{1.170E-02}$                                             & 5.653 $\pm{0.073}$                                                       & 2085.080 $\pm{31.648}$                                                     \\ \hline
Weight Decay                                                         & 41.579 $\pm{0.107}$                                                         & 0.584 $\pm{0.001}$                                                     & 0.138 $\pm{0.000}$                                                & 41.266 $\pm{0.112}$                                                          & 0.384 $\pm{0.000}$                                                               & 3.302E-04 $\pm{3.256E-05}$                                             & 0.678 $\pm{0.008}$                                                       & 45.728 $\pm{0.073}$                                                        \\ \hline
\begin{tabular}[c]{@{}c@{}}Weight Decay\\ + SAM\end{tabular}         & 44.631 $\pm{0.228}$                                                         & 0.553 $\pm{0.002}$                                                     & 0.189 $\pm{0.002}$                                                & 38.961 $\pm{0.191}$                                                          & 0.429 $\pm{0.001}$                                                               & 2.630E-03 $\pm{2.111E-04}$                                             & 2.138 $\pm{0.084}$                                                       & 153.194 $\pm{6.495}$                                                       \\ \hline
\end{tabular}}
\caption{Results for VGG-19 trained on CIFAR100, the Mean and $\pm{}$ 1 SEM are recorded over 10 models. Numbers in bold indicate best scores for metrics. For sharpness metrics lower values represent flatter models.}
\label{tab:c100_vgg19_256_0.001}
\end{table}
\begin{table}[H]
\centering
\resizebox{\textwidth}{!}{
\begin{tabular}{|c|c|c|c|c|c|c|c|c|}
\hline
\textbf{\begin{tabular}[c]{@{}c@{}}Control\\ Condition\end{tabular}} & \textbf{\begin{tabular}[c]{@{}c@{}}Generalisation \\ Gap\end{tabular}} & \textbf{\begin{tabular}[c]{@{}c@{}}Test \\ Accuracy\end{tabular}} & \textbf{\begin{tabular}[c]{@{}c@{}}Test \\ ECE\end{tabular}} & \textbf{\begin{tabular}[c]{@{}c@{}}Corruption \\ Accuracy\end{tabular}} & \textbf{\begin{tabular}[c]{@{}c@{}}Prediction \\ Disagreement\end{tabular}} & \textbf{\begin{tabular}[c]{@{}c@{}}SAM \\ Sharpness\end{tabular}} & \textbf{\begin{tabular}[c]{@{}c@{}}Fisher-Rao \\ norm\end{tabular}} & \textbf{\begin{tabular}[c]{@{}c@{}}Relative \\ Flatness\end{tabular}} \\ \hline
\begin{tabular}[c]{@{}c@{}}Baseline \\ + SAM\end{tabular}            & \xmark~(N/A)  & \xmark~(N/A) & \cmark~(1.0) & \xmark~(N/A)  & \xmark~(N/A)  & \cmark~(1.0) & \cmark~(1.0) & \cmark~(1.0) \\ \hline
Augmentation                                                         & \cmark~(1.0) & \cmark~(1.0) & \cmark~(1.0) & \xmark~(N/A)  & \cmark~(1.0) & \cmark~(1.0) & \cmark~(1.0) & \cmark~(1.0) \\ \hline
\begin{tabular}[c]{@{}c@{}}Augmentation \\ + SAM\end{tabular}        & \cmark~(1.0) & \cmark~(1.0) & \cmark~(1.0) & \cmark~(0.71) & \cmark~(1.0) & \cmark~(1.0) & \cmark~(1.0) & \cmark~(1.0) \\ \hline
Weight Decay                                                         & \cmark~(1.0) & \cmark~(1.0) & \cmark~(1.0) & \cmark~(0.89) & \cmark~(1.0) & \cmark~(1.0) & \cmark~(1.0) & \cmark~(1.0) \\ \hline
\begin{tabular}[c]{@{}c@{}}Weight Decay \\ + SAM\end{tabular}        & \xmark~(N/A)  & \xmark~(N/A) & \cmark~(1.0) & \xmark~(N/A)  & \xmark~(N/A)  & \cmark~(1.0) & \cmark~(1.0) & \cmark~(1.0) \\ \hline
\end{tabular}}
\caption{Significance results for the control conditions against the baseline for VGG19 trained on CIFAR100.~\cmark~indicates significant difference compared to the baseline; \xmark~indicates no significance, the value in the brackets represents the  effect sizes using the absolute RBC.}
\label{sig:c100_vgg19}
\end{table}

\begin{table}[H]
\centering
\begin{tabular}{|c|c|c|c|}
\hline
                     & \textbf{Baseline} & \textbf{Weight Decay} & \textbf{Augmentation} \\ \hline
\textbf{\begin{tabular}[c]{@{}c@{}}Without \\ SAM\end{tabular}} &  \includegraphics[width=0.25\textwidth]{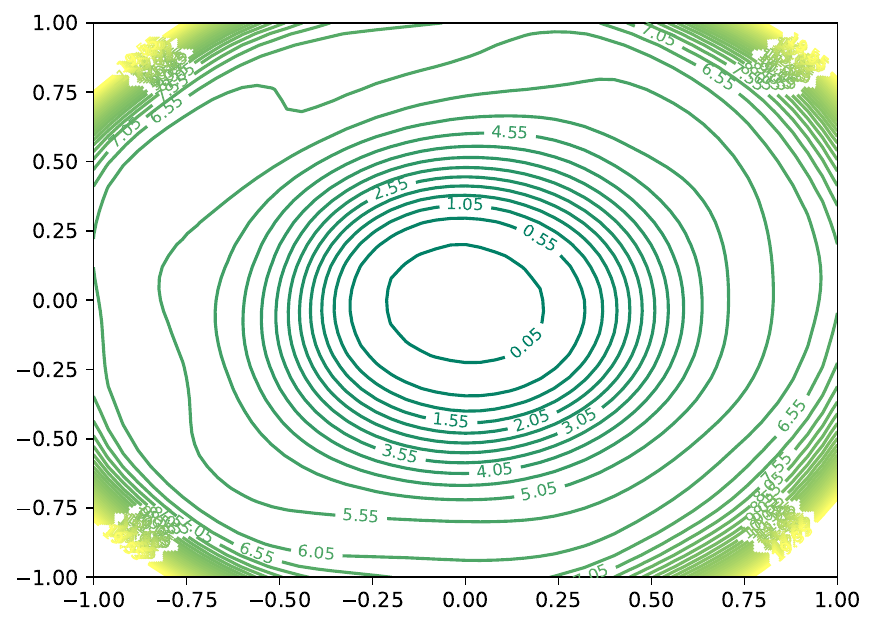}                  & \includegraphics[width=0.25\textwidth]{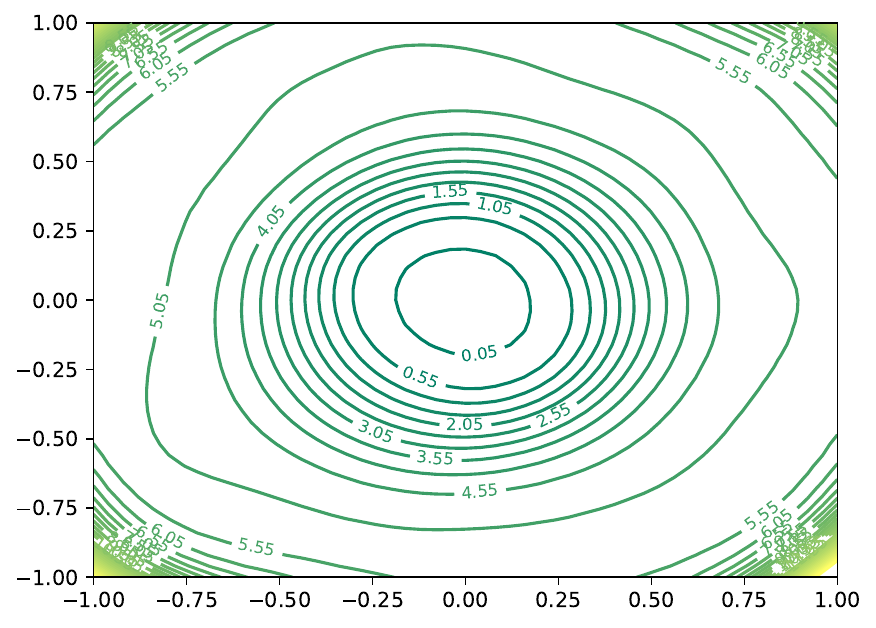}                          &  \includegraphics[width=0.25\textwidth]{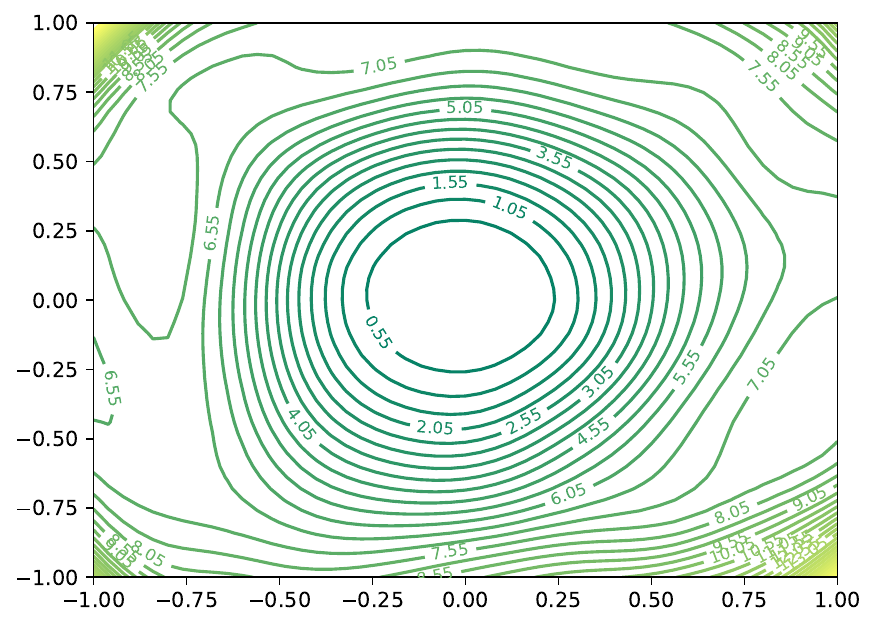}                       \\ \hline
\textbf{\begin{tabular}[c]{@{}c@{}}With \\ SAM\end{tabular}}    & \includegraphics[width=0.25\textwidth]{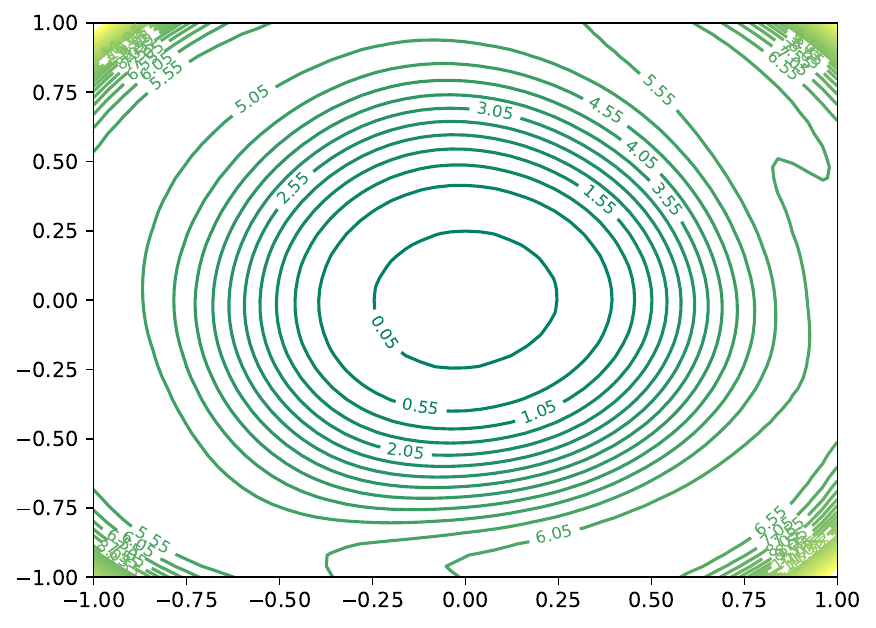}                    & \includegraphics[width=0.25\textwidth]{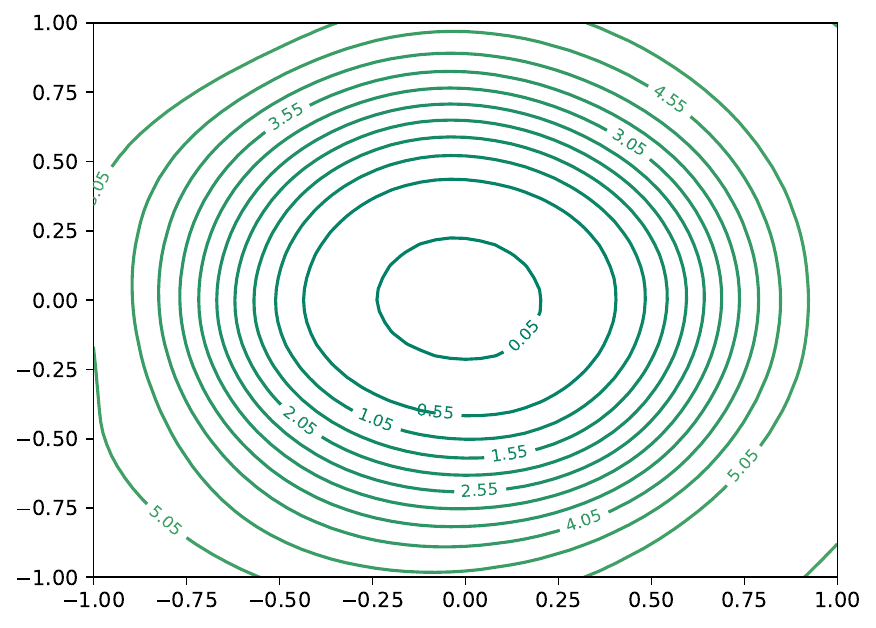}                        & \includegraphics[width=0.25\textwidth]{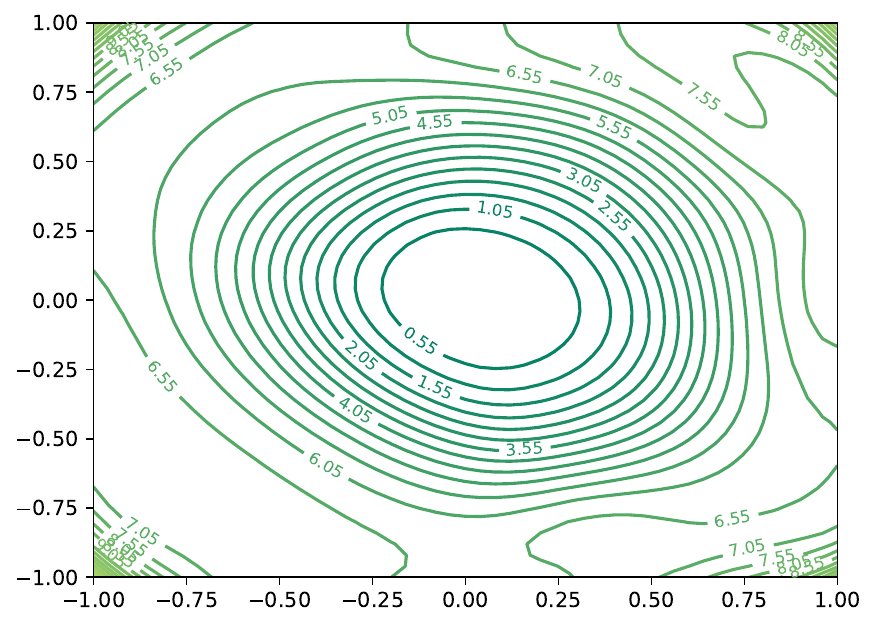}                          \\ \hline
\end{tabular}
\caption{Loss landscape visualisation  \citep{li2018visualizing} VGG19 landscape on CIFAR100 exploring the loss in the domain of the perturbations $[-1,1]^2$ with 51 steps in both directions}.
\label{cifar100_vgg_vis}
\end{table}
When considering the significance of these results, we see that the overall findings hold. In Table~\ref{sig:c100_vgg19}, it can be observed that regularisers weight decay, augmentation and augmentation +SAM result in models that have statistically supported improvements across generalisation and reliability metrics as well as reaching statistically supported increases in minima sharpness. As a result, the findings in the main paper hold in this case. We also observe that there are instances where models reach sharper minima but are not as performant in the case of baseline + SAM and weight decay + SAM, which shows that SAM regularisation may be too strong and that the sharpness evoked by the SAM regulariser can be suboptimal for this learning task on this architecture.  
\textit{CIFAR100 Landscape Visualisation:} Once again, we confirm through the loss landscape visualisation in Table~\ref{cifar100_vgg_vis}, that the application of regularisers does indeed change the properties of the minima that a network reaches at the end of training. This, corroborates our findings that state that regularisation can change the function complexity of a network and thus impact the geometric properties of the minima found at the end of training.

\subsection{TinyImageNet:}  
For the VGG19BN on TinyImageNet it can be observed that the weight decay + SAM condition performs best for test accuracy and prediction disagreement. For weight decay + SAM condition we see no real difference in the sharpness values. For ECE we see that augmentation + SAM is the best condition. Augmentation + SAM is the second sharpest model for Fisher-Rao norm and SAM sharpness.  In this instance, the results indicate that the models trained in the augmentation and augmentation + SAM conditions are under optimised as their generalisation gaps are small while the test accuracy remains low indicating that the models have not fully fit the training set for this setting.
\begin{table}[H]
\centering
\resizebox{\textwidth}{!}{
\begin{tabular}{|c|c|c|c|c|c|c|c|}
\hline
\textbf{\begin{tabular}[c]{@{}c@{}}Control\\ Condition\end{tabular}} & \textbf{\begin{tabular}[c]{@{}c@{}}Generalisation \\ Gap\end{tabular}} & \textbf{\begin{tabular}[c]{@{}c@{}}Test \\ Accuracy\end{tabular}} & \textbf{\begin{tabular}[c]{@{}c@{}}Test\\  ECE\end{tabular}} & \textbf{\begin{tabular}[c]{@{}c@{}}Corruption \\ Accuracy\end{tabular}} & \textbf{\begin{tabular}[c]{@{}c@{}}Prediction \\ Disagreement\end{tabular}} & \textbf{\begin{tabular}[c]{@{}c@{}}SAM \\ Sharpness\end{tabular}} & \textbf{\begin{tabular}[c]{@{}c@{}}Fisher-Rao \\ norm\end{tabular}} \\ \hline
Baseline                                                             & 39.588 $\pm{0.063}$                                                         & 0.604 $\pm{0.001}$                                                     & 0.303 $\pm{0.001}$                                                & 39.082 $\pm{0.042}$                                                          & 0.238 $\pm{0.000}$                                                               & 2.642E-04 $\pm{1.986E-05}$                                             & 0.337 $\pm{0.118}$                                                       \\ \hline
\begin{tabular}[c]{@{}c@{}}Baseline\\ + SAM\end{tabular}             & 36.131 $\pm{0.048}$                                                         & 0.638 $\pm{0.000}$                                                     & 0.199 $\pm{0.001}$                                                & 41.815 $\pm{0.038}$                                                          & 0.186 $\pm{0.000}$                                                               & 3.364E-04 $\pm{2.684E-05}$                                             & 0.419 $\pm{0.097}$                                                       \\ \hline
Augmentation                                                         & 20.952 $\pm{0.080}$                                                         & 0.578 $\pm{0.001}$                                                     & 0.119 $\pm{0.001}$                                                & 42.359 $\pm{0.123}$                                                          & 0.473 $\pm{0.000}$                                                               & 1.893E+00 $\pm{7.702E-02}$                                             & 20.033 $\pm{0.076}$                                                      \\ \hline
\begin{tabular}[c]{@{}c@{}}Augmentation\\ + SAM\end{tabular}         & 17.927 $\pm{0.048}$                                                         & 0.594 $\pm{0.000}$                                                     & 0.056 $\pm{0.002}$                                                & 44.449 $\pm{0.125}$                                                          & 0.440 $\pm{0.000}$                                                               & 1.665E+00 $\pm{5.137E-02}$                                             & 19.230 $\pm{0.035}$                                                      \\ \hline
Weight Decay                                                         & 39.622 $\pm{0.069}$                                                         & 0.604 $\pm{0.001}$                                                     & 0.265 $\pm{0.000}$                                                & 39.148 $\pm{0.026}$                                                          & 0.222 $\pm{0.000}$                                                               & 2.679E-04 $\pm{1.084E-05}$                                             & 0.207 $\pm{0.026}$                                                       \\ \hline
\begin{tabular}[c]{@{}c@{}}Weight Decay\\ + SAM\end{tabular}         & 35.922 $\pm{0.050}$                                                         & 0.641 $\pm{0.001}$                                                     & 0.180 $\pm{0.001}$                                                & 42.106 $\pm{0.048}$                                                          & 0.185 $\pm{0.000}$                                                               & 3.072E-04 $\pm{4.621E-06}$                                             & 0.342 $\pm{0.015}$                                                       \\ \hline
\end{tabular}}
\caption{Results for VGG19-BN (Pre-Trained) on TinyImageNet. Numbers in bold indicate best scores for metrics. For sharpness metrics lower values represent flatter models.}
\label{tab:TiN_vgg19}
\end{table}

\begin{table}[H]
\centering
\resizebox{\textwidth}{!}{
\begin{tabular}{|c|c|c|c|c|c|c|c|c|}
\hline
\textbf{\begin{tabular}[c]{@{}c@{}}Control\\ Condition\end{tabular}} & \textbf{\begin{tabular}[c]{@{}c@{}}Generalisation \\ Gap\end{tabular}} & \textbf{\begin{tabular}[c]{@{}c@{}}Test \\ Accuracy\end{tabular}} & \textbf{\begin{tabular}[c]{@{}c@{}}Test \\ ECE\end{tabular}} & \textbf{\begin{tabular}[c]{@{}c@{}}Corruption \\ Accuracy\end{tabular}} & \textbf{\begin{tabular}[c]{@{}c@{}}Prediction \\ Disagreement\end{tabular}} & \textbf{\begin{tabular}[c]{@{}c@{}}SAM \\ Sharpness\end{tabular}} & \textbf{\begin{tabular}[c]{@{}c@{}}Fisher-Rao \\ norm\end{tabular}} \\ \hline
\begin{tabular}[c]{@{}c@{}}Baseline \\ + SAM\end{tabular} & \cmark~(1.0) & \cmark~(1.0) & \cmark~(1.0) & \cmark~(1.0) & \cmark~(1.0) & \xmark~(N/A) & \xmark~(N/A) \\ \hline
Augmentation & \cmark~(1.0) & \xmark~(N/A) & \cmark~(1.0) & \cmark~(1.0) & \xmark~(N/A) & \cmark~(1.0) & \cmark~(1.0) \\ \hline
\begin{tabular}[c]{@{}c@{}}Augmentation\\ + SAM\end{tabular} & \cmark~(1.0) & \xmark~(N/A) & \cmark~(1.0) & \cmark~(1.0) & \xmark~(N/A) & \cmark~(1.0) & \cmark~(1.0) \\ \hline
Weight Decay & \xmark~(N/A) & \xmark~(N/A) & \cmark~(1.0) & \xmark~(N/A) & \cmark~(1.0) & \xmark~(N/A) & \xmark~(N/A) \\ \hline
\begin{tabular}[c]{@{}c@{}}Weight Decay\\ + SAM\end{tabular} & \cmark~(1.0) & \cmark~(1.0) & \cmark~(1.0) & \cmark~(1.0) & \cmark~(1.0) & \xmark~(N/A) & \xmark~(N/A) \\ \hline
\end{tabular}}
\caption{Significance results for the control conditions against the baseline for VGG19 trained on TinyImageNet.~\cmark~indicates significant difference compared to the baseline; \xmark~indicates no significance the value in the brackets represents the  effect sizes using the absolute RBC.}
\label{sig:TiN_vgg19}
\end{table}

These results show that both SAM and weight decay + SAM produce flatter models than the baseline, with improved reliability and generalisation, as shown in Table~\ref{sig:TiN_vgg19}. In contrast, the augmentation and augmentation + SAM conditions may be under-optimised and perform worse; they are also sharper. This case demonstrates a preference for flatness and serves as a reminder to apply caution when making broad statements about geometric property preference. 

\section{Vision Transformer}
For the Vision Transformer (ViT), we see that our findings are consistent with the findings observed with ResNet and VGG architectures. Often regularisers increase the sharpness of the resulting model; more frequently than not, this increased sharpness coincides with improves generalisation and reliability. 
\label{app:ViT}

\subsection{CIFAR10} 

For this architecture on CIFAR10 the baseline sharpness exceeds that of ResNet and VGG, which are trained with the same batch size and learning rate. This may indicate that ViT's learning biases the models towards narrower decision boundaries in general. However, further investigation is needed to confirm this. In Table~\ref{tab:c10_ViT_256_0.001}, augmentation and augmentation + SAM conditions achieve the best performance and exhibit the highest sharpness values across metrics, which is consistent with our previous findings.

\begin{table}[H]
\centering
\resizebox{\textwidth}{!}{
\begin{tabular}{|c|c|c|c|c|c|c|c|c|}
\hline
\textbf{\begin{tabular}[c]{@{}c@{}}Control\\ Condition\end{tabular}} & \textbf{\begin{tabular}[c]{@{}c@{}}Generalisation \\ Gap\end{tabular}} & \textbf{\begin{tabular}[c]{@{}c@{}}Test \\ Accuracy\end{tabular}} & \textbf{\begin{tabular}[c]{@{}c@{}}Test \\ ECE\end{tabular}} & \textbf{\begin{tabular}[c]{@{}c@{}}Corruption \\ Accuracy\end{tabular}} & \textbf{\begin{tabular}[c]{@{}c@{}}Prediction \\ Disagreement\end{tabular}} & \textbf{\begin{tabular}[c]{@{}c@{}}SAM \\ Sharpness\end{tabular}} & \textbf{\begin{tabular}[c]{@{}c@{}}Fisher-Rao \\ norm\end{tabular}} & \textbf{\begin{tabular}[c]{@{}c@{}}Relative \\ Flatness\end{tabular}} \\ \hline
Baseline                                                             & 39.040 $\pm{0.177}$                                                         & 0.610 $\pm{0.002}$                                                     & 0.308 $\pm{0.002}$                                                & 54.805 $\pm{0.147}$                                                          & 0.408 $\pm{0.001}$                                                               & 8.769E-05 $\pm{4.974E-06}$                                             & 0.221 $\pm{0.003}$                                                       & 347.198 $\pm{6.425}$                                                       \\ \hline
\begin{tabular}[c]{@{}c@{}}Baseline\\ + SAM\end{tabular}             & 39.935 $\pm{0.144}$                                                         & 0.600 $\pm{0.001}$                                                     & 0.276 $\pm{0.001}$                                                & 54.792 $\pm{0.113}$                                                          & 0.421 $\pm{0.001}$                                                               & 1.458E-03 $\pm{8.995E-05}$                                             & 1.576 $\pm{0.083}$                                                       & 1459.292 $\pm{82.220}$                                                     \\ \hline
Augmentation                                                         & 1.305 $\pm{0.076}$                                                          & \textbf{0.724 $\pm{0.001}$}                                            & \textbf{0.019 $\pm{0.001}$}                                       & \textbf{64.092 $\pm{0.152}$}                                                 & 0.217 $\pm{0.001}$                                                      & 4.741E-01 $\pm{3.822E-02}$                                             & 22.809 $\pm{0.117}$                                                      & 38465.647 $\pm{139.905}$                                                   \\ \hline
\begin{tabular}[c]{@{}c@{}}Augmentation\\ + SAM\end{tabular}         & \textbf{-1.199 $\pm{0.097}$}                                                & 0.668 $\pm{0.002}$                                                     & 0.030 $\pm{0.001}$                                                & 60.535 $\pm{0.179}$                                                          & \textbf{0.201 $\pm{0.001}$}                                                               & 4.352E-01 $\pm{2.420E-02}$                                             & 22.372 $\pm{0.042}$                                                      & 18412.664 $\pm{617.822}$                                                   \\ \hline
Weight Decay                                                         & 38.746 $\pm{0.196}$                                                         & 0.613 $\pm{0.002}$                                                     & 0.301 $\pm{0.002}$                                                & 55.077 $\pm{0.159}$                                                          & 0.402 $\pm{0.001}$                                                               & 1.359E-04 $\pm{1.030E-05}$                                             & 0.328 $\pm{0.003}$                                                       & 422.966 $\pm{6.897}$                                                       \\ \hline
\begin{tabular}[c]{@{}c@{}}Weight Decay\\ + SAM\end{tabular}         & 39.881 $\pm{0.162}$                                                         & 0.600 $\pm{0.002}$                                                     & 0.268 $\pm{0.001}$                                                & 54.797 $\pm{0.125}$                                                          & 0.419 $\pm{0.001}$                                                               & 2.890E-03 $\pm{3.102E-04}$                                             & 2.250 $\pm{0.099}$                                                       & 1908.688 $\pm{97.800}$                                                     \\ \hline
\end{tabular}}
\caption{Results for ViT trained on CIFAR10, the Mean and $\pm{}$ 1 SEM are recorded over 10 models. Numbers in bold indicate best scores for metrics. For sharpness metrics lower values represent flatter models.}
\label{tab:c10_ViT_256_0.001}
\end{table}

\begin{table}[H]
\centering
\resizebox{\textwidth}{!}{
\begin{tabular}{|c|c|c|c|c|c|c|c|c|}
\hline
\textbf{\begin{tabular}[c]{@{}c@{}}Control\\ Condition\end{tabular}} & \textbf{\begin{tabular}[c]{@{}c@{}}Generalisation \\ Gap\end{tabular}} & \textbf{\begin{tabular}[c]{@{}c@{}}Test \\ Accuracy\end{tabular}} & \textbf{\begin{tabular}[c]{@{}c@{}}Test \\ ECE\end{tabular}} & \textbf{\begin{tabular}[c]{@{}c@{}}Corruption \\ Accuracy\end{tabular}} & \textbf{\begin{tabular}[c]{@{}c@{}}Prediction \\ Disagreement\end{tabular}} & \textbf{\begin{tabular}[c]{@{}c@{}}SAM \\ Sharpness\end{tabular}} & \textbf{\begin{tabular}[c]{@{}c@{}}Fisher-Rao \\ norm\end{tabular}} & \textbf{\begin{tabular}[c]{@{}c@{}}Relative \\ Flatness\end{tabular}} \\ \hline
\begin{tabular}[c]{@{}c@{}}Baseline \\ + SAM\end{tabular}             & \xmark~(N/A)  & \xmark~(N/A) & \cmark~(1.0) & \xmark~(N/A) & \xmark~(N/A)  & \cmark~(1.0) & \cmark~(1.0) & \cmark~(1.0) \\ \hline
Augmentation                                                          & \cmark~(1.0) & \cmark~(1.0) & \cmark~(1.0) & \cmark~(1.0) & \cmark~(1.0) & \cmark~(1.0) & \cmark~(1.0) & \cmark~(1.0) \\ \hline
\begin{tabular}[c]{@{}c@{}}Augmentation \\ + SAM\end{tabular}         & \cmark~(1.0) & \cmark~(1.0) & \cmark~(1.0) & \cmark~(1.0) & \cmark~(1.0) & \cmark~(1.0) & \cmark~(1.0) & \cmark~(1.0) \\ \hline
Weight Decay                                                          & \cmark~(1.0) & \cmark~(1.0) & \cmark~(1.0) & \cmark~(1.0) & \cmark~(1.0) & \cmark~(1.0) & \cmark~(1.0) & \cmark~(1.0) \\ \hline
\begin{tabular}[c]{@{}c@{}}Weight Decay \\ + SAM\end{tabular}         & \xmark~(N/A)  & \xmark~(N/A) & \cmark~(1.0) & \xmark~(N/A) & \xmark~(N/A)  & \cmark~(1.0) & \cmark~(1.0) & \cmark~(1.0) \\ \hline
\end{tabular}}
\caption{Significance results for the control conditions against the baseline for ViT trained on CIFAR10.~\cmark~indicates significant difference compared to the baseline; \xmark~indicates significance, the value in the brackets represents the  effect sizes using the absolute RBC.}
\label{tab:vit-c10-sig}
\end{table}

With respect to significance in Table~\ref{tab:vit-c10-sig}, we can see that the application of regularisers has statistically supported sharpness for each condition. These results further align with previous data that shows regularisation can improve performance. However, we do find that only for augmentation, augmentation + SAM and weight decay this results in statistically supported improved reliability and generalisation performance.

\subsection{CIFAR100} In line with the CIFAR10 results for this architecture, we find that regularisers increase the sharpness of the resulting model and, again, that more frequently than not, this increased sharpness coincides with improved generalisation and reliability performance. The ViT's baseline model sharpness also exceeds that of ResNet and VGG, which are trained with the same batch size and learning rate. As a result, this adds strength to the idea that ViT's learning biases the models towards narrower decision boundaries in general. In Table~\ref{tab:c100_ViT_256_0.001}, augmentation and augmentation + SAM conditions achieve the best performance and exhibit the highest sharpness values across metrics, which is consistent with our previous findings.
\begin{table}[H]
\centering
\resizebox{\textwidth}{!}{
\begin{tabular}{|c|c|c|c|c|c|c|c|c|}
\hline
\textbf{\begin{tabular}[c]{@{}c@{}}Control\\ Condition\end{tabular}} & \textbf{\begin{tabular}[c]{@{}c@{}}Generalisation \\ Gap\end{tabular}} & \textbf{\begin{tabular}[c]{@{}c@{}}Test \\ Accuracy\end{tabular}} & \textbf{\begin{tabular}[c]{@{}c@{}}Test\\ ECE\end{tabular}} & \textbf{\begin{tabular}[c]{@{}c@{}}Corruption \\ Accuracy\end{tabular}} & \textbf{\begin{tabular}[c]{@{}c@{}}Prediction \\ Disagreement\end{tabular}} & \textbf{\begin{tabular}[c]{@{}c@{}}SAM \\ Sharpness\end{tabular}} & \textbf{\begin{tabular}[c]{@{}c@{}}Fisher-Rao \\ norm\end{tabular}} & \textbf{\begin{tabular}[c]{@{}c@{}}Relative \\ Flatness\end{tabular}} \\ \hline
Baseline                                                             & 69.048 $\pm{0.164}$                                                         & 0.309 $\pm{0.002}$                                                     & 0.402 $\pm{0.002}$                                               & 25.936 $\pm{0.088}$                                                          & 0.723 $\pm{0.000}$                                                               & 3.428E-04 $\pm{4.954E-05}$                                             & 0.646 $\pm{0.061}$                                                       & 112.185 $\pm{4.246}$                                                       \\ \hline
\begin{tabular}[c]{@{}c@{}}Baseline\\ + SAM\end{tabular}             & 67.376 $\pm{0.126}$                                                         & 0.326 $\pm{0.001}$                                                     & 0.386 $\pm{0.001}$                                               & 27.628 $\pm{0.097}$                                                          & 0.697 $\pm{0.000}$                                                               & 4.539E-04 $\pm{6.066E-05}$                                             & 0.821 $\pm{0.070}$                                                       & 124.472 $\pm{30.314}$                                                      \\ \hline
Augmentation                                                         & 37.472 $\pm{0.249}$                                                         & 0.508 $\pm{0.001}$                                                     & 0.227 $\pm{0.001}$                                               & 38.680 $\pm{0.091}$                                                          & 0.483 $\pm{0.001}$                                                               & 5.995E-01 $\pm{8.815E-02}$                                             & 17.321 $\pm{0.192}$                                                      & 17401.462 $\pm{143.009}$                                                   \\ \hline
\begin{tabular}[c]{@{}c@{}}Augmentation\\ + SAM\end{tabular}         & \textbf{32.136 $\pm{0.262}$}                                                & \textbf{0.523 $\pm{0.001}$}                                            & \textbf{0.146 $\pm{0.002}$}                                      & \textbf{40.275 $\pm{0.097}$}                                                 & 0.446 $\pm{0.000}$                                                               & 4.649E-01 $\pm{2.505E-02}$                                             & 19.664 $\pm{0.127}$                                                      & 17812.985 $\pm{55.523}$                                                    \\ \hline
Weight Decay                                                         & 67.524 $\pm{0.148}$                                                         & 0.325 $\pm{0.001}$                                                     & 0.324 $\pm{0.001}$                                               & 27.364 $\pm{0.103}$                                                          & 0.700 $\pm{0.000}$                                                               & 8.440E-04 $\pm{1.251E-04}$                                             & 1.563 $\pm{0.073}$                                                       & 251.148 $\pm{15.330}$                                                      \\ \hline
\begin{tabular}[c]{@{}c@{}}Weight Decay\\ + SAM\end{tabular}         & 67.227 $\pm{0.077}$                                                         & 0.327 $\pm{0.001}$                                                     & 0.284 $\pm{0.001}$                                               & 27.739 $\pm{0.069}$                                                          & 0.695 $\pm{0.001}$                                                               & 4.323E-03 $\pm{3.837E-04}$                                             & 5.181 $\pm{0.260}$                                                       & 1554.595 $\pm{91.649}$                                                     \\ \hline
\end{tabular}}
\caption{Results for ViT trained on CIFAR100, the Mean and $\pm{}$ 1 SEM are recorded over 10 models. Numbers in bold indicate best scores for metrics. For sharpness metrics lower values represent flatter models.}
\label{tab:c100_ViT_256_0.001}
\end{table}
\begin{table}[H]
\centering
\resizebox{\textwidth}{!}{
\begin{tabular}{|c|c|c|c|c|c|c|c|c|}
\hline
\textbf{\begin{tabular}[c]{@{}c@{}}Control\\ Condition\end{tabular}} & \textbf{\begin{tabular}[c]{@{}c@{}}Generalisation \\ Gap\end{tabular}} & \textbf{\begin{tabular}[c]{@{}c@{}}Test \\ Accuracy\end{tabular}} & \textbf{\begin{tabular}[c]{@{}c@{}}Test \\ ECE\end{tabular}} & \textbf{\begin{tabular}[c]{@{}c@{}}Corruption \\ Accuracy\end{tabular}} & \textbf{\begin{tabular}[c]{@{}c@{}}Prediction \\ Disagreement\end{tabular}} & \textbf{\begin{tabular}[c]{@{}c@{}}SAM \\ Sharpness\end{tabular}} & \textbf{\begin{tabular}[c]{@{}c@{}}Fisher-Rao\\ norm\end{tabular}} & \textbf{\begin{tabular}[c]{@{}c@{}}Relative \\ Flatness\end{tabular}} \\ \hline
\begin{tabular}[c]{@{}c@{}}Baseline \\ + SAM\end{tabular}            & \cmark~(1.0) & \cmark~(1.0) & \cmark~(1.0) & \cmark~(1.0) & \cmark~(1.0) & \xmark~(N/A)  & \xmark~(N/A) & \xmark~(N/A) \\ \hline
Augmentation                                                         & \cmark~(1.0) & \cmark~(1.0) & \cmark~(1.0) & \cmark~(1.0) & \cmark~(1.0) & \cmark~(1.0)  & \cmark~(1.0) & \cmark~(1.0) \\ \hline
\begin{tabular}[c]{@{}c@{}}Augmentation \\ + SAM\end{tabular}        & \cmark~(1.0) & \cmark~(1.0) & \cmark~(1.0) & \cmark~(1.0) & \cmark~(1.0) & \cmark~(1.0)  & \cmark~(1.0) & \cmark~(1.0) \\ \hline
Weight Decay                                                         & \cmark~(1.0) & \cmark~(1.0) & \cmark~(1.0) & \cmark~(1.0) & \cmark~(1.0) & \cmark~(0.93) & \cmark~(1.0) & \cmark~(1.0) \\ \hline
\begin{tabular}[c]{@{}c@{}}Weight Decay \\ + SAM\end{tabular}        & \cmark~(1.0) & \cmark~(1.0) & \cmark~(1.0) & \cmark~(1.0) & \cmark~(1.0) & \cmark~(1.0)  & \cmark~(1.0) & \cmark~(1.0) \\ \hline
\end{tabular}}
\caption{Significance results for the control conditions against the baseline for ViT trained on CIFAR100.~\cmark~indicates significant difference compared to the baseline; \xmark~indicates no significance, the value in the brackets represents the  effect sizes using the absolute RBC.}
\label{tab:vit-c100-sig}
\end{table}

It can be observed in Table~\ref{tab:vit-c100-sig} that the application of regularisation results in statistically supported increased sharpness. However, only for the conditions of weight decay, weight decay+SAM, augmentation and augmentation + SAM do we observe that this increased sharpness coincides with improved generalisation and reliability. We hypothesise that the nuanced relationship with the application of SAM for the ViT could be related to its initial sharp inductive bias, however, more experiments would be needed to confirm this. 

\section{SAM Sharpness \texorpdfstring{$\boldsymbol{\rho}$}{ρ}  Radius Correlation}
\label{app:sam-sharpness-rho}
\begin{figure}[H]
    \centering
    \includegraphics[width=\linewidth]{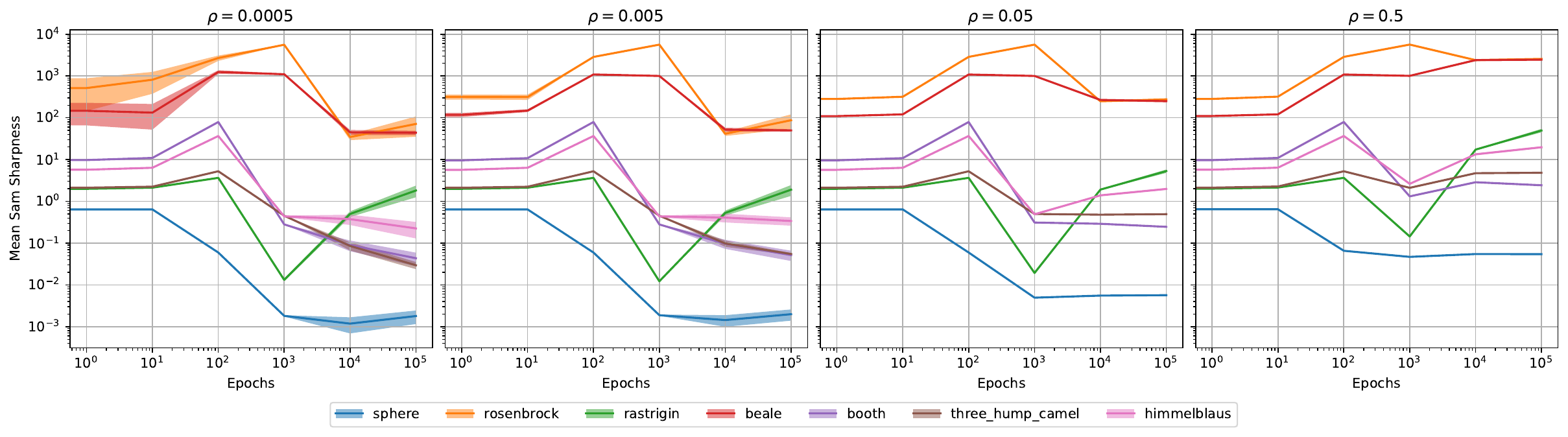}
    \caption{Training an MLP on single-objective problems mean SAM sharpness vs epochs with varying $\rho$. First $\rho=0.0005$, second $\rho=0.005$, third $\rho=0.05$ and fourth $\rho=0.5$. The results are averaged over 10 runs.}
    \label{fig:rho_and_sam_sharpness}
\end{figure}

This section explores the robustness of the mean SAM sharpness value with respect to the $\rho$ hyperparameter. We conduct the same experimental setup for single objective optimisation and record the corresponding mean sharpness with $\rho \in [0.005, 0.005, 0.05,0.5]$ at epochs $0, 10^1, 10^2, 10^3, 10^4$ and $10^5$. Figure \ref{fig:rho_and_sam_sharpness} shows the mean SAM sharpness across epochs when using different values of $\rho$ to calculate the mean SAM sharpness value. Across $\rho$ values there is a consistent trend, the sphere task always results in the lowest sharpness, the rastrigin task has a dip at epoch $10^3$ which subsequently increases. The rosenbrock and beale tasks result in the sharpest models at every epoch. The marked difference between the plots is that as $\rho$ increases the mean SAM sharpness across the board, however the general relationship between tasks is maintained as shown with the results in Tables~\ref{ss-per-epoch} and \ref{ss-global}. These tables  compare the mean SAM sharpness global relationship with a $\rho$ of 0.005, 0.05 and 0.5 at different training epochs (Table~\ref{ss-per-epoch} and across epochs and Table~\ref{ss-global} with $\rho=0.005$ the $\rho$ used in results above and in the main paper). The tables show that there is a strong correlation with the results across the $\rho$ hyperparameter.

\begin{table}[H]

    \begin{minipage}{.475\linewidth}

      \centering
        \begin{tabular}{|c|ccc|}
            \hline
            \multirow{2}{*}{\textbf{Epoch}} & \multicolumn{3}{c|}{\textbf{$\boldsymbol{\rho}$ Value}} \\ \cline{2-4} 
             & \multicolumn{1}{c|}{\textbf{0.0005}} & \multicolumn{1}{c|}{\textbf{0.05}} & \textbf{0.5} \\ \hline
            $0$ & \multicolumn{1}{c|}{0.352} & \multicolumn{1}{c|}{0.878} & 0.876 \\ \hline
            $10^0$ & \multicolumn{1}{c|}{0.400} & \multicolumn{1}{c|}{0.895} & 0.894 \\ \hline
            $10^1$ & \multicolumn{1}{c|}{0.655} & \multicolumn{1}{c|}{0.888} & 0.886 \\ \hline
            $10^2$ & \multicolumn{1}{c|}{0.889} & \multicolumn{1}{c|}{1.000} & 1.000 \\ \hline
            $10^3$ & \multicolumn{1}{c|}{0.999} & \multicolumn{1}{c|}{1.000} & 1.000 \\ \hline
            $10^4$ & \multicolumn{1}{c|}{0.992} & \multicolumn{1}{c|}{0.949} & 0.947 \\ \hline
            $10^5$ & \multicolumn{1}{c|}{0.990} & \multicolumn{1}{c|}{0.676} & 0.593 \\ \hline
        \end{tabular}%
      \caption{The mean SAM sharpness global correlation to $\rho$ = 0.005 at different training epochs.}
      \label{ss-per-epoch}
    \end{minipage} \hfill
    \begin{minipage}{.475\linewidth}
      \centering
        \begin{tabular}{|c|c|c|c|}
            \hline
            \textbf{$\boldsymbol{\rho}$ Value} & \textbf{0.0005} & \textbf{0.05} & \textbf{0.5} \\ \hline
            Correlation & 0.939 & 0.997 & 0.791 \\ \hline
        \end{tabular}%
            \caption{The mean SAM sharpness radius global correlation to $\rho$ = 0.005 across epochs.}
            \label{ss-global}
    \end{minipage}     
\end{table}

\section{Language Model Results}
\label{Language_model_results}

While models trained in the computer vision domain have dominated the study of neural networks geometry, language models have recently become an object of attention due to the advent of large language models. As such, we see them as of interest when establishing relationships between regularisation, memorisation, generalisation and minima geometry in the language domain. Additionally, exploring this domain allows a study of the generality of our results in high-dimensional settings on the language domain.

To do this we generated the NAMES language dataset using the Faker Python package\footnote{\url{https://faker.readthedocs.io/}}. The dataset can be found online\footnote{\url{https://huggingface.co/datasets/ifm23/NAMES}}. The dataset consisted of 15 country categories:  

\begin{multicols}{3}
\begin{itemize}
    \item English (en\_GB)
    \item German (de\_DE)
    \item French (fr\_FR)
    \item Italian (it\_IT)
    \item Spanish (es\_ES)
    \item Portuguese (pt\_PT)
    \item Russian (ru\_RU)
    \item Polish (pl\_PL)
    \item Dutch (nl\_NL)
    \item Japanese (ja\_JP)
    \item Norwegian (no\_NO)
    \item Chinese (zh\_CN)
    \item Czech (cs\_CZ)
    \item Greek (el\_GR)
    \item Turkish (tr\_TR)
\end{itemize}
\end{multicols}
 
 For the Japanese and Chinese names, the romanised form was used. To ensure class balance for each country category, we generated 1000 unique first and last name combinations with a maximum length of 20, totalling 15000 unique names.  Each category was split into an 80-20 train (800 samples) and test (200 samples) split to ensure class balance during training and evaluation. The names were character tokenised using the Roman alphabet with lowercase and uppercase characters, as well as the "'" and "[SPACE]" characters. The "[UNK]" token was used for characters not in this set, and "[PAD]" was used for applying padding. 

For the neural network we train a GPT2 architecture \citep{radford2019language}, with embedding dimension of 128, 4 layers, 4 attention heads, a maximum sequence length of 25 characters and a vocab size of 56, on the NAMES dataset for 200 epochs using SGD and a momentum of 0.9 with a learning rate of $1e^{-3}$. Left padding was applied to all inputs had a length of 25. To ensure a consistent experimental setup we use the same regularisers: weight decay ($5e^{-3}$), SAM and augmentation and record the same sharpness metrics. For this dataset augmentation uses the nlpaug.augmenter.char package using KeyboardAug at a probability of 0.1~\citep{ma2019nlpaug}. All reliability metrics are maintained apart from corruption accuracy as there is no such dataset for this task. 

\begin{table}[H]
\centering
\resizebox{\textwidth}{!}{
\begin{tabular}{|c|c|c|c|c|c|c|c|}
\hline
\textbf{\begin{tabular}[c]{@{}c@{}}Control\\ Conditon\end{tabular}} & \textbf{\begin{tabular}[c]{@{}c@{}}Generalisation \\ Gap\end{tabular}} & \textbf{\begin{tabular}[c]{@{}c@{}}Test \\ Accuracy\end{tabular}} & \textbf{\begin{tabular}[c]{@{}c@{}}Test \\ ECE\end{tabular}} & \textbf{\begin{tabular}[c]{@{}c@{}}Prediction \\ Disagreement\end{tabular}} & \textbf{\begin{tabular}[c]{@{}c@{}}Fisher Rao \\ norm\end{tabular}} & \textbf{\begin{tabular}[c]{@{}c@{}}SAM \\ Sharpness\end{tabular}} & \textbf{\begin{tabular}[c]{@{}c@{}}Relative \\ Flatness\end{tabular}} \\ \hline
Baseline                                                            & 12.393 $\pm{0.158}$                                                         & 0.876 $\pm{0.002}$                                                     & 0.080 $\pm{0.001}$                                                & 0.109 $\pm{0.001}$                                                               & 0.044 $\pm{0.000}$                                                       & 3.658E-05 $\pm{3.077E-06}$                                             & 13.968 $\pm{0.081}$                                                        \\ \hline
\begin{tabular}[c]{@{}c@{}}Baseline\\ + SAM\end{tabular}            & 12.282 $\pm{0.147}$                                                         & 0.874 $\pm{0.001}$                                                     & 0.054 $\pm{0.001}$                                                & 0.117 $\pm{0.001}$                                                               & 0.457 $\pm{0.012}$                                                       & 5.861E-03 $\pm{1.046E-03}$                                             & 188.148 $\pm{3.816}$                                                       \\ \hline
Augmentation                                                        & 10.239 $\pm{0.110}$                                                         & \textbf{0.888 $\pm{0.001}$}                                            & 0.062 $\pm{0.001}$                                                & 0.114 $\pm{0.001}$                                                               & 0.767 $\pm{0.021}$                                                       & 4.238E-01 $\pm{5.088E-02}$                                             & 218.364 $\pm{3.721}$                                                       \\ \hline
\begin{tabular}[c]{@{}c@{}}Augmentation\\ + SAM\end{tabular}        & \textbf{5.187 $\pm{0.166}$}                                                 & 0.871 $\pm{0.002}$                                                     & \textbf{0.016 $\pm{0.001}$}                                       & \textbf{0.094 $\pm{0.001}$}                                                      & 1.810 $\pm{0.024}$                                                       & 6.950E-01 $\pm{7.846E-02}$                                             & 1080.307 $\pm{6.488}$                                                      \\ \hline
Weight Decay                                                        & 12.320 $\pm{0.155}$                                                         & 0.877 $\pm{0.002}$                                                     & 0.078 $\pm{0.001}$                                                & 0.108 $\pm{0.001}$                                                               & 0.050 $\pm{0.000}$                                                       & 4.043E-05 $\pm{3.204E-06}$                                             & 15.427 $\pm{0.095}$                                                        \\ \hline
\begin{tabular}[c]{@{}c@{}}Weight Decay\\ + SAM\end{tabular}        & 12.242 $\pm{0.190}$                                                         & 0.874 $\pm{0.002}$                                                     & 0.052 $\pm{0.002}$                                                & 0.116 $\pm{0.001}$                                                               & 0.499 $\pm{0.013}$                                                       & 6.311E-03 $\pm{1.028E-03}$                                             & 203.940 $\pm{3.967}$                                                       \\ \hline
\end{tabular}}
\caption{Results for GPT2 architecture trained on the NAMES dataset.}
\label{tab:GPT2-classifcation}
\end{table}

\begin{table}[H]
\centering
\resizebox{\textwidth}{!}{
\begin{tabular}{|c|c|c|c|c|c|c|c|}
\hline
\textbf{\begin{tabular}[c]{@{}c@{}}Control\\ Condition\end{tabular}} & \textbf{\begin{tabular}[c]{@{}c@{}}Generalisation \\ Gap\end{tabular}} & \textbf{\begin{tabular}[c]{@{}c@{}}Test \\ Accuracy\end{tabular}} & \textbf{\begin{tabular}[c]{@{}c@{}}Test \\ ECE\end{tabular}} & \textbf{\begin{tabular}[c]{@{}c@{}}Prediction \\ Disagreement\end{tabular}} & \textbf{\begin{tabular}[c]{@{}c@{}}SAM \\ Sharpness\end{tabular}} & \textbf{\begin{tabular}[c]{@{}c@{}}Fisher-Rao \\ norm\end{tabular}} & \textbf{\begin{tabular}[c]{@{}c@{}}Relative \\ Flatness\end{tabular}} \\ \hline
\begin{tabular}[c]{@{}c@{}}Baseline\\ + SAM\end{tabular}             & \xmark (N/A)                                            & \xmark (N/A)                                       & \cmark (1.0)                                 & \xmark (N/A)                                                 & \cmark (1.0)                                       & \cmark (1.0)                                         & \cmark (1.0)                                           \\ \hline
Augmentation                                                         & \cmark (1.0)                                           & \cmark (0.96)                                      & \cmark (1.0)                                 & \xmark (N/A)                                                 & \cmark (1.0)                                       & \cmark (1.0)                                         & \cmark (1.0)                                           \\ \hline
\begin{tabular}[c]{@{}c@{}}Augmentation\\ + SAM\end{tabular}         & \cmark (1.0)                                           & \xmark (N/A)                                       & \cmark (1.0)                                 & \cmark (1.0)                                                & \cmark (1.0)                                       & \cmark (1.0)                                         & \cmark (1.0)                                           \\ \hline
Weight Decay                                                         & \xmark (N/A)                                            & \xmark (N/A)                                       & \cmark (0.82)                                & \cmark (0.42)                                               & \cmark (1.0)                                       & \cmark (1.0)                                         & \cmark (1.0)                                           \\ \hline
\begin{tabular}[c]{@{}c@{}}Weight Decay\\ + SAM\end{tabular}         & \xmark (N/A)                                            & \xmark (N/A)                                       & \cmark (1.0)                                 & \xmark (N/A)                                                 & \cmark (1.0)                                       & \cmark (1.0)                                         & \cmark (1.0)                                           \\ \hline
\end{tabular}}
\caption{Significance results for the control conditions against the baseline for GPT2 trained on NAMES.~\cmark~indicates significant difference compared to the baseline; \xmark~indicates no significance, the value in the brackets represents the  effect sizes using the absolute RBC.}
\label{tab:GPT2-classifcation-sig}
\end{table}

When considering the results in the language domain, it is clear that they largely mirror those of the vision domain. As seen in Table~\ref{tab:GPT2-classifcation} it is clear that the use of regularisation such as augmentation can increase the test accuracy and reduce the generational gap of the model as well as improving the calibration of the model. With respect to sharpness we observe that the model becomes sharper under the application of regularisation in Table~\ref{tab:GPT2-classifcation-sig}. When regularises are applied together such as weight decay and SAM as well as augmentation and SAM it can be seen that sharpness increases with additional benefits to reliability metrics compared to the unregularised baselines. 

\textit{Relevance for Large Language Models and Reinforcement Learning Domains}. We are keen not to overextend the insights of this paper outside of the domains studied. However, with respect to language models and reinforcement learning, we believe that given the empirical evidence provided in the paper, these results and insights would likely extend to these cases. We show in this appendix section for language classification that regularisation can improve model performance from a generalisation and reliability perspective while increasing sharpness and that there is a multiplicative effect when regularisers are applied together. We further posit that function complexity can interact with language models via disjoint concepts, different language structures in a myriad of task depended manners, however future work should focus on establishing this at scale. For reinforcement learning, it could also be expected that the number of environment interactions could represent a similar decision boundary complexity proxy. However, we would like to caveat these points with the acknowledgement that the results in this paper are predominately centred on computer vision, synthetic datasets and a small language dataset and, therefore, the function-centric lens of minima sharpness would be the most useful contribution when considering how these results extend to large language models and reinforcement learning domains. We think that it would be interesting future work to verify these impacts in these settings, starting with the hypothesis of regularisation and sharpness that this work offers.  

\end{document}